\documentclass[11pt]{article}

\usepackage[margin=1in]{geometry}
\usepackage[T1]{fontenc}
\usepackage[utf8]{inputenc}
\usepackage{lmodern}
\usepackage{microtype}
\usepackage{amsmath,amssymb}
\usepackage{graphicx}
\usepackage[hyphens]{url}
\usepackage{natbib}
\usepackage{caption}
\usepackage{tabularx}
\usepackage{array}
\usepackage{ragged2e}
\usepackage{float}
\usepackage{afterpage}
\usepackage{xcolor}
\usepackage{fvextra}
\usepackage{needspace}
\usepackage[
  colorlinks=true,
  urlcolor=blue,
  linkcolor=black,
  citecolor=black
]{hyperref}

\hypersetup{
  pdftitle={ScienceCUA: Resource-to-Skill Construction and Execution-Guided Skill Evolution for Scientific Computer Use Agents},
  pdfauthor={Daiqiang Li, Jingyi Peng, Rui Hu, Jia Li, Xuyang Liu, Guanren Qiao, Yujia Xiang, Qiuzhuang Sun, Zhiping Xu, Mingrui Luo},
  pdfsubject={Computer-use agents for scientific software},
  pdfkeywords={ScienceCUA, computer-use agents, scientific software, resource-to-skill construction, skill evolution},
  pdfdisplaydoctitle=true
}
\usepackage{booktabs}
\usepackage{wrapfig}
\usepackage{longtable}
\usepackage{tikz}
\usetikzlibrary{arrows.meta,positioning,fit}

\definecolor{DeepAutonomyBlue}{HTML}{007EFF}
\definecolor{DeepAutonomyDarkBlue}{HTML}{004C99}
\newcommand{\sciencecuawordmark}{%
  \raisebox{-0.015em}{\includegraphics[height=0.76em]{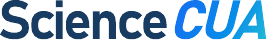}}%
}
\newsavebox{\scienceabstractbox}
\renewenvironment{abstract}{%
  \par\addvspace{0.5em}%
  \begin{lrbox}{\scienceabstractbox}%
    \begin{minipage}{\dimexpr\textwidth-28pt-0.6pt\relax}
      \fontsize{9.5}{11.2}\selectfont
      \setlength{\parindent}{0pt}%
      {\centering\large\bfseries\color{DeepAutonomyDarkBlue}Abstract\par}
      \vspace{0.45em}%
      \ignorespaces
}{%
      \par
    \end{minipage}%
  \end{lrbox}%
  \noindent
  \begin{tikzpicture}
    \node[draw=DeepAutonomyBlue!38,fill=DeepAutonomyBlue!4,
      rounded corners=5pt,line width=0.6pt,
      inner xsep=14pt,inner ysep=10pt,outer sep=0pt]
      {\usebox{\scienceabstractbox}};
  \end{tikzpicture}%
  \par\addvspace{0.45em}%
}

\usepackage{amsmath,amsfonts,bm}

\def\eqref#1{equation~\ref{#1}}

\def\1{\bm{1}}

\DeclareMathAlphabet{\mathsfit}{\encodingdefault}{\sfdefault}{m}{sl}
\SetMathAlphabet{\mathsfit}{bold}{\encodingdefault}{\sfdefault}{bx}{n}

\usepackage{sciencecua-tables}

\makeatletter
\renewcommand{\section}{\@startsection{section}{1}{\z@}%
  {-3.5ex\@plus -1ex\@minus -.2ex}%
  {2.3ex\@plus .2ex}%
  {\normalfont\Large\bfseries\color{DeepAutonomyDarkBlue}}}

\newcommand{\rqsubsection}{\@startsection{subsection}{2}{\z@}%
  {-3.25ex\@plus -1ex\@minus -.2ex}%
  {1.5ex\@plus .2ex}%
  {\normalfont\fontsize{8.5}{10.5}\selectfont\bfseries}}
\makeatother

\begin{document}

\vspace*{-0.55in}

\noindent\begin{minipage}[t]{\textwidth}
  \includegraphics[width=5.0cm,trim=73.44bp 246.72bp 73.92bp 252.72bp,clip,keepaspectratio]{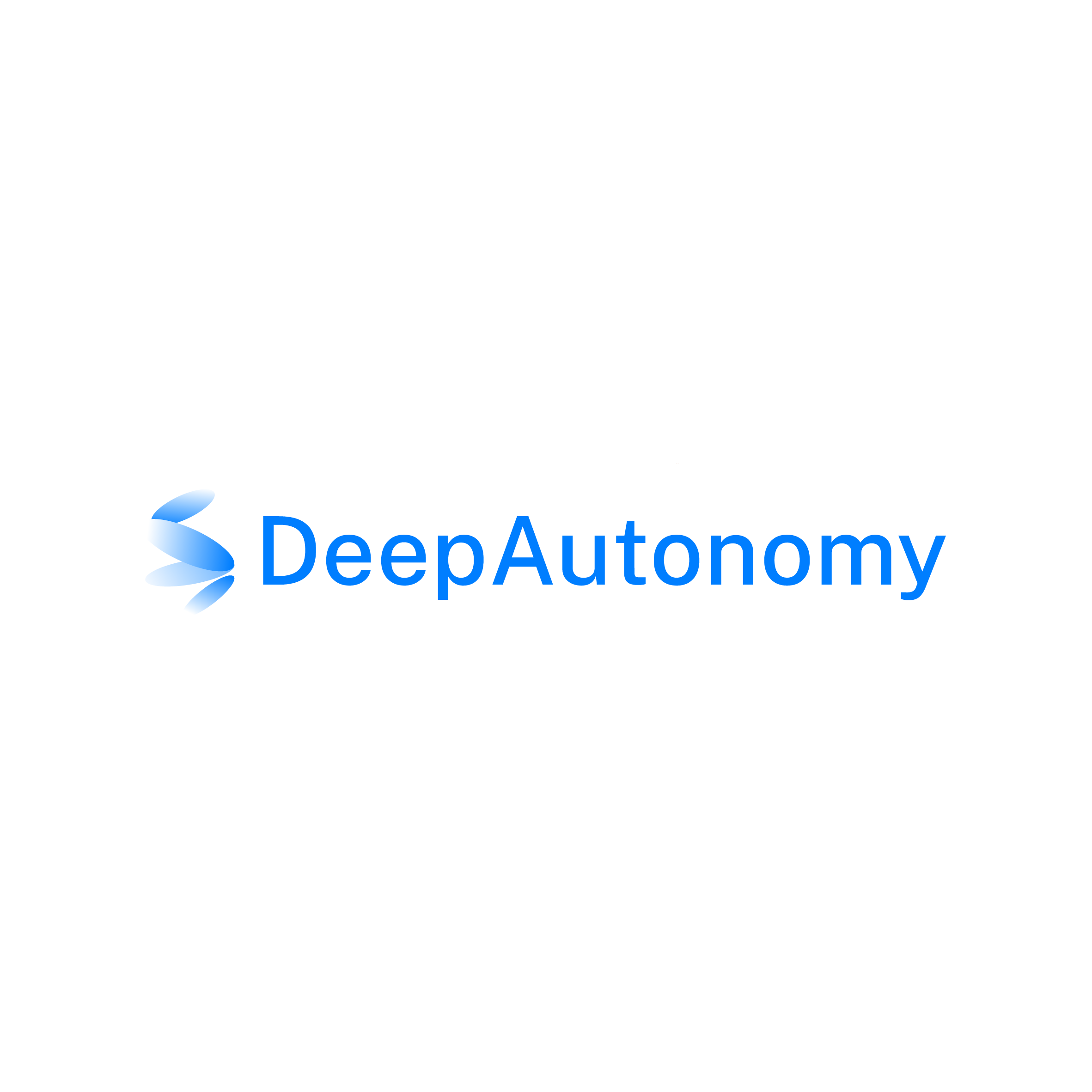}
\end{minipage}

\vspace{0.3em}

\noindent\rule{\textwidth}{0.4pt}

\vspace{0.75em}

\begin{center}
  {\fontsize{14.5}{18}\selectfont\bfseries
  \sciencecuawordmark{}: Resource-to-Skill Construction and Execution-\\
  Guided Skill Evolution for Scientific Computer Use Agents
  \par}

  \vspace{0.6em}

  {\normalsize
  Daiqiang Li\textsuperscript{1},
  Jingyi Peng\textsuperscript{1},
  Rui Hu\textsuperscript{2},
  Jia Li\textsuperscript{3},
  Xuyang Liu\textsuperscript{4},
  Guanren Qiao\textsuperscript{1},
  Yujia Xiang\textsuperscript{1},
  Qiuzhuang Sun\textsuperscript{5},
  Zhiping Xu\textsuperscript{1},
  Mingrui Luo\textsuperscript{1,\dag}
  \par}

  \vspace{0.3em}

  {\small
  \makebox[\textwidth][c]{%
    \textsuperscript{1}DeepAutonomy Technology Co., Ltd.\quad
    \textsuperscript{2}Peking University\quad
    \textsuperscript{3}College of AI, Tsinghua University}\\[0.25em]
  \makebox[\textwidth][c]{%
    \textsuperscript{4}The Hong Kong Polytechnic University\quad
    \textsuperscript{5}Singapore Management University}
  \par}

  \vspace{0.45em}

  {\small\bfseries
  \href{https://hszd-team.github.io/ScienceCUA-website/}{%
    \raisebox{-0.22em}{\includegraphics[height=1.1em]{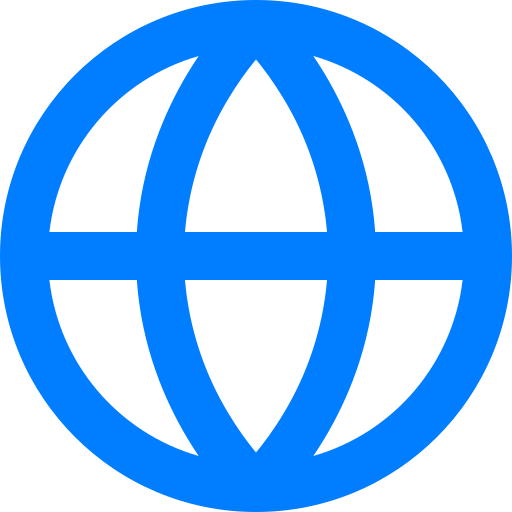}}\hspace{0.35em}Project Page}%
  \hspace{1.8em}%
  \href{https://github.com/HSZD-Team/ScienceCUA}{%
    \raisebox{-0.22em}{\includegraphics[height=1.1em]{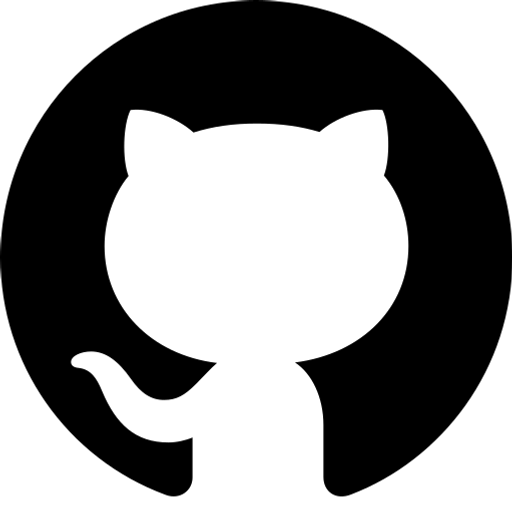}}\hspace{0.35em}Code}%
  \hspace{1.8em}%
  \href{https://huggingface.co/lidaiqiang/ScienceBoard_Verified_Env}{%
    \raisebox{-0.22em}{\includegraphics[height=1.1em]{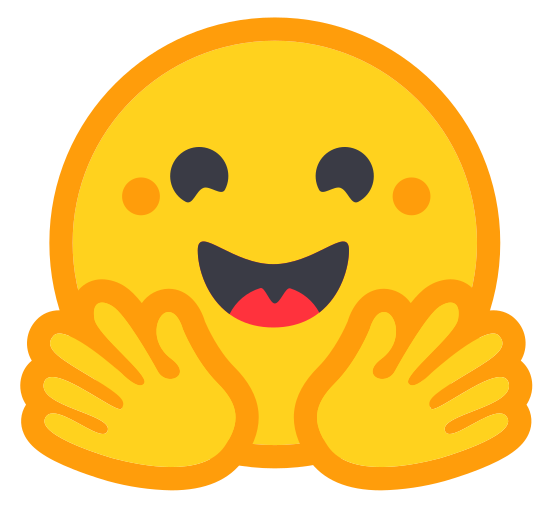}}\hspace{0.35em}Benchmark Environment}
  \par}

\end{center}

\begingroup
  \let\footnotesize\small
  \renewcommand{\thefootnote}{\textdagger}
  \footnotetext[1]{Corresponding author}
\endgroup

\vspace{-0.8em}

\begin{abstract}
Computer-use agents for scientific software require application-specific expertise. Much of this expertise is already documented in human-authored software manuals and web pages, making them readily available resources for improving agents’ operational capabilities. However, turning these resources into executable, reusable skills remains challenging due to mismatches between documented and deployed software versions and differences in scope between documentation coverage and task requirements. To address these challenges, we introduce ScienceCUA, a two-stage framework combining resource-to-skill construction with execution-guided skill evolution. First, ScienceCUA identifies tasks supported by the software from its documentation, infers operational trajectories, and generalizes them into reusable Skills. Command catalogs, syntax constraints, and shortcuts are organized separately as References for on-demand lookup. Then, ScienceCUA uses execution feedback to drive skill self-evolution by contrasting successful and failed trajectories for the same task, identifying deficiencies, revising existing skills or adding new ones, and updating the skill library only after repeated validation in real environments. To enable reliable evaluation without benchmark noise or task leakage, we introduce two improved benchmarks. ScienceBoard-Verified provides an audited and corrected basis for measuring task performance, while ScienceBoard-VE separates skill evolution from final testing through disjoint evolution and test tasks and further supports native-agent evaluation. Experiments show that the initial Skills and References improve average task success rate by up to 7.23 percentage points over the base model, while skill self-evolution increases the total gain over the baseline to as much as 17.12 percentage points. ScienceCUA turns software documentation into an evolving set of executable skills and queryable references, offering a practical way for general-purpose agents to use scientific software reliably and at scale.
\end{abstract}

\noindent\begin{minipage}{\textwidth}
  \centering
  \includegraphics[width=\linewidth]{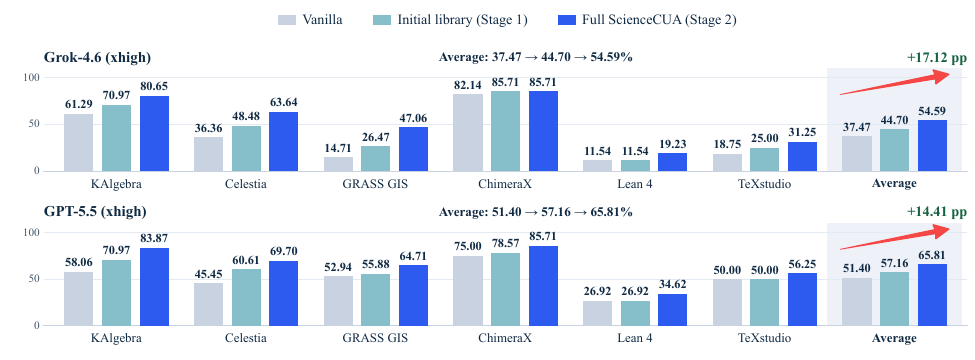}
  \captionsetup{font=footnotesize,skip=3pt,justification=raggedright,singlelinecheck=false}
  \captionof{figure}{Two-stage gains on ScienceBoard-Verified. Success rates (\%) for Vanilla, the initial library after Resource-to-Skill construction, and Full ScienceCUA after skill self-evolution.}
  \label{fig:two_stage_gains}
\end{minipage}

\clearpage
\section{Introduction}
\label{sec:introduction}

Computer-use agents (CUAs) built on foundation models can now execute multi-step tasks through graphical and command-line interfaces~\citep{qin2025ui,sager2026survey,xie2024osworld}. Most existing benchmarks, including WebArena, OSWorld, and Windows Agent Arena, focus on general-purpose web, office, and desktop tasks, while scientific software has received comparatively less attention~\citep{xie2024osworld,zhou2024webarena,bonatti2024windowsagentarena,sun2026scienceboard}. Scientific software is widely used for experiments, simulations, and data analysis, but operating these tools often requires application-specific workflows, commands, syntax, and parameter constraints~\citep{sun2026scienceboard,ai2026scivisagentbench}. To complete scientific-software tasks, agents must use software-specific knowledge to select appropriate operations and organize them into effective workflows~\citep{shao2026bioinfoysis}.

\begin{figure}[!t]
    \centering
    \includegraphics[width=\linewidth]{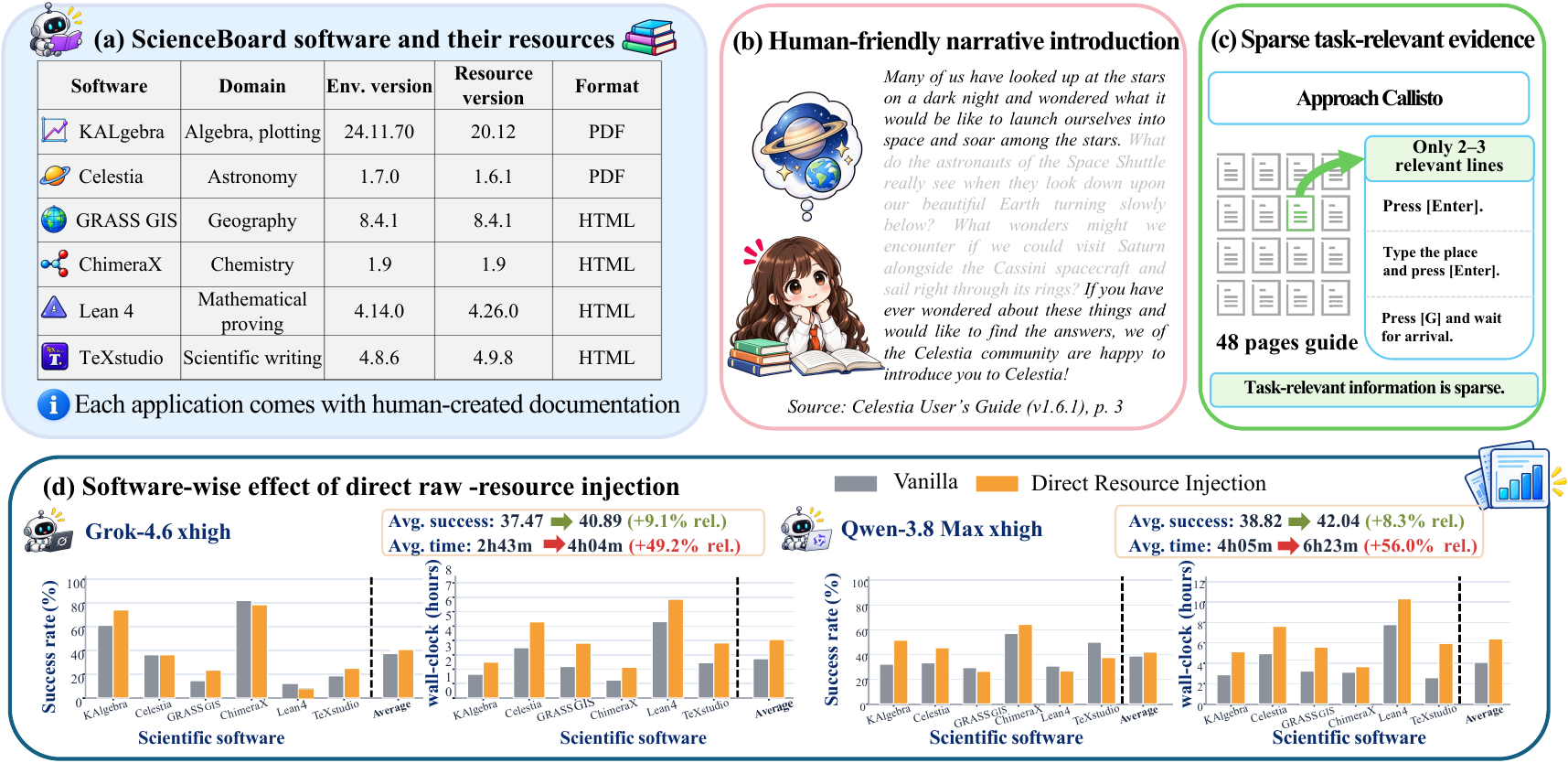}
    \caption{Motivation for resource-to-skill construction. (a) Scientific applications and their resources. (b) Human-oriented narrative content. (c) Sparse task-relevant information. (d) Effects of directly injecting raw resources on success rate and total execution time.}
    \label{fig:motivation}
\end{figure}

We study this problem using ScienceBoard~\citep{sun2026scienceboard}, which evaluates CUAs on six scientific applications. Figure~\ref{fig:motivation}(a) shows that each application provides user manuals or official tutorials containing application-specific knowledge. However, Figure~\ref{fig:motivation}(d) shows that directly placing these resources in the model context yields only limited gains in task success while substantially increasing execution time. Figure~\ref{fig:motivation}(b) illustrates that human-authored documentation mixes operational instructions with conceptual explanations and other narrative content. A particular task, however, typically depends on only a small fraction of the available information, as shown in Figure~\ref{fig:motivation}(c). The central question is how to extract task-relevant information from these human-oriented resources and organize it as structured knowledge that the model can use effectively.

Prior work has constructed reusable skills from human-created resources for content creation, gaming, software operation, and related applications~\citep{fan2026resource2skill,che2026mmg2skill,jiang2026visualskill,shen2026skillfoundry}. For scientific software, these resources must supply two kinds of knowledge with different lifetimes: procedural guidance for a specific task, and shared software knowledge queried on demand. We therefore introduce ScienceCUA, which extracts task-oriented procedural knowledge and cross-task shared software knowledge from raw software resources, organizes them as Skills and References, respectively, and combines them on demand during execution.

However, the initial Skills and References constructed from raw resources have inherent limitations. First, the available PDF or HTML resources may not match the version of the software deployed in the execution environment, since perfectly matched resources are not always available. Second, for some applications, the resource coverage is broader than the range of tasks being evaluated, while the task scope is unknown during construction. The resulting knowledge library therefore contains many entries that are irrelevant to the evaluation tasks, reducing the density of useful information. Third, some task requirements may fall outside the coverage of the resources themselves. 

To address these limitations, ScienceCUA drives the continual evolution of Skills and References through execution feedback. Prior work has used task feedback and execution trajectories to build reflective memory, optimize external skills, or adapt agent behavior~\citep{shinn2023reflexion,zhao2024expel,ni2026trace2skill,liu2026skillforge,yang2026skillopt,chen2026skillcat,zhang2026coevoskills,sun2025seagent}. In contrast, ScienceCUA improves a scientific-software knowledge library initialized from software resources, using historical failure trajectories to guide subsequent exploration and help agents avoid repeating failed paths or continuing to follow potentially erroneous Skills and References; when knowledge is insufficient, it revisits the original resources to retrieve and construct task-relevant Skills or References. Combined with on-demand composition, this design both helps agents bypass ineffective paths caused by version mismatch and, where the original resources support it, supplements omitted or overly general knowledge in the initial library, thereby reducing the mismatch between the resources and the actual tasks. Candidate updates are added only after independent validation in the target environment. 

To support reliable evaluation, we audit and repair ScienceBoard to create ScienceBoard-Verified. We also add a set of tasks that do not overlap with the final test tasks, use them exclusively for skill evolution, and combine the audited benchmark and the new tasks into ScienceBoard-VE; in addition, we extend the evaluation framework to support native agents. Across six scientific applications, the initial Skills and References improve the average task success rate over the base model by up to 7.23 percentage points. Execution-guided skill evolution increases the total gain over the baseline to as much as 17.12 percentage points. Our contributions are:

\begin{enumerate}
    \item \textbf{Resource-to-skill construction for scientific software.}
    We present a method for constructing a knowledge library from raw software resources. It organizes task-oriented procedural knowledge as Skills and cross-task shared software knowledge as References, and supports their on-demand composition during execution.

    \item \textbf{Execution-feedback-driven skill evolution.}
    We use failure trajectories to guide exploration and help agents avoid repeatedly following erroneous Skills or References. When knowledge is missing or an existing entry is inapplicable, the system supplements or revises Skills and References from the original resources and updates the knowledge library only after validation in the target environment.

    \item \textbf{Reliable evaluation of skill construction and evolution.}
    We audit and repair ScienceBoard to create ScienceBoard-Verified, add skill-evolution tasks disjoint from the final test set to form ScienceBoard-VE, and extend evaluation to native agents.
\end{enumerate}

\section{Related Work}

\paragraph{Resource-to-Skill Learning.}
Automatically constructing reusable agent knowledge from external resources is a relatively recent research direction. MMG2Skill~\cite{che2026mmg2skill} converts public guides into editable Skills, but assumes that tasks are paired with relevant guides in advance and does not realize an end-to-end pipeline for constructing Skills directly from raw resources. VisualSkill~\cite{jiang2026visualskill} leverages document images to support GUI element identification and state verification. However, some HTML documentation contains no images and thus provides no additional visual evidence; even when screenshots are available, their interfaces may differ from the current software version, potentially introducing misleading visual cues. SkillFoundry~\cite{shen2026skillfoundry} supports workflow generation for concrete scientific tasks, but focuses on domain capability coverage and task workflows rather than on systematically organizing reusable knowledge for a given scientific software package. Overall, existing methods differ in their resource formats, task assumptions, and construction objectives, but systematic investigation is still lacking into how to automatically discover tasks from raw scientific-software resources and organize cross-task reusable knowledge without pre-specified task–resource pairs.

\paragraph{Self-Evolving Skills and Experience-Driven Agent Adaptation.}
Skill self-evolution improves reusable agent knowledge from execution experience without modifying model parameters. Early work such as Reflexion~\cite{shinn2023reflexion} used verbal feedback to improve subsequent behavior, and SEAgent~\cite{sun2025seagent} extended this experience-driven paradigm to computer-use agents. Later work made such experience explicit and reusable as skills. EvoSkill~\cite{alzubi2026evoskill} creates or revises skills from failure feedback and selects candidates through validation, while Trace2Skill~\cite{ni2026trace2skill} distills trajectory-local lessons into transferable skill catalogs. As skill updates become central, subsequent studies increasingly focus on their reliability. SkillOpt~\cite{yang2026skillopt} constrains the scope of skill edits and accepts only validation-improving updates, while SkillCAT~\cite{chen2026skillcat} contrasts successful and failed trajectories for the same task and uses task replay before merging candidate edits. Most prior methods start from execution experience or an existing skill library. ScienceCUA instead initializes its library from scientific-software documentation and addresses version and task-coverage mismatch through failure-history-guided exploration and source-grounded additions or revisions.

\afterpage{%
\begin{figure}[!t]
    \centering
    \includegraphics[width=1.0\linewidth]{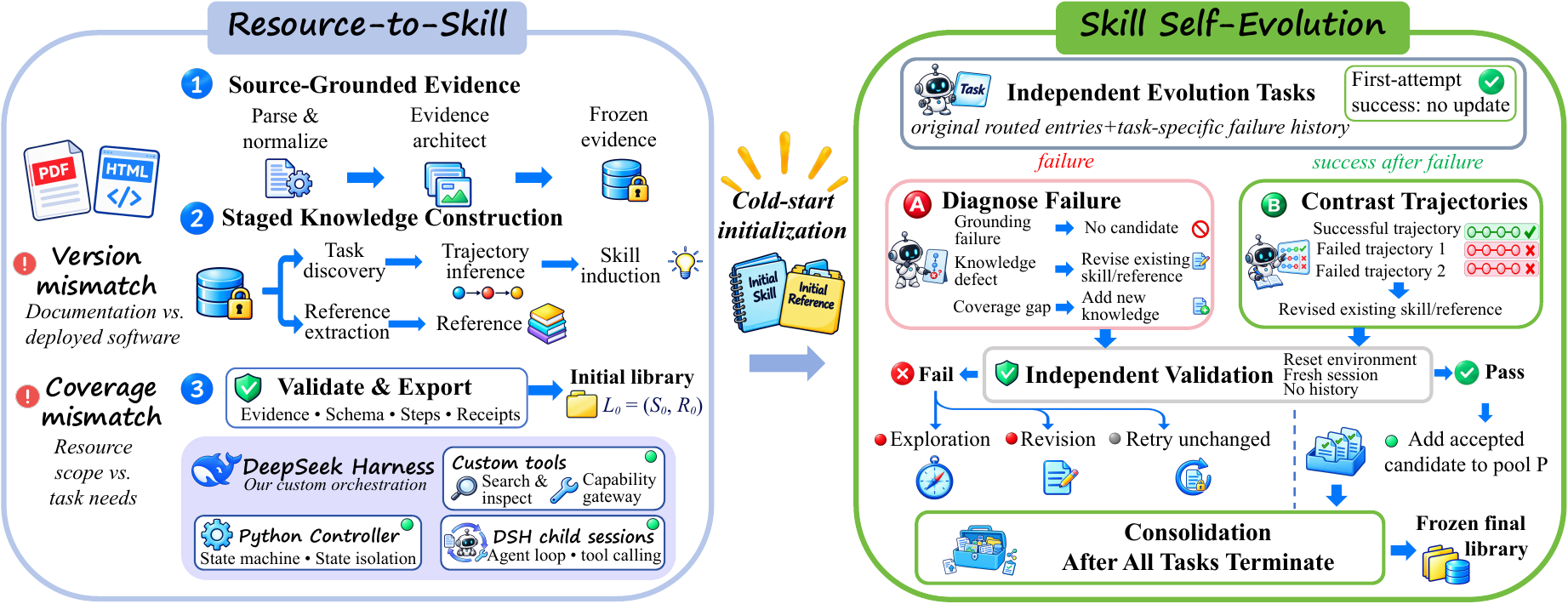}
    \caption{ScienceCUA overview. Resource-to-Skill construction builds Skills and References from PDF/HTML documentation to initialize execution-guided Skill self-evolution.}
    \label{fig:method_overview}
    \vspace{-1em}
\end{figure}
}

\section{Method}
\label{sec:method}

\subsection{Resource-to-Skill Construction}
\label{sec:resource_to_skill}

ScienceCUA first transforms software documentation into an initial library of reusable Skills and queryable References, and then evolves this library using execution feedback from a disjoint set of evolution tasks.

\subsubsection{Decoupling Skills and References}
\label{sec:decoupling_skills_references}

\begin{wrapfigure}[10]{r}{0.5\textwidth}
    \vspace{-\intextsep}
    \centering
    \includegraphics[width=\linewidth,pagebox=cropbox,clip]{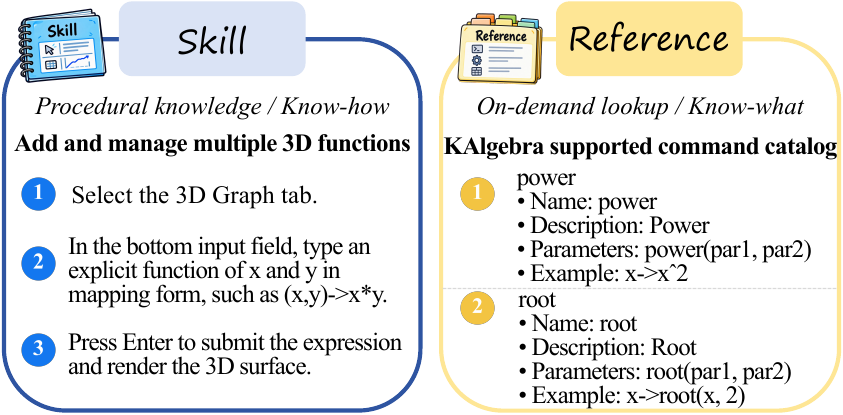}
    \captionsetup{font=small,justification=raggedright,singlelinecheck=false,position=bottom,skip=7pt}
    \caption{An example in KAlgebra: a procedural Skill for plotting 3D functions (left) and a Reference cataloging supported commands (right).}
    \label{fig:skill_reference_examples}
\end{wrapfigure}

Given PDF/HTML software documentation $\mathcal{D}$, we construct an initial library $\mathcal{L}_0=(\mathcal{S}_0,\mathcal{R}_0)$, where Skills $\mathcal{S}_0$ encode \textbf{reusable task-level operational procedures} and References $\mathcal{R}_0$ organize \textbf{cross-task shared knowledge}, such as commands, syntax, and parameter constraints. For example, KAlgebra's 2D and 3D plotting workflows form separate Skills, while syntax for square roots and exponentiation is stored in a Reference (Figure~\ref{fig:skill_reference_examples}). During execution, the agent selects a Skill for the required workflow and consults References as needed to incorporate software knowledge into its actions.

\subsubsection{Construction Pipeline}
\label{sec:construction_pipeline}

We represent a Skill as $s=(n_s,d_s,a_s,p_s,b_s)$, where $n_s$ is its name, $d_s$ is its capability description, $a_s$ specifies its applicability, $p_s$ lists its preconditions, and $b_s$ contains its procedural body, including the steps needed to accomplish the goal. We represent a Reference as $r=(n_r,d_r,c_r)$, where $n_r$ is its name, $d_r$ describes its purpose, and $c_r$ contains the concrete lookup content. Construction first organizes the documentation into a shared evidence corpus. From this corpus, one branch extracts References, while the other discovers Tasks, infers trajectories, and induces Skills.

\paragraph{Traceable evidence organization.}
To support all downstream stages and reduce hallucinations, we organize documentation into traceable evidence. A parser uses authored headings and navigation to structure text and visual content while retaining source locations. An agent refines this organization without rewriting the content. Every downstream Reference, Task, and inferred action or observation must be grounded in this evidence.

\paragraph{Reference extraction.}
To make shared software knowledge accessible during execution, the Reference branch turns relevant evidence into focused lookup entries. A program first extracts and deduplicates candidate lookup content from the documentation, including command tables, syntax constraints, and shortcut descriptions. An agent reviews the remaining candidates to decide what to retain and how to group them, preserving documented syntax, parameters, and examples.

\paragraph{Task discovery, trajectory inference, and Skill induction.}
To ground reusable Skills in concrete operational requirements, the other branch proceeds from goals to trajectories before induction. An agent first discovers source-supported user goals without benchmark task input. For each Task, it infers ordered actions and expected observations, retaining documented controls. These trajectories make the required operations explicit before related procedures are grouped into a coherent Skill, helping preserve actionable detail during abstraction. After induction, we check whether the operational capability expressed by each trajectory is included in a Skill and explicitly record any exclusions.

\newpage

Each stage produces structured outputs that serve as inputs to subsequent stages, yielding the initial library $\mathcal{L}_0=(\mathcal{S}_0,\mathcal{R}_0)$. Reference extraction, Task discovery, trajectory inference, and Skill induction are assigned to independent agents, each with its own tools, prompt, and required output format. A unified description of the Resource-to-Skill pipeline, including agent roles and tools, prompts, and output schemas, is provided in Appendix~\ref{app:design}.

\vspace{0.6\baselineskip}
\subsection{Execution-Guided Skill Self-Evolution}
\label{sec:evolution}

Differences between documented and deployed software versions, together with gaps between resource coverage and task requirements, can leave the initial library misaligned with the target environment or incomplete for its tasks. ScienceCUA therefore keeps the execution model fixed and continually evolves Skills and References through two mechanisms: \textbf{failure-history-guided exploration} and \textbf{source-grounded knowledge addition}.

\paragraph{Evolution setting and overall workflow.}
As described in Section~\ref{sec:scienceboard_evolution_tasks}, ScienceBoard-VE contains an evolution set $E=\mathrm{Expanded}\setminus\mathrm{Verified}$ and a held-out test set $V=\mathrm{Verified}$, where $E\cap V=\emptyset$. Evolution uses $E$, while $V$ remains unseen until final evaluation. Given the initial library $\mathcal{L}_0=(\mathcal{S}_0,\mathcal{R}_0)$, ScienceCUA processes each task $q\in E$ independently and maintains its history $\mathcal{H}_q$. Histories and provisional updates are not shared across tasks.

Each task uses two types of execution. \textbf{Task exploration} attempts the task with the frozen initial library, together with any prior failure records available, to search for a feasible path and collect environmental feedback. \textbf{Candidate validation} temporarily loads one modified or newly added Skill or Reference and reruns the task to test whether that candidate can independently support successful execution.

Each task begins with task exploration. A first-exploration success requires no update and exits the evolution process. After a failure, the system either makes no change, revises existing knowledge, or adds source-supported knowledge according to its diagnosis. Any modified or newly added Skill or Reference undergoes candidate validation. A successful candidate enters the candidate pool $\mathcal{P}$. A failed candidate and its validation trajectory enter $\mathcal{H}_q$ to guide the next task exploration. If a later exploration succeeds with the help of failure history, the system contrasts it with earlier failures, derives a revision candidate when needed, and validates it. Each task contributes at most one validated candidate to $\mathcal{P}$. These candidates are then consolidated into $\mathcal{L}_{\mathrm{final}}$, merged targets undergo regression certification, and the library is frozen for evaluation. Further details are provided in Appendix~\ref{app:self_evolution}.

\paragraph{Failure-history-guided exploration.}
The first exploration of task $q$ performs normal routing over $\mathcal{L}_0$ by selecting entries from their names and descriptions; the routed Skills and References define $\mathcal{F}_q\subseteq\mathcal{L}_0$. Later task explorations load the same entries in their initial versions and additionally receive the accumulated failure history. Formally, the $i$-th exploration is $\tau_q^{(i)}=\operatorname{Execute}_{\theta}(q;\mathcal{L}_0,\mathcal{F}_q,\mathcal{H}_q^{(i)})$, where $\mathcal{H}_q^{(1)}=\emptyset$. For $i>1$, $\mathcal{H}_q^{(i)}$ contains the task's earlier failed task explorations and candidate validations, including their trajectories, screenshots, and outcomes. These failure records help the model avoid paths that failed because of version mismatch, rather than mechanically follow initial knowledge that conflicts with the deployed software. Later explorations always start from the same $\mathcal{L}_0$ and $\mathcal{F}_q$: failure history changes exploration behavior, whereas knowledge changes still require independent validation. Appendix~\ref{app:evolution_history} specifies the archived evidence and its input format.

\paragraph{Failure diagnosis and candidate construction.}
For a failed task exploration, the diagnosis module analyzes its loaded Skills and References, trajectory, screenshots, and evaluator outcome, and assigns one primary cause: (1) \textbf{grounding failure}, where interface perception or manipulation is the main obstacle and no candidate is produced; (2) \textbf{modify existing knowledge}, where one entry in $\mathcal{F}_q$ is relevant but contains an incorrect or incomplete procedure, condition, parameter, or version-dependent instruction; or (3) \textbf{add new knowledge}, where the routed knowledge has a genuine coverage gap and triggers retrieval from the original resources $\mathcal{D}$. The modify branch produces one complete revision candidate $\Delta_q^{(i)}$; the add branch produces a new Skill or Reference candidate only when retrieval finds sufficient source support. Any resulting candidate is independently validated: success adds it to $\mathcal{P}$ and terminates task $q$, whereas failure records it in $\mathcal{H}_q$ before task exploration resumes. Separately, only when a later task exploration itself succeeds does the system contrast that success with earlier failures and either revise one entry in $\mathcal{F}_q$ or make no change; this branch cannot add knowledge. Details are provided in Appendices~\ref{app:evolution_diagnosis} and~\ref{app:evolution_submission}.

\paragraph{Independent candidate validation.}
For candidate $\Delta_q^{(i)}$, ScienceCUA constructs the temporary library $\widetilde{\mathcal{L}}_q^{(i)}=\mathcal{L}_0\oplus\Delta_q^{(i)}$, where $\oplus$ replaces one initial entry with its revision or inserts one source-supported new entry. Validation then executes $\widetilde{\tau}_q^{(i)}=\operatorname{Execute}_{\theta}(q;\widetilde{\mathcal{L}}_q^{(i)},\widetilde{\mathcal{F}}_q^{(i)},\emptyset)$ in a reset environment and a fresh session. It fixes the route obtained by the first task exploration, with the candidate entry substituted or added, and receives no prior trajectory, diagnosis, or candidate rationale, thereby testing whether the candidate knowledge itself supports the task. A successful candidate enters $\mathcal{P}$; a failed candidate remains inactive, and its execution record is added to the task's failure history. Appendix~\ref{app:evolution_validation} gives the complete validation protocol.

\section{ScienceBoard-VE: A Verified and Expanded Benchmark for Scientific Computer-Use Agents}
\label{sec:scienceboard_ve}

\begin{wrapfigure}{r}{0.5\textwidth}
    \vspace{-\intextsep}
    \centering
    \includegraphics[width=\linewidth,pagebox=cropbox,clip]{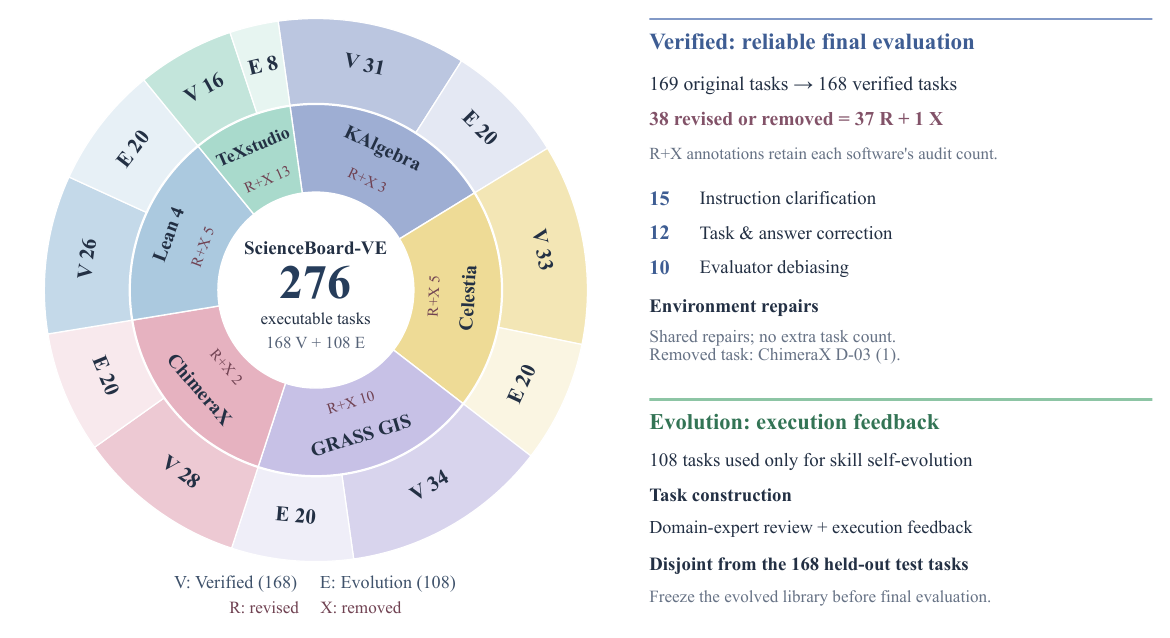}
    \captionsetup{font=small,justification=raggedright,singlelinecheck=false,position=bottom,skip=7pt}
    \caption{Task distribution in ScienceBoard-VE. The inner ring groups tasks by application. The outer ring separates verified evaluation tasks (V) from evolution tasks (E). R+X denotes revised or removed tasks in the original benchmark. The removed task is excluded from the sector sizes.}
    \label{fig:scienceboard_ve}
\end{wrapfigure}

ScienceBoard~\citep{sun2026scienceboard} is a benchmark for scientific computer-use agents and has been adopted in subsequent studies on agent evaluation, training, and task expansion~\citep{lv2026scalecua,lu2026qwen}. During reproduction, we audited all 169 original tasks and identified issues in task instructions, task statements, reference answers, evaluators, and execution environments. We therefore construct \textbf{ScienceBoard-Verified} and add a disjoint set of tasks for skill self-evolution. Together, they form \textbf{ScienceBoard-VE}. We also extend the evaluation framework to support native-agent evaluation. Figure~\ref{fig:scienceboard_ve} summarizes the task distribution and benchmark quality assurance.

\subsection{ScienceBoard-Verified}
\label{sec:scienceboard_verified}

We substantively revise 37 tasks and remove one additional task, yielding 168 executable tasks reserved for final evaluation. We organize the revisions and supporting repairs into four categories: (1) \textbf{Instruction Ambiguity}, which covers 15 revised tasks with unclear goals or omitted completion requirements; (2) \textbf{Task \& Answer Errors}, which covers 12 revised tasks involving incorrect task statements, reference answers, or expected evaluator results; (3) \textbf{Evaluator Bias}, which covers 10 revised tasks in which unnecessary restrictions exclude valid execution paths, output representations, or output locations; and (4) \textbf{Environment Repairs}, which covers application prompts, response handling, runtime and initialization settings, and unavailable external dependencies. The fourth category includes shared evaluation-environment repairs and the removed task; these do not add to the 37 revised tasks. Appendix~\ref{app:scienceboard_verified_repairs} details all 37 task revisions, the task removal rationale, and the associated evaluation-environment repairs.

\subsection{Expanding the Benchmark for Skill Self-Evolution}
\label{sec:scienceboard_evolution_tasks}

To separate skill improvement from final evaluation, we add 108 tasks that do not overlap with the verified test set: 20 each for KAlgebra, Celestia, GRASS GIS, ChimeraX, and Lean 4, and 8 for TeXstudio. These tasks are used exclusively for \textbf{skill self-evolution}, while the 168 tasks in ScienceBoard-Verified remain unseen during skill construction and evolution and are used only for final evaluation. The resulting \textbf{ScienceBoard-VE} contains 276 executable tasks in total. The evolution tasks use the same types of software operations as the test tasks but differ in their specific targets or requirements, allowing execution experience to transfer without reusing the test tasks themselves. See Appendix~\ref{app:task_deduplication} for details.

\subsection{Native-Agent Evaluation}
\label{sec:native_agent_evaluation}

The original ScienceBoard evaluates models through a unified API-based interaction loop. We extend it to support native evaluation of Codex and Grok Build, keeping tasks, software environments, and evaluators unchanged. In both settings, ScienceBoard supplies screenshots and accessibility trees, executes the returned action code or control instructions, and evaluates task outcomes. In the model-based setting, each API request includes the system prompt, task instruction, screenshots and accessibility trees from the three most recent steps together with the model-generated actions for those steps, and the current screenshot and accessibility tree. For ScienceCUA, we implement two-stage resource routing through progressive disclosure. Before execution, the evaluated model selects relevant Skills and References from their names and descriptions based on the task instruction. The selected entries are then loaded in full into the system prompt and included in every subsequent execution request. In the native-agent setting, each task uses a separate persistent session, to which the outer evaluation framework appends only the current screenshot and accessibility tree at each step. The native harness maintains the task history and compacts context as needed. Skills and References are registered as independently loadable native resources, which the agent selects and reads during execution. Each turn concludes with one ScienceBoard action block for the evaluation framework to execute. To prevent agents from obtaining additional reference material through web search, we disable web search for native agents. To preserve task independence, conversation history and runtime context are maintained separately for each task and are not shared between tasks.

\section{Experiments}

\paragraph{Experimental setup.}
We evaluate ScienceCUA on ScienceBoard-Verified. Model-based evaluation uses Qwen-3.8 Max, GLM-5.2, Grok-4.6, and GPT-5.5. Native-agent evaluation uses Grok Build with Grok-4.6 and Codex with GPT-5.5. The two settings follow the respective interaction and resource access protocols described in Section~\ref{sec:native_agent_evaluation}. In model-based ScienceCUA evaluation, at most three Skill and Reference entries are selected in total before each task. In native-agent evaluation, the agent autonomously selects and reads resources during execution, with no limit on the number of Skill and Reference entries it can access. We report task success rates for each of the six applications and use their unweighted mean as the overall performance metric.

Model-based evaluation compares ScienceCUA with Vanilla, direct document injection, RAG, VisualSkill, and SkillFoundry. Native-agent evaluation compares Vanilla, VisualSkill, SkillFoundry, and ScienceCUA. ScienceCUA and VisualSkill use the same documentation inputs. SkillFoundry does not support direct documentation input and instead automatically searches for and collects online resources based on the name and version of the software installed in the execution environment. Implementation details are provided in Appendix~\ref{app:direct-injection-details} for direct injection, Appendix~\ref{app:rag-details} for RAG, Appendix~\ref{app:skillfoundry-reproduction} for SkillFoundry, and Appendix~\ref{app:visualskill-reproduction} for VisualSkill.

Skill self-evolution uses the 108 additional tasks in ScienceBoard-VE. After evolution, the knowledge library is frozen and evaluated on ScienceBoard-Verified without access to evolution trajectories or histories.

\newsavebox{\mainresultstablebox}
\begin{table*}[t]
    \centering
    \begin{lrbox}{\mainresultstablebox}%
    \begingroup%
    \renewcommand{\resultgain}[1]{\raisebox{-0.4pt}{\fontsize{5.5}{6}\selectfont\textcolor{resultgain}{(#1)}}}%
    \renewcommand{\resultloss}[1]{\raisebox{-0.4pt}{\fontsize{5.5}{6}\selectfont\textcolor{resultloss}{(#1)}}}%
    \resulttablefont%
    \fontsize{6.5}{7.8}\selectfont%
    \setlength{\tabcolsep}{6.8pt}%
    \renewcommand{\arraystretch}{1.13}%
    \setlength{\aboverulesep}{1.5pt}%
    \setlength{\belowrulesep}{1.5pt}%
    \begin{tabular}{ll@{\hspace{4pt}}*{6}{c}c@{\hspace{4pt}}l}%
        \toprule[0.9pt]
        \textbf{Model} & \textbf{Method} & {\fontsize{6.2}{7.2}\selectfont\bfseries KAlgebra} & {\fontsize{6.2}{7.2}\selectfont\bfseries Celestia} & {\fontsize{6.2}{7.2}\selectfont\bfseries GRASS GIS} & {\fontsize{6.2}{7.2}\selectfont\bfseries ChimeraX} & {\fontsize{6.2}{7.2}\selectfont\bfseries Lean 4} & {\fontsize{6.2}{7.2}\selectfont\bfseries TeXstudio} & \textbf{Average} & \\
        \midrule[0.5pt]
         & Vanilla & 32.26 & 33.33 & 29.41 & 57.14 & 30.77 & 50.00 & 38.82 &  \\
         & RAG & 38.71 & 36.36 & \bfseries 32.35 & 53.57 & 30.77 & 50.00 & 40.29 & \resultgain{$\uparrow$ 3.79\%} \\
         & Direct injection & \bfseries 51.61 & \bfseries 45.45 & 26.47 & \bfseries 64.29 & 26.92 & 37.50 & 42.04 & \resultgain{$\uparrow$ 8.29\%} \\
        \resultmodel{Qwen-3.8 Max\\(xhigh)} & visualskill & 38.71 & 33.33 & \bfseries 32.35 & 57.14 & 30.77 & 50.00 & 40.38 & \resultgain{$\uparrow$ 4.02\%} \\
         & skillfoundry & 35.48 & 36.36 & 20.59 & 53.57 & \bfseries 34.62 & 50.00 & 38.44 & \resultloss{$\downarrow$ 0.98\%} \\
        \rowcolor{oursbackground}
        \cellcolor{white} & ScienceCUA & \bfseries 51.61 & 36.36 & \bfseries 32.35 & \bfseries 64.29 & 30.77 & \bfseries 56.25 & \bfseries 45.27 & \resultgain{$\uparrow$ 16.62\%} \\
        \midrule[0.35pt]
         & Vanilla & 19.35 & 12.12 & 0.00 & 53.57 & 3.85 & 6.25 & 15.86 &  \\
         & RAG & 22.58 & \bfseries 18.18 & 0.00 & 60.71 & 3.85 & \bfseries 12.50 & 19.64 & \resultgain{$\uparrow$ 23.83\%} \\
         & Direct injection & 25.81 & 15.15 & 0.00 & 57.14 & 0.00 & 6.25 & 17.39 & \resultgain{$\uparrow$ 9.65\%} \\
        \resultmodel{GLM-5.2\\(Max)} & visualskill & \bfseries 29.03 & 9.09 & 0.00 & \bfseries 64.29 & \bfseries 7.69 & 6.25 & 19.39 & \resultgain{$\uparrow$ 22.26\%} \\
         & skillfoundry & 19.35 & 12.12 & \bfseries 2.94 & 50.00 & 3.85 & \bfseries 12.50 & 16.79 & \resultgain{$\uparrow$ 5.86\%} \\
        \rowcolor{oursbackground}
        \cellcolor{white} & ScienceCUA & \bfseries 29.03 & 15.15 & \bfseries 2.94 & \bfseries 64.29 & \bfseries 7.69 & \bfseries 12.50 & \bfseries 21.93 & \resultgain{$\uparrow$ 38.27\%} \\
        \midrule[0.35pt]
         & Vanilla & 61.29 & 36.36 & 14.71 & 82.14 & \bfseries 11.54 & 18.75 & 37.47 &  \\
         & RAG & 70.97 & 36.36 & 20.59 & 82.14 & 7.69 & \bfseries 31.25 & 41.50 & \resultgain{$\uparrow$ 10.76\%} \\
         & Direct injection & \bfseries 74.19 & 36.36 & 23.53 & 78.57 & 7.69 & 25.00 & 40.89 & \resultgain{$\uparrow$ 9.13\%} \\
        \resultmodel{Grok-4.6\\(xhigh)} & visualskill & 61.29 & 36.36 & 23.53 & 78.57 & \bfseries 11.54 & \bfseries 31.25 & 40.42 & \resultgain{$\uparrow$ 7.87\%} \\
         & skillfoundry & 45.16 & 33.33 & 20.59 & \bfseries 85.71 & \bfseries 11.54 & \bfseries 31.25 & 37.93 & \resultgain{$\uparrow$ 1.23\%} \\
        \rowcolor{oursbackground}
        \cellcolor{white} & ScienceCUA & 70.97 & \bfseries 48.48 & \bfseries 26.47 & \bfseries 85.71 & \bfseries 11.54 & 25.00 & \bfseries 44.70 & \resultgain{$\uparrow$ 19.30\%} \\
        \midrule[0.35pt]
         & Vanilla & 58.06 & 45.45 & 52.94 & 75.00 & 26.92 & \bfseries 50.00 & 51.40 &  \\
         & RAG & 64.52 & 54.55 & 50.00 & \bfseries 78.57 & \bfseries 30.77 & 50.00 & 54.73 & \resultgain{$\uparrow$ 6.48\%} \\
         & Direct injection & 67.74 & 54.55 & \bfseries 55.88 & 75.00 & 23.08 & 43.75 & 53.33 & \resultgain{$\uparrow$ 3.75\%} \\
        \resultmodel{GPT-5.5\\(xhigh)} & visualskill & 64.52 & 57.58 & 52.94 & 75.00 & 26.92 & \bfseries 50.00 & 54.49 & \resultgain{$\uparrow$ 6.01\%} \\
         & skillfoundry & 61.29 & 54.55 & 50.00 & \bfseries 78.57 & 26.92 & \bfseries 50.00 & 53.56 & \resultgain{$\uparrow$ 4.20\%} \\
        \rowcolor{oursbackground}
        \cellcolor{white} & ScienceCUA & \bfseries 70.97 & \bfseries 60.61 & \bfseries 55.88 & \bfseries 78.57 & 26.92 & \bfseries 50.00 & \bfseries 57.16 & \resultgain{$\uparrow$ 11.21\%} \\
        \midrule[0.9pt]
        \textbf{Agent} & \textbf{Method} & {\fontsize{6.2}{7.2}\selectfont\bfseries KAlgebra} & {\fontsize{6.2}{7.2}\selectfont\bfseries Celestia} & {\fontsize{6.2}{7.2}\selectfont\bfseries GRASS GIS} & {\fontsize{6.2}{7.2}\selectfont\bfseries ChimeraX} & {\fontsize{6.2}{7.2}\selectfont\bfseries Lean 4} & {\fontsize{6.2}{7.2}\selectfont\bfseries TeXstudio} & \textbf{Average} & \\
        \midrule[0.5pt]
         & Vanilla & 70.97 & 42.42 & 35.29 & 78.57 & 19.23 & 31.25 & 46.29 &  \\
        \resultmodel{Grok Build\\Grok-4.6\\(xhigh)} & visualskill & 70.97 & 63.64 & 35.29 & 75.00 & 26.92 & 31.25 & 50.51 & \resultgain{$\uparrow$ 9.12\%} \\
         & skillfoundry & 64.52 & 51.52 & 26.47 & 78.57 & \bfseries 30.77 & \bfseries 43.75 & 49.27 & \resultgain{$\uparrow$ 6.44\%} \\
        \rowcolor{oursbackground}
        \cellcolor{white} & ScienceCUA & \bfseries 77.42 & \bfseries 66.67 & \bfseries 38.24 & \bfseries 85.71 & 26.92 & 37.50 & \bfseries 55.41 & \resultgain{$\uparrow$ 19.70\%} \\
        \midrule[0.35pt]
         & Vanilla & 58.06 & 51.52 & 52.94 & 75.00 & \bfseries 30.77 & \bfseries 56.25 & 54.09 &  \\
        \resultmodel{GPT-5.5\\(xhigh)+Codex} & visualskill & 61.29 & 57.58 & 55.88 & \bfseries 78.57 & 26.92 & \bfseries 56.25 & 56.08 & \resultgain{$\uparrow$ 3.68\%} \\
         & skillfoundry & 61.29 & 54.55 & 55.88 & 75.00 & \bfseries 30.77 & 50.00 & 54.58 & \resultgain{$\uparrow$ 0.91\%} \\
        \rowcolor{oursbackground}
        \cellcolor{white} & ScienceCUA & \bfseries 70.97 & \bfseries 60.61 & \bfseries 61.76 & \bfseries 78.57 & \bfseries 30.77 & \bfseries 56.25 & \bfseries 59.82 & \resultgain{$\uparrow$ 10.59\%} \\
        \bottomrule[0.9pt]
    \end{tabular}%
    \endgroup%
    \end{lrbox}
    \begin{minipage}{\textwidth}
    \caption{Success rates on ScienceBoard-Verified under model-based (top) and native-agent (bottom) evaluation. Bold marks column-wise best results within each group.}
    \label{tab:main_results}
    \label{tab:native_agent_results}
    \noindent\resizebox{\linewidth}{!}{\usebox{\mainresultstablebox}}
    \end{minipage}
    \vspace{-0.6em}
\end{table*}

\begin{figure*}[t]
    \centering
    \includegraphics[width=\textwidth]{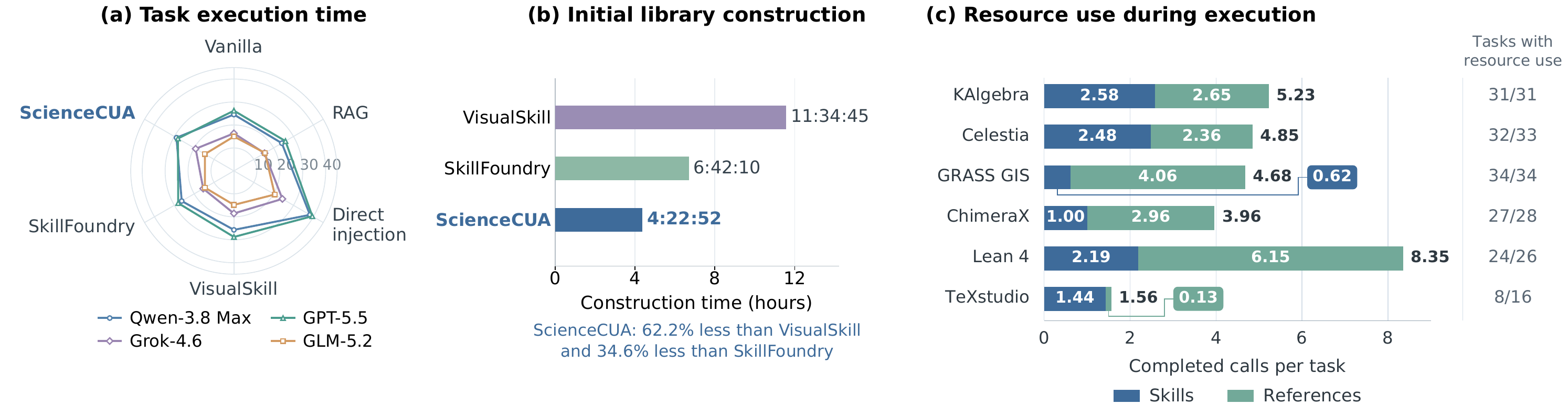}
    \caption{Task execution time, initial library construction cost, and resource use during execution. (a) Total task execution time summed across all six applications for four models under six methods. Each curve represents one model, and radial values indicate hours. (b) Total initial library construction time across the six applications for VisualSkill, SkillFoundry, and ScienceCUA, using GPT-5.5 (xhigh). (c) Completed calls to Skills and References in final task attempts, averaged over all tasks in each application, for native Grok Build with Grok-4.6 (xhigh). White labels give Skill and Reference calls per task in their respective colors; the narrow GRASS GIS Skill and TeXstudio Reference segments have connected labels. Bar-end labels give the combined calls per task. Values are rounded independently. The fractions on the right show the number of tasks with at least one resource call over the total number of tasks in each application.}
    \label{fig:resource_cost_and_usage}
\end{figure*}

\rqsubsection{RQ1: Can ScienceCUA-constructed resources improve task success in scientific software operation?}
\label{sec:rq1}

Table~\ref{tab:main_results} shows that the initial ScienceCUA library achieves the highest average success rate in all four model-based settings. ScienceCUA improves over Vanilla by 5.76--7.23 percentage points, compared with 1.53--3.42 points for direct document injection, and outperforms RAG, VisualSkill, and SkillFoundry across all four models. The gains also extend to native-agent evaluation. ScienceCUA reaches average success rates of 55.41\% with Grok Build and 59.82\% with Codex, improving over their respective Vanilla baselines by 9.12 and 5.73 percentage points and outperforming VisualSkill and SkillFoundry in both settings. Figure~\ref{fig:resource_cost_and_usage}(a) shows that ScienceCUA requires 21.1--28.9\% less total task execution time than direct document injection across the four models. ScienceCUA also reduces total initial library construction time by 62.2\% and 34.6\% relative to VisualSkill and SkillFoundry, respectively (Figure~\ref{fig:resource_cost_and_usage}(b)). These results support the usefulness of the constructed Skills and References as initial operational knowledge for scientific software.

\rqsubsection{RQ2: Do Skills and References make complementary contributions to task success?}
\label{sec:rq2}

We ablate Skills and References separately to assess the individual and joint contributions of procedural guidance and cross-task shared lookup knowledge to task success. Skills organize complete operational procedures around specific goals, while References provide shared knowledge that is not tied to any single procedure, including syntax constraints, parameter rules, command and function catalogs, and keyboard shortcuts. The two resource types have no one-to-one correspondence that permits direct, equivalent merging. We therefore do not use a forced merger of the two resource types as an information-matched control for knowledge organization, and instead use resource ablations to evaluate the value of providing each knowledge type individually and both jointly.

Table~\ref{tab:skill_reference_ablation} shows that providing either resource type alone improves average success rates over Vanilla, while providing both achieves the best results. In model-based evaluation with Grok-4.6, the average success rates are 40.09\% with Skills alone, 40.62\% with References alone, and 44.70\% with both. The corresponding results for native Grok Build are 49.89\%, 51.14\%, and 55.41\%. The relative gains from the two resource types vary across applications, while providing both further improves overall performance, supporting the complementary contributions of procedural guidance and shared lookup knowledge to task execution.

The KAlgebra C-05 comparison in Appendix~\ref{app:case-kalgebra-c05} illustrates these roles under the same budget of five interaction turns. Each condition retains autonomous routing over the permitted resource types. Reference-only spends its budget locating the script-loading interface; Skill-only loads the script but leaves the calculation symbolic. With both types, the agent combines the script-loading Skill's guidance with application-specific knowledge from the function-catalog Reference, then corrects an intermediate vector-valued result using execution feedback to complete the task.

Figure~\ref{fig:resource_cost_and_usage}(c) further illustrates actual calls to both resource types during task execution. Native Grok Build with Grok-4.6 (xhigh) accesses at least one resource in 156 of the 168 tasks (92.9\%). Final task attempts contain 291 completed Skill calls and 543 completed Reference calls, with repeated calls to the same entry counted separately. These records show that the agent calls both Skills and References during actual task execution and, together with the preceding ablation results, support the contributions of both knowledge types to task completion.

\newsavebox{\ablationtablebox}
\begin{table*}[t]
    \centering
    \begin{lrbox}{\ablationtablebox}%
    \begingroup%
    \renewcommand{\resultgain}[1]{\raisebox{-1.2pt}{\fontsize{5.5}{6}\selectfont\textcolor{resultgain}{(#1)}}}%
    \renewcommand{\resultloss}[1]{\raisebox{-1.2pt}{\fontsize{5.5}{6}\selectfont\textcolor{resultloss}{(#1)}}}%
    \resulttablefont%
    \fontsize{7}{8.5}\selectfont%
    \setlength{\tabcolsep}{7pt}%
    \renewcommand{\arraystretch}{0.95}%
    \setlength{\aboverulesep}{1.2pt}%
    \setlength{\belowrulesep}{1.2pt}%
    \begin{tabular}{ll@{\hspace{6pt}}*{6}{c}c@{\hspace{3pt}}l}%
        \toprule[0.9pt]
        \textbf{Model} & \textbf{Method} & {\fontsize{6.5}{7.5}\selectfont\textbf{KAlgebra}} & {\fontsize{6.5}{7.5}\selectfont\textbf{Celestia}} & {\fontsize{6.5}{7.5}\selectfont\textbf{GRASS GIS}} & {\fontsize{6.5}{7.5}\selectfont\textbf{ChimeraX}} & {\fontsize{6.5}{7.5}\selectfont\textbf{Lean 4}} & {\fontsize{6.5}{7.5}\selectfont\textbf{TeXstudio}} & \textbf{Average} & \\
        \midrule[0.5pt]
         & Vanilla & 61.29 & 36.36 & 14.71 & 82.14 & \textbf{11.54} & 18.75 & 37.47 &  \\
        \resultmodel{Grok-4.6\\(xhigh)} & skill-only & 67.74 & 39.39 & 14.71 & 82.14 & \textbf{11.54} & \textbf{25.00} & 40.09 & \resultgain{$\uparrow$ 6.99\%} \\
         & reference-only & 64.52 & 36.36 & 20.59 & \textbf{85.71} & \textbf{11.54} & \textbf{25.00} & 40.62 & \resultgain{$\uparrow$ 8.41\%} \\
        \rowcolor{oursbackground}
        \cellcolor{white} & skill+reference\hspace{1pt} & \textbf{70.97} & \textbf{48.48} & \textbf{26.47} & \textbf{85.71} & \textbf{11.54} & \textbf{25.00} & \textbf{44.70} & \resultgain{$\uparrow$ 19.30\%} \\
        \midrule[0.35pt]
         & Vanilla & 70.97 & 42.42 & 35.29 & 78.57 & 19.23 & 31.25 & 46.29 &  \\
        \resultmodel{Grok Build\\Grok-4.6\\(xhigh)} & skill-only & 74.19 & 54.55 & 35.29 & 78.57 & 19.23 & \textbf{37.50} & 49.89 & \resultgain{$\uparrow$ 7.78\%} \\
         & reference-only & \textbf{77.42} & 48.48 & \textbf{38.24} & 82.14 & 23.08 & \textbf{37.50} & 51.14 & \resultgain{$\uparrow$ 10.48\%} \\
        \rowcolor{oursbackground}
        \cellcolor{white} & skill+reference\hspace{1pt} & \textbf{77.42} & \textbf{66.67} & \textbf{38.24} & \textbf{85.71} & \textbf{26.92} & \textbf{37.50} & \textbf{55.41} & \resultgain{$\uparrow$ 19.70\%} \\
        \bottomrule[0.9pt]
    \end{tabular}%
    \endgroup%
    \end{lrbox}
    \begin{minipage}{\textwidth}
    \caption{Ablation of skills and references on ScienceBoard-Verified.}
    \label{tab:skill_reference_ablation}
    \noindent\resizebox{\linewidth}{!}{\usebox{\ablationtablebox}}
    \end{minipage}
\end{table*}

\rqsubsection{RQ3: Can self-evolution mitigate the limitations of Resource-to-Skill construction?}
\label{sec:rq3}

The initial library may lack task-relevant knowledge or contain instructions incompatible with the deployed software version. We use Grok-4.6 (xhigh) and GPT-5.5 (xhigh) to compare the initial library, Full ScienceCUA, two ablated variants, Trace2Skill~\citep{ni2026trace2skill}, and SkillCAT~\citep{chen2026skillcat}, examining whether execution-guided self-evolution improves performance on unseen tasks. The \textbf{$-$Knowledge addition} variant only revises existing entries without adding new Skills or References. The \textbf{$-$Failure history} variant withholds prior exploration and candidate-validation records from subsequent exploration, leaving the other mechanisms unchanged. Within each model setting, all evolution variants use the same initial library, evolution tasks, software environments, and runtime configurations. Detailed ScienceCUA settings are provided in Appendix~\ref{app:self_evolution}.

Table~\ref{tab:self_evolution_ablation} shows that self-evolution improves the average success rate from 44.70\% to 54.59\% for Grok-4.6 and from 57.16\% to 65.81\% for GPT-5.5, yielding gains of 9.89 and 8.65 percentage points, respectively. Full ScienceCUA also outperforms Trace2Skill and SkillCAT, which achieve average success rates of 46.44\% and 53.02\% with Grok-4.6, and 60.46\% and 63.74\% with GPT-5.5, respectively. These results indicate that knowledge improvements obtained during evolution transfer to final test tasks not used during evolution.

Removing either knowledge addition or failure history reduces average success rates, but the dominant source of improvement varies by application. KAlgebra and Celestia are particularly sensitive to failure history: removing it reduces success by 9.68 percentage points for KAlgebra under both models, and by 9.09 and 6.06 points for Celestia under Grok-4.6 and GPT-5.5, respectively. Documentation-version mismatches leave some initial entries with inapplicable capability constraints or interface paths. Appendix~\ref{app:case-self-evolve-kalgebra} traces how execution feedback repairs such a constraint in KAlgebra, while Appendix~\ref{app:case-celestia-a05} shows how outdated interface guidance consumes execution steps in an initial Celestia run. GRASS GIS benefits especially from knowledge addition: removing it reduces success by 14.71 and 5.89 points under the two models. Its broad documentation leaves task-relevant module options underspecified in the initial library; Appendix~\ref{app:case-grass-reference} shows how adding a module-level Reference fills this gap and supports related Verified tasks. Gains are smaller or less consistent for the other applications. Lean 4 tasks require specialized formal proofs that the general theorem-proving guide rarely supplies directly, while many TeXstudio tasks require information from other software beyond the TeXstudio manual. Appendix~\ref{app:lean-texstudio-scope} compares these task requirements with the source documentation.

\begin{table*}[t]
    \centering
    \caption{Self-evolution comparison and ablation on ScienceBoard-Verified. Results are success rates in percent. All variants are evaluated with frozen libraries and no evolution history.}
        \label{tab:self_evolution_ablation}
    \begingroup
    \resulttablefont
    \fontsize{7.2}{8.6}\selectfont
    \setlength{\tabcolsep}{2.4pt}
    \renewcommand{\arraystretch}{1.03}
    \newcommand{\ablationdelta}[1]{\textcolor{resultloss}{($\downarrow$ #1\%)}}%
    \begin{tabular*}{\linewidth}{@{\extracolsep{\fill}}ll*{6}{c}c@{\hspace{3pt}}l@{}}
        \toprule[0.9pt]
        \textbf{Model} & \textbf{Method} & \textbf{KAlgebra} & \textbf{Celestia} & \textbf{GRASS GIS} & \textbf{ChimeraX} & \textbf{Lean 4} & \textbf{TeXstudio} & \textbf{Average} & \\
        \midrule[0.5pt]
         & Full ScienceCUA & 80.65\% & 63.64\% & 47.06\% & 85.71\% & 19.23\% & 31.25\% & \bfseries 54.59\% & \\
         & $-$Knowledge addition & 80.65\% & 63.64\% & 32.35\% & 85.71\% & 15.38\% & 31.25\% & 51.50\% & \ablationdelta{5.66} \\
        \resultmodel{Grok-4.6\\(xhigh)} & $-$Failure history & 70.97\% & 54.55\% & 41.18\% & 82.14\% & 15.38\% & 25.00\% & 48.20\% & \ablationdelta{11.71} \\
         & initial library & 70.97\% & 48.48\% & 26.47\% & 85.71\% & 11.54\% & 25.00\% & 44.70\% & \ablationdelta{18.12} \\
         & Trace2Skill & 70.97\% & 51.52\% & 23.53\% & 82.14\% & 19.23\% & 31.25\% & 46.44\% & \\
         & SkillCAT & 74.19\% & 63.64\% & 44.12\% & 85.71\% & 19.23\% & 31.25\% & 53.02\% & \\
        \midrule[0.35pt]
         & Full ScienceCUA & 83.87\% & 69.70\% & 64.71\% & 85.71\% & 34.62\% & 56.25\% & \bfseries 65.81\% & \\
         & $-$Knowledge addition & 80.65\% & 69.70\% & 58.82\% & 82.14\% & 30.77\% & 56.25\% & 63.05\% & \ablationdelta{4.19} \\
        \resultmodel{GPT-5.5\\(xhigh)} & $-$Failure history & 74.19\% & 63.64\% & 61.76\% & 82.14\% & 30.77\% & 50.00\% & 60.42\% & \ablationdelta{8.19} \\
         & initial library & 70.97\% & 60.61\% & 55.88\% & 78.57\% & 26.92\% & 50.00\% & 57.16\% & \ablationdelta{13.14} \\
         & Trace2Skill & 74.19\% & 60.61\% & 58.82\% & 82.14\% & 30.77\% & 56.25\% & 60.46\% & \\
         & SkillCAT & 77.42\% & 66.67\% & 61.76\% & 85.71\% & 34.62\% & 56.25\% & 63.74\% & \\
        \bottomrule[0.9pt]
    \end{tabular*}
    \endgroup
    \vspace{-1em}
\end{table*}

\section{Conclusion}

We present ScienceCUA, a two-stage framework for computer-use agents operating scientific software. The first stage extracts source-grounded operational knowledge from software documentation, distilling task-level workflows into executable Skills and organizing commands, syntax, and parameter constraints as queryable References. The second stage refines Skills and References using execution feedback from the target software environment. We also construct the audited ScienceBoard-Verified and add evolution tasks disjoint from the test set to form ScienceBoard-VE, supporting knowledge evolution and native-agent evaluation. Experiments across six scientific applications show that the initial Skills and References improve average task success by up to 7.23 percentage points over the baseline in model-based evaluation. Knowledge self-evolution increases the total gain to as much as 17.12 percentage points. These results suggest that human-authored software resources can be transformed into executable, verifiable operational knowledge that continually improves through environmental feedback, enhancing general-purpose agents' ability to use scientific software. By maintaining Skills and References outside the model parameters, ScienceCUA offers a practical path for agents to adapt to changes in scientific software environments and task requirements without retraining the model.


\clearpage
\bibliography{references}

\begin{thebibliography}{23}
\providecommand{\natexlab}[1]{#1}
\providecommand{\url}[1]{\texttt{#1}}
\expandafter\ifx\csname urlstyle\endcsname\relax
  \providecommand{\doi}[1]{doi: #1}\else
  \providecommand{\doi}{doi: \begingroup \urlstyle{rm}\Url}\fi

\bibitem[Ai et~al.(2026)Ai, Miao, Tang, Gorski, Sun, Liu, Ingolfsson, Lenz,
  Guo, Yu, Leburu, Molash, Wang, Peterka, Wang, and
  Liu]{ai2026scivisagentbench}
Kuangshi Ai, Haichao Miao, Kaiyuan Tang, Nathaniel Gorski, Jianxin Sun, Guoxi
  Liu, Helgi~I. Ingolfsson, David Lenz, Hanqi Guo, Hongfeng Yu, Teja Leburu,
  Michael Molash, Bei Wang, Tom Peterka, Chaoli Wang, and Shusen Liu.
\newblock {SciVisAgentBench: A Benchmark for Evaluating Scientific Data
  Analysis and Visualization Agents}.
\newblock \emph{arXiv preprint arXiv:2603.29139}, 2026.
\newblock URL \url{https://arxiv.org/abs/2603.29139}.

\bibitem[Alzubi et~al.(2026)Alzubi, Provenzano, Bingham, Chen, and
  Vu]{alzubi2026evoskill}
Salaheddin Alzubi, Noah Provenzano, Jaydon Bingham, Weiyuan Chen, and Tu~Vu.
\newblock Evoskill: Automated skill discovery for multi-agent systems.
\newblock \emph{arXiv preprint arXiv:2603.02766}, 2026.

\bibitem[Bonatti et~al.(2024)Bonatti, Zhao, Bonacci, Dupont, Abdali, Li, Lu,
  Wagle, Koishida, Bucker, Jang, and Hui]{bonatti2024windowsagentarena}
Rogerio Bonatti, Dan Zhao, Francesco Bonacci, Dillon Dupont, Sara Abdali,
  Yinheng Li, Yadong Lu, Justin Wagle, Kazuhito Koishida, Arthur Bucker,
  Lawrence Jang, and Zack Hui.
\newblock {Windows Agent Arena: Evaluating Multi-Modal OS Agents at Scale}.
\newblock \emph{arXiv preprint arXiv:2409.08264}, 2024.
\newblock URL \url{https://arxiv.org/abs/2409.08264}.

\bibitem[Che et~al.(2026)Che, Xiong, Ge, Lei, Li, Yan, Li, Zhang, Bai, Hao,
  Sun, Li, and Liu]{che2026mmg2skill}
Xinyu Che, Junqi Xiong, Yunfei Ge, Xinping Lei, Shihao Li, Hang Yan, Han Li,
  Yuanxing Zhang, Zhiqi Bai, Jinhua Hao, Ming Sun, Han Li, and Jiaheng Liu.
\newblock {MMG2Skill: Can Agents Distill In-the-Wild Guides into Self-Evolving
  Skills?}
\newblock \emph{arXiv preprint arXiv:2606.01993}, 2026.
\newblock URL \url{https://arxiv.org/abs/2606.01993}.

\bibitem[Chen et~al.(2026)Chen, Zhong, Liu, and Du]{chen2026skillcat}
Kunfeng Chen, Qihuang Zhong, Juhua Liu, and Bo~Du.
\newblock {SkillCAT: Contrastive Assessment and Topology-Aware Skill
  Self-Evolution for LLM Agents}.
\newblock \emph{arXiv preprint arXiv:2606.13317}, 2026.
\newblock URL \url{https://arxiv.org/abs/2606.13317v1}.

\bibitem[Fan et~al.(2026)Fan, Di, Wen, Yang, Cheng, Dai, Liu, Qiu, Dong, Li,
  and Luo]{fan2026resource2skill}
Yijia Fan, Zonglin Di, Zimo Wen, Yifan Yang, Mingxi Cheng, Qi~Dai, Bei Liu, Kai
  Qiu, Yue Dong, Ji~Li, and Chong Luo.
\newblock {RESOURCE2SKILL: Distilling Executable Agent Skills from
  Human-Created Multimodal Resources}.
\newblock \emph{arXiv preprint arXiv:2606.29538}, 2026.
\newblock URL \url{https://arxiv.org/abs/2606.29538}.

\bibitem[Jiang et~al.(2026)Jiang, An, Liu, Ji, Wu, Andreas, Zhang, and
  Chang]{jiang2026visualskill}
Ziyan Jiang, Li~An, Yujian Liu, Jiabao Ji, Qiucheng Wu, Jacob Andreas, Yang
  Zhang, and Shiyu Chang.
\newblock {VISUALSKILL: Multimodal Skills for Computer-Use Agents}.
\newblock \emph{arXiv preprint arXiv:2606.18448}, 2026.
\newblock URL \url{https://arxiv.org/abs/2606.18448}.

\bibitem[Liu et~al.(2026)Liu, Luo, Li, Huang, Liu, and Qiao]{liu2026skillforge}
Xingyan Liu, Xiyue Luo, Linyu Li, Ganghong Huang, Jianfeng Liu, and Honglin
  Qiao.
\newblock {SkillForge: Forging Domain-Specific, Self-Evolving Agent Skills in
  Cloud Technical Support}.
\newblock \emph{arXiv preprint arXiv:2604.08618}, 2026.
\newblock URL \url{https://arxiv.org/abs/2604.08618}.

\bibitem[Lu et~al.(2026)Lu, Bai, Bai, Fan, Gao, Guan, Hu, Huang, Huang, Jiang,
  et~al.]{lu2026qwen}
Dunjie Lu, Shuai Bai, Tianyi Bai, Sicheng Fan, Chang Gao, Jian Guan, Feng Hu,
  Mianqiu Huang, Xingyang Huang, Yizhen Jiang, et~al.
\newblock Qwen-cua: Native computer use for (almost) everything.
\newblock \emph{arXiv preprint arXiv:2608.02352}, 2026.

\bibitem[Lv et~al.(2026)Lv, Liu, Ren, Lai, Jing, Zhang, Zhao, Yao, Tang, and
  Dong]{lv2026scalecua}
Bowen Lv, Xiao Liu, Yanyu Ren, Hanyu Lai, Bohao Jing, Hanchen Zhang, Yanxiao
  Zhao, Shuntian Yao, Jie Tang, and Yuxiao Dong.
\newblock Scalecua: Scaling computer use agents with verifiable task synthesis
  and efficient online rl.
\newblock \emph{arXiv preprint arXiv:2607.11185}, 2026.

\bibitem[Ni et~al.(2026)Ni, Liu, Liu, Sun, Zhou, Cheng, Wang, Zhao, Jiang, and
  Jiang]{ni2026trace2skill}
Jingwei Ni, Yihao Liu, Xinpeng Liu, Yutao Sun, Mengyu Zhou, Pengyu Cheng, Dexin
  Wang, Erchao Zhao, Xiaoxi Jiang, and Guanjun Jiang.
\newblock {Trace2Skill: Distill Trajectory-Local Lessons into Transferable
  Agent Skills}.
\newblock \emph{arXiv preprint arXiv:2603.25158}, 2026.
\newblock URL \url{https://arxiv.org/abs/2603.25158}.

\bibitem[Qin et~al.(2025)Qin, Ye, Fang, Wang, Liang, Tian, Zhang, Li, Li,
  Huang, Zhong, Li, Yang, Miao, Lin, Liu, Jiang, Ma, Li, Xiao, Cai, Li, Zheng,
  Jin, Li, Zhou, Wang, Chen, Li, Yang, Liu, Lin, Peng, Liu, and Shi]{qin2025ui}
Yujia Qin, Yining Ye, Junjie Fang, Haoming Wang, Shihao Liang, Shizuo Tian,
  Junda Zhang, Jiahao Li, Yunxin Li, Shijue Huang, Wanjun Zhong, Kuanye Li,
  Jiale Yang, Yu~Miao, Woyu Lin, Longxiang Liu, Xu~Jiang, Qianli Ma, Jingyu Li,
  Xiaojun Xiao, Kai Cai, Chuang Li, Yaowei Zheng, Chaolin Jin, Chen Li, Xiao
  Zhou, Minchao Wang, Haoli Chen, Zhaojian Li, Haihua Yang, Haifeng Liu, Feng
  Lin, Tao Peng, Xin Liu, and Guang Shi.
\newblock {UI-TARS: Pioneering Automated GUI Interaction with Native Agents}.
\newblock \emph{arXiv preprint arXiv:2501.12326}, 2025.
\newblock URL \url{https://arxiv.org/abs/2501.12326}.

\bibitem[Sager et~al.(2026)Sager, Meyer, Yan, von Wartburg-Kottler, Etaiwi,
  Enayati, Nobel, Abdulkadir, Grewe, and Stadelmann]{sager2026survey}
Pascal~J. Sager, Benjamin Meyer, Peng Yan, Rebekka von Wartburg-Kottler, Layan
  Etaiwi, Aref Enayati, Gabriel Nobel, Ahmed Abdulkadir, Benjamin~F. Grewe, and
  Thilo Stadelmann.
\newblock {A Comprehensive Survey of Agents for Computer Use: Foundations,
  Challenges, and Future Directions}.
\newblock \emph{Journal of Artificial Intelligence Research}, 85, 2026.
\newblock \doi{10.1613/jair.1.19490}.
\newblock URL \url{https://doi.org/10.1613/jair.1.19490}.

\bibitem[Shao et~al.(2026)Shao, Zhang, Yuan, Chen, Xiang, Yang, Ye, Zhang, Xu,
  Yan, et~al.]{shao2026bioinfoysis}
Qingyang Shao, Xin Zhang, Zhouyang Yuan, Xianying Chen, Yujia Xiang, Zihao
  Yang, Tong Ye, Yangqi Zhang, Jiakang Xu, Xiaoqing Yan, et~al.
\newblock Bioinfoysis technical report.
\newblock \emph{arXiv preprint arXiv:2609.03871}, 2026.

\bibitem[Shen et~al.(2026)Shen, Cheng, Ma, Turcan, Zhang, and
  Ma]{shen2026skillfoundry}
Shuaike Shen, Wenduo Cheng, Mingqian Ma, Alistair Turcan, Martin~Jinye Zhang,
  and Jian Ma.
\newblock {SKILLFOUNDRY: Building Self-Evolving Agent Skill Libraries from
  Heterogeneous Scientific Resources}.
\newblock \emph{arXiv preprint arXiv:2604.03964}, 2026.
\newblock URL \url{https://arxiv.org/abs/2604.03964}.

\bibitem[Shinn et~al.(2023)Shinn, Cassano, Gopinath, Narasimhan, and
  Yao]{shinn2023reflexion}
Noah Shinn, Federico Cassano, Ashwin Gopinath, Karthik Narasimhan, and Shunyu
  Yao.
\newblock {Reflexion: Language Agents with Verbal Reinforcement Learning}.
\newblock In \emph{Advances in Neural Information Processing Systems},
  volume~36, 2023.
\newblock URL
  \url{https://proceedings.neurips.cc/paper_files/paper/2023/hash/1b44b878bb782e6954cd888628510e90-Abstract-Conference.html}.

\bibitem[Sun et~al.(2026)Sun, Liu, Ma, Ding, Xu, Yin, Zhao, Wu, Cheng, Liu,
  Wang, Li, Tang, Xie, Feng, Li, Kao, Wang, Qi, Kong, and
  Wu]{sun2026scienceboard}
Qiushi Sun, Zhoumianze Liu, Chang Ma, Zichen Ding, Fangzhi Xu, Zhangyue Yin,
  Haiteng Zhao, Zhenyu Wu, Kanzhi Cheng, Zhaoyang Liu, Jianing Wang, Qintong
  Li, Xiangru Tang, Tianbao Xie, Xiachong Feng, Xiang Li, Ben Kao, Wenhai Wang,
  Biqing Qi, Lingpeng Kong, and Zhiyong Wu.
\newblock {ScienceBoard: Evaluating Multimodal Autonomous Agents in Realistic
  Scientific Workflows}.
\newblock In \emph{International Conference on Learning Representations}, 2026.
\newblock URL
  \url{https://proceedings.iclr.cc/paper_files/paper/2026/hash/7a9745f251508a053425a256490b0665-Abstract-Conference.html}.

\bibitem[Sun et~al.(2025)Sun, Liu, Zang, Cao, Dong, Wu, Lin, and
  Wang]{sun2025seagent}
Zeyi Sun, Ziyu Liu, Yuhang Zang, Yuhang Cao, Xiaoyi Dong, Tong Wu, Dahua Lin,
  and Jiaqi Wang.
\newblock {SEAgent: Self-Evolving Computer Use Agent with Autonomous Learning
  from Experience}.
\newblock \emph{arXiv preprint arXiv:2508.04700}, 2025.
\newblock URL \url{https://arxiv.org/abs/2508.04700}.

\bibitem[Xie et~al.(2024)Xie, Zhang, Chen, Li, Zhao, Cao, Hua, Cheng, Shin,
  Lei, Liu, Xu, Zhou, Savarese, Xiong, Zhong, and Yu]{xie2024osworld}
Tianbao Xie, Danyang Zhang, Jixuan Chen, Xiaochuan Li, Siheng Zhao, Ruisheng
  Cao, Toh~Jing Hua, Zhoujun Cheng, Dongchan Shin, Fangyu Lei, Yitao Liu,
  Yiheng Xu, Shuyan Zhou, Silvio Savarese, Caiming Xiong, Victor Zhong, and Tao
  Yu.
\newblock {OSWorld: Benchmarking Multimodal Agents for Open-Ended Tasks in Real
  Computer Environments}.
\newblock In \emph{Advances in Neural Information Processing Systems},
  volume~37, 2024.
\newblock URL
  \url{https://proceedings.neurips.cc/paper_files/paper/2024/hash/5d413e48f84dc61244b6be550f1cd8f5-Abstract-Datasets_and_Benchmarks_Track.html}.

\bibitem[Yang et~al.(2026)Yang, Gong, Huang, Yang, Zhou, Huang, Li, Gao, Dai,
  Liu, Qiu, Yang, Chen, Yang, and Luo]{yang2026skillopt}
Yifan Yang, Ziyang Gong, Weiquan Huang, Qihao Yang, Ziwei Zhou, Zisu Huang, Yan
  Li, Xuemei Gao, Qi~Dai, Bei Liu, Kai Qiu, Yuqing Yang, Dongdong Chen, Xue
  Yang, and Chong Luo.
\newblock {SkillOpt: Executive Strategy for Self-Evolving Agent Skills}.
\newblock \emph{arXiv preprint arXiv:2605.23904}, 2026.
\newblock URL \url{https://arxiv.org/abs/2605.23904}.

\bibitem[Zhang et~al.(2026)Zhang, Fan, Zou, Chen, Wang, Zhou, Li, Huang, Yao,
  Zheng, Liu, Li, and Yu]{zhang2026coevoskills}
Hanrong Zhang, Shicheng Fan, Henry~Peng Zou, Yankai Chen, Zhenting Wang, Jiayu
  Zhou, Chengze Li, Wei-Chieh Huang, Yifei Yao, Kening Zheng, Xue~(Steve) Liu,
  Xiaoxiao Li, and Philip~S. Yu.
\newblock {CoEvoSkills: Self-Evolving Agent Skills via Co-Evolutionary
  Verification}.
\newblock \emph{arXiv preprint arXiv:2604.01687}, 2026.
\newblock URL \url{https://arxiv.org/abs/2604.01687}.

\bibitem[Zhao et~al.(2024)Zhao, Huang, Xu, Lin, Liu, and Huang]{zhao2024expel}
Andrew Zhao, Daniel Huang, Quentin Xu, Matthieu Lin, Yong-Jin Liu, and Gao
  Huang.
\newblock {ExpeL: LLM Agents Are Experiential Learners}.
\newblock In \emph{Proceedings of the AAAI Conference on Artificial
  Intelligence}, volume~38, pp.\  19632--19642, 2024.
\newblock \doi{10.1609/aaai.v38i17.29936}.
\newblock URL \url{https://ojs.aaai.org/index.php/AAAI/article/view/29936}.

\bibitem[Zhou et~al.(2024)Zhou, Xu, Zhu, Zhou, Lo, Sridhar, Cheng, Ou, Bisk,
  Fried, Alon, and Neubig]{zhou2024webarena}
Shuyan Zhou, Frank~F. Xu, Hao Zhu, Xuhui Zhou, Robert Lo, Abishek Sridhar,
  Xianyi Cheng, Tianyue Ou, Yonatan Bisk, Daniel Fried, Uri Alon, and Graham
  Neubig.
\newblock {WebArena: A Realistic Web Environment for Building Autonomous
  Agents}.
\newblock In \emph{International Conference on Learning Representations}, 2024.
\newblock URL \url{https://openreview.net/forum?id=oKn9c6ytLx}.

\end{thebibliography}
\bibliographystyle{iclr2027_conference}

\appendix

\setlength{\LTcapwidth}{\textwidth}

\clearpage
\begingroup
\begin{center}
  {\Large\bfseries\color{DeepAutonomyDarkBlue}Appendix Overview\par}
\end{center}

\vspace{1.0em}

\normalsize
\setlength{\emergencystretch}{1em}
\noindent This appendix contains the following parts:\par

\begin{list}{\labelitemi}{%
  \setlength{\leftmargin}{\leftmargini}%
  \setlength{\topsep}{3pt}%
  \setlength{\partopsep}{0pt}%
  \setlength{\itemsep}{4pt}%
  \setlength{\parsep}{0pt}%
}
  \item[{\textcolor[HTML]{F2C200}{\raisebox{-0.1ex}{\scalebox{1.25}{\(\bigstar\)}}}}] \hyperref[app:research-takeaways]{\textbf{Appendix~\ref*{app:research-takeaways}: Discussion and Implications}}\quad\textbf{Recommended reading} --- applicability, knowledge organization, and base model capabilities.
  \item \hyperref[app:direct-injection-details]{\textbf{Appendix~\ref*{app:direct-injection-details}: Direct Documentation Injection Details}} --- model-based injection protocol, context-window constraints, and selected documentation statistics.
  \item \hyperref[app:rag-details]{\textbf{Appendix~\ref*{app:rag-details}: Retrieval-Augmented Generation Details}} --- offline corpus preparation, structure-preserving chunking, hybrid retrieval with reranking, and frozen task bundles.
  \item \hyperref[app:skillfoundry-reproduction]{\textbf{Appendix~\ref*{app:skillfoundry-reproduction}: SkillFoundry Reproduction Details}} --- reproduction of SkillFoundry as a Resource-to-Skill baseline, including setup, inputs, and procedure.
  \item \hyperref[app:visualskill-reproduction]{\textbf{Appendix~\ref*{app:visualskill-reproduction}: VisualSkill Reproduction Details}} --- reproduction of VisualSkill as a Resource-to-Skill baseline, including setup, inputs, and procedure.
  \item \hyperref[app:design]{\textbf{Appendix~\ref*{app:design}: Resource-to-Skill Implementation Details}} --- pipeline details, including child-session contracts and deterministic validation.
  \item \hyperref[app:scienceboard_verified_repairs]{\textbf{Appendix~\ref*{app:scienceboard_verified_repairs}: ScienceBoard-Verified: Detailed Repairs}} --- audited revisions: instruction ambiguity, task and answer errors, evaluator bias, and environment repairs.
  \item \hyperref[app:task_deduplication]{\textbf{Appendix~\ref*{app:task_deduplication}: Task Construction and Deduplication}} --- construction of the 108 evolution tasks and deduplication against the test set.
  \item \hyperref[app:self_evolution]{\textbf{Appendix~\ref*{app:self_evolution}: Execution-Guided Skill Self-Evolution: Additional Details}} --- history format, failure diagnosis, validation, submission, and consolidation protocols.
  \item \hyperref[app:case-kalgebra-c05]{\textbf{Appendix~\ref*{app:case-kalgebra-c05}: Case Study: Complementary Roles of Skills and References in KAlgebra}} --- three-way C-05 comparison of script-loading guidance, expression knowledge, and task completion.
  \item \hyperref[app:case-self-evolve-kalgebra]{\textbf{Appendix~\ref*{app:case-self-evolve-kalgebra}: Case Study: How Self-Evolve Repairs Resource--Environment Mismatches and Improves Task Performance}} --- two KAlgebra skill revisions, from execution failure to transfer.
  \item \hyperref[app:case-grass-reference]{\textbf{Appendix~\ref*{app:case-grass-reference}: Case Study: Adding a Module-Level Reference for GRASS GIS}} --- addition of task-relevant reference knowledge and its selection on related test tasks.
  \item \hyperref[app:case-celestia-a05]{\textbf{Appendix~\ref*{app:case-celestia-a05}: Case Study: Outdated Interface Guidance in Celestia}} --- how guidance inherited from an outdated source interface consumes execution steps and causes task failure.
  \item \hyperref[app:lean-texstudio-scope]{\textbf{Appendix~\ref*{app:lean-texstudio-scope}: Task Requirements and Documentation Scope for Lean 4 and TeXstudio}} --- Verified task requirements compared with the source documentation used for the initial libraries.
  \item \hyperref[app:resource2skill-scope]{\textbf{Appendix~\ref*{app:resource2skill-scope}: Research Setting and Baseline Reproduction Constraints of Resource2Skill}} --- research setting, knowledge objectives, and limitations in adapting the released implementation.
  \item \hyperref[app:frontier-models]{\textbf{Appendix~\ref*{app:frontier-models}: Supplementary Evaluation of Frontier Models}} --- model performance and gains from the self-evolved ScienceCUA library.
\end{list}
\endgroup

\clearpage

\section{Discussion and Implications}
\label{app:research-takeaways}

\paragraph{Niche and private software are priority settings for our approach.}
Knowledge of niche, private, or self-developed software may be underrepresented during general models' pre-training and post-training, leaving gaps in their understanding of specific functions, workflows, and usage constraints. Extracting knowledge from documentation therefore has the potential to yield larger performance gains for such software. Execution feedback can then guide revisions and additions that adapt this knowledge to the deployed environment and task requirements. For widely used software whose operational knowledge is well covered during training, models may already know much of the documented material, leaving less room for improvement through external knowledge.

\paragraph{Knowledge organization should reflect software and task requirements.}
Scientific software requires both operational workflows and precise parameter constraints, syntax rules, and command and function catalogs. These details are often shared across tasks and consulted only when a step requires them. ScienceCUA therefore organizes goal-oriented procedural knowledge as Skills and shared lookup knowledge as References, allowing agents to follow workflows while accessing specific rules and details on demand. Other software and scenarios may require different forms of knowledge organization. Applying our approach to these settings can retain the process of constructing knowledge from resources and improving it through execution, while adapting knowledge organization to the target setting. Such adaptation could include exploring code-based or visual representations, whose effectiveness we have not systematically compared within the same framework.

\paragraph{Base model capabilities affect the value of external knowledge augmentation.}
Recent frontier models, such as GPT-6 Astra, demonstrate stronger computer-use capabilities. As models become able to complete more scientific-software tasks on their own, the additional gains from ScienceCUA's self-evolved Skills and References may diminish, a pattern observed in our supplementary experiments (Appendix~\ref{app:frontier-models}). Similar effects may arise in other CUA settings: when a base model already possesses the relevant operational knowledge, the room for improvement from Skills may shrink; if their guidance is outdated, inapplicable to the task, or imposes unnecessary constraints on execution paths, it may even interfere with the model's own judgment and lead to worse performance than the base model without Skills. The value of external knowledge therefore needs to be evaluated in relation to the base model's capabilities and the specific task. Practical deployment must also balance task performance against inference cost. In budget-constrained settings that call for lower-cost models, constructing and continually improving external knowledge from software resources retains practical value. Moreover, the software covered by ScienceBoard has publicly available documentation, and models may already have learned some of this operational knowledge. For niche, private, or self-developed software whose knowledge is underrepresented during model training, external knowledge may still fill capability gaps and yield larger gains. This expectation warrants further empirical validation.

\clearpage
\section{Direct Documentation Injection Details}
\label{app:direct-injection-details}

Direct injection follows the model-based evaluation protocol in Section~\ref{sec:native_agent_evaluation}. Each API request includes the system prompt, task instruction, screenshots and accessibility trees from the three most recent steps together with the model-generated actions for those steps, and the current screenshot and accessibility tree. Documentation for the corresponding application is added to every request. ScienceBoard executes the action code or control instructions returned by the model.

Direct injection is constrained by the model's context window. Table~\ref{tab:direct-injection-context-windows} lists the officially documented context limits of the four evaluated models. These windows must accommodate documentation, task instructions, interaction history, and other inputs while leaving room for model output; they cannot be allocated entirely to documentation.

\begin{center}
    \centering
    \captionof{table}{Official context-window limits of the evaluated models.}
    \label{tab:direct-injection-context-windows}
    \begingroup
    \small
    \setlength{\tabcolsep}{6pt}
    \renewcommand{\arraystretch}{1.12}
    \begin{tabular}{@{}lll@{}}
        \toprule
        \textbf{Model} & \textbf{Context window} & \textbf{Official source} \\
        \midrule
        Qwen-3.8 Max & 1M tokens & \href{https://docs.qwencloud.com/developer-guides/getting-started/vision-models}{Qwen documentation} \\
        GLM-5.2 & 1M tokens & \href{https://docs.z.ai/guides/llm/glm-5.2}{Z.AI documentation} \\
        Grok-4.6 & 500,000 tokens & \href{https://docs.x.ai/developers/models/grok-4.6}{Grok model documentation} \\
        GPT-5.5 & 1,050,000 tokens & \href{https://developers.openai.com/api/docs/models/gpt-5.5}{OpenAI model documentation} \\
        \bottomrule
    \end{tabular}
    \endgroup
\end{center}

The complete documentation for some applications also contains substantial amounts of text, requiring considerable context space for full injection. As shown in Table~\ref{tab:direct-injection-documentation-size}, the full GRASS GIS and ChimeraX documentation contains approximately 12.2 million and 1.7 million text characters, respectively. Injecting these materials in full at every step, alongside task and interaction inputs, creates a risk of exceeding the context window.

\begin{center}
    \centering
    \captionof{table}{Text sizes of the full documentation and the documentation selected for direct injection.}
    \label{tab:direct-injection-documentation-size}
    \begingroup
    \small
    \setlength{\tabcolsep}{6pt}
    \renewcommand{\arraystretch}{1.12}
    \begin{tabular}{@{}lcrr@{}}
        \toprule
        \textbf{Software} & \textbf{Format} & \begin{tabular}[c]{@{}r@{}}\textbf{Full documentation}\\\textbf{(characters)}\end{tabular} & \begin{tabular}[c]{@{}r@{}}\textbf{Selected documentation}\\\textbf{(characters)}\end{tabular} \\
        \midrule
        KAlgebra & PDF & 27,572 & 27,572 \\
        Celestia & PDF & 146,761 & 146,761 \\
        GRASS GIS & HTML & 12,232,449 & 367,177 \\
        ChimeraX & HTML & 1,704,478 & 254,697 \\
        Lean 4 & HTML & 527,662 & 527,662 \\
        TeXstudio & HTML & 243,645 & 243,645 \\
        \bottomrule
    \end{tabular}
    \endgroup
\end{center}

Full documentation refers to all materials within the documentation scope specified in this work. Both columns use the same character-counting procedure: text is extracted from PDF text layers or HTML with tags removed, whitespace is normalized, and characters are counted including whitespace. Navigation text is retained for HTML.

Given these context limits and documentation sizes, we use a fixed documentation configuration for each application. KAlgebra and Celestia receive their complete original PDF files. For Lean 4 and TeXstudio, we acquire all pages within the specified documentation scope and inject their normalized text in full. For GRASS GIS and ChimeraX, we use bounded page acquisition: the root page is retained, and 40 child pages are selected in traversal-depth order with round-robin scheduling across root-level chapter branches. The normalized text of all selected pages is then injected in full. All documentation is fixed before evaluation, shared across tasks and evaluated models for the same application, and included in full in every execution request, without task-specific retrieval or dynamic selection during execution.

\section{Retrieval-Augmented Generation Details}
\label{app:rag-details}

This appendix describes the retrieval-augmented generation (RAG) pipeline. Offline preprocessing freezes one corpus per application, applies structure-preserving chunking and hybrid retrieval, and materializes task-specific context bundles for deterministic loading at execution. All tasks of an application share the same corpus. Documents are never manually paired with tasks.

\subsection{Frozen Documentation Corpora}

We construct one physically isolated corpus for each of the six applications. Table~\ref{tab:rag-corpora} reports its documentation input and size. Raw inputs, normalized text, source locators, acquisition records, and manifests are all frozen before retrieval. Bundle generation and benchmark execution use only these local snapshots, shared across all tasks and evaluated models for an application.

\begin{center}
    \centering
    \captionof{table}{Frozen documentation corpora used by RAG. Documents are semantic sections when reliable boundaries are available and page-level fallbacks otherwise. GRASS GIS and ChimeraX counts cover all successfully acquired pages.}
    \label{tab:rag-corpora}
    \begingroup
    \small
    \setlength{\tabcolsep}{4pt}
    \renewcommand{\arraystretch}{1.12}
    \begin{tabularx}{\textwidth}{@{}lcc>{\raggedright\arraybackslash}Xrrr@{}}
        \toprule
        \textbf{Software} & \textbf{Environment} & \textbf{Format} & \textbf{Frozen documentation source} & \textbf{Pages} & \textbf{Documents} & \textbf{Chunks} \\
        \midrule
        KAlgebra & 24.11.70 & PDF & \emph{KAlgebra Handbook} & 32 & 88 & 96 \\
        Celestia & 1.7.0 & PDF & \emph{Celestia 1.6.1 User's Guide} & 48 & 48 & 57 \\
        GRASS GIS & 8.4.1 & HTML & GRASS GIS 8.4.1 Reference Manual & 1,468 & 11,444 & 24,236 \\
        ChimeraX & 1.9 & HTML & UCSF ChimeraX online User Guide, frozen at acquisition & 277 & 726 & 859 \\
        Lean 4 & 4.14.0 & HTML & \emph{Theorem Proving in Lean 4}, version 4.26.0 & 13 & 113 & 146 \\
        TeXstudio & 4.8.6 & HTML & TeXstudio 4.9.8 documentation & 9 & 269 & 294 \\
        \midrule
        \textbf{Total} & & & & \textbf{1,847} & \textbf{12,688} & \textbf{25,688} \\
        \bottomrule
    \end{tabularx}
    \endgroup
\end{center}

\paragraph{PDF normalization.}
PDF text is extracted with PyMuPDF from the embedded text layer. OCR is disabled and raster images are not indexed. Repeated headers and footers are removed. We infer a hierarchy from chapter, section, appendix, and numbered headings and retain page ranges as source locators. When a reliable semantic heading is unavailable, the corresponding page is kept as a page-level document rather than discarded. This yields 81 semantic sections and 7 page fallbacks for KAlgebra. The Celestia guide is retained as 48 page-level documents because its layout does not provide sufficiently reliable semantic section boundaries.

\paragraph{HTML acquisition and freezing.}
Each HTML corpus begins at a fixed official documentation root. URLs are canonicalized and restricted to the documentation scope. The reachable in-scope link queue is traversed to exhaustion, with no page-count limit, and every successful response is frozen with its requested URL, final URL, acquisition time, raw HTML hash, outgoing links, and local file locator. GRASS GIS contains 1,468 successful pages and ChimeraX contains 277. Both acquisition queues have zero pending URLs. Images, video, CSS, JavaScript, PDFs, and pages outside the documentation scope are excluded. The Lean 4 and TeXstudio documentation trees contain 13 and 9 in-scope pages, respectively, all of which are retained.

\paragraph{HTML normalization.}
For every retained page, the extractor selects the \texttt{main}, \texttt{article}, or body content and removes scripts, styles, templates, and navigation chrome. It segments the remaining content at \texttt{h1}--\texttt{h6} boundaries. Paragraphs, lists, tables, and preformatted code are preserved in reading order, with code represented as fenced blocks. Exact duplicate pages are removed by content hash. Raw HTML and normalized Markdown-compatible text are both retained for audit, but only normalized text is indexed. This procedure is deterministic and uses no language model.

\subsection{Structure-Preserving Chunking}

Chunk boundaries are computed with the tokenizer of \texttt{BAAI/bge-m3}. Each indexable text is prefixed with its document title and complete section breadcrumb. The target length is 700 tokens and the hard maximum, including this prefix, is 1,000 tokens. Text that exceeds the limit is split preferentially at paragraph, line, or sentence boundaries with an 80-token overlap between consecutive pieces. Chunks never cross documents or semantic sections. Exact duplicates are removed using the SHA-256 hash of the complete index text. Near-duplicate removal is not applied. The counts in Table~\ref{tab:rag-corpora} are post-deduplication, and no resulting chunk exceeds the hard limit.

\subsection{Hybrid Retrieval and Reranking}

Retrieval is performed independently for each application. The query is exactly the stripped task instruction. The software name is used only to select the corresponding corpus and is not appended to the query. Cross-application retrieval is disabled.

We retrieve 30 candidates with BM25 and 30 with a dense encoder. BM25 uses $k_1=1.2$, $b=0.75$, binary query term frequency, and a Unicode-aware tokenizer that preserves compound identifiers. Dense retrieval uses \texttt{BAAI/bge-m3} with a maximum input length of 1,024 tokens, CLS pooling, and L2-normalized embeddings. Encoding uses FP16 on CUDA with batch size 2. The resulting 1,024-dimensional vectors are stored in FP32. Similarity is exact cosine similarity, implemented as a dot product over normalized vectors, without approximate-nearest-neighbor search.

The two ranked lists are combined by equal-weight reciprocal rank fusion (RRF):
\begin{equation}
    \operatorname{RRF}(d)
    = \sum_{s \in \{\mathrm{BM25},\mathrm{dense}\}}
      \frac{1}{60 + \operatorname{rank}_{s}(d)}.
\end{equation}
The 40 highest-scoring fused candidates are reranked with \texttt{BAAI/bge-reranker-v2-m3}. The reranker uses a maximum input length of 1,024 tokens, FP16 CUDA inference, batch size 4, and its unnormalized output score. Ties are resolved deterministically using the fused score and chunk identifier.

We select up to five reranked chunks whose rendered references fit a 6,000-token context budget. Each numbered reference gives its rank, document title, section breadcrumb, and chunk text. Retrieval scores are hidden. The references are enclosed by documentation markers and prepended to the unmodified original task instruction.

\subsection{Frozen Task Bundles and Runtime Protocol}

Offline retrieval produces one frozen bundle per ScienceBoard-Verified task, storing the query, corpus and index identities, configuration hash, candidate rankings, final five chunks, and rendered instruction. The 168 bundles comprise 31 for KAlgebra, 33 for Celestia, 34 for GRASS GIS, 28 for ChimeraX, 26 for Lean 4, and 16 for TeXstudio. Each is reused across all evaluated models.

At initialization, hashes and identities are verified. Invalid bundles fail without live fallback. Runtime uses no retrieval assets or evaluation data.

\clearpage
\section{SkillFoundry Reproduction Details}
\label{app:skillfoundry-reproduction}

This appendix describes our reproduction of SkillFoundry as a Resource-to-Skill baseline.

\subsection{Reproduction Setup}

We reproduced the complete SkillFoundry pipeline from an empty repository state, without an external taxonomy or any pre-existing resource or skill library. Each target software package was processed in an isolated repository, and no task list or task-specific resource pairing was provided. Under this setting, SkillFoundry was responsible for constructing the software-specific taxonomy, discovering supporting resources, and building reusable skills from the retrieved evidence.

\par\medskip\noindent
\begin{minipage}{\linewidth}
\centering
\captionof{table}{Inputs used for the SkillFoundry reproduction.}
\label{tab:skillfoundry-inputs}
\begingroup
\small
\setlength{\tabcolsep}{12pt}
\renewcommand{\arraystretch}{1.12}
\begin{tabular}{@{}lc@{}}
\toprule
\textbf{Software} & \textbf{Environment version} \\
\midrule
KAlgebra & 24.11.70 \\
Celestia & 1.7.0 \\
GRASS GIS & 8.4.1 \\
Lean 4 & 4.14.0 \\
TeXstudio & 4.8.6 \\
ChimeraX & 1.9 \\
\bottomrule
\end{tabular}
\endgroup
\end{minipage}
\par\medskip

\subsection{Inputs and Configuration}

Each run received only the software name and environment version. Table~\ref{tab:skillfoundry-inputs} lists the six targets used in our experiments. We used the same execution configuration for all targets: stage-workers=1, background-validation-limit=0, and verification-mode=standard.

\subsection{Reproduction Procedure}

For each target, we executed one complete five-stage cycle in the standard order: \texttt{tree\_check}, \texttt{resource\_search}, \texttt{skill\_build}, \texttt{skill\_test}, and \texttt{refresh}. The pipeline first constructed a software-specific capability taxonomy, then searched for and registered relevant external resources. It subsequently synthesized skills from the registered evidence, ran SkillFoundry's repository-local validation, and synchronized the resulting taxonomy, resource registry, and skill registry. The generated skills were then evaluated on ScienceBoard-Verified using the same downstream protocol as the other Resource-to-Skill methods.

\section{VisualSkill Reproduction Details}
\label{app:visualskill-reproduction}

This appendix describes our reproduction of VisualSkill as a Resource-to-Skill baseline.

\subsection{Reproduction Setup}

We reproduced VisualSkill's standard multimodal Stage~1 pipeline rather than its task-filtered mode. Each run processed one target application with exactly one curated documentation source, either a PDF manual or an HTML documentation root. PDF and HTML inputs were not mixed within a run, no external task list was provided for taxonomy filtering, and the optional text-derivation phase was not used.

\subsection{Inputs and Configuration}

Table~\ref{tab:visualskill-inputs} lists the documentation source supplied for each target, with public links to the documentation resources.

\begin{table}[htbp]
\centering
\caption{Inputs used for the VisualSkill Stage~1 reproduction.}
\label{tab:visualskill-inputs}
\begingroup
\small
\setlength{\tabcolsep}{3pt}
\renewcommand{\arraystretch}{1.15}
\begin{tabularx}{\linewidth}{@{}>{\raggedright\arraybackslash}p{0.14\linewidth}>{\centering\arraybackslash}p{0.14\linewidth}>{\centering\arraybackslash}p{0.14\linewidth}>{\centering\arraybackslash}p{0.07\linewidth}>{\raggedright\arraybackslash}X@{}}
\toprule
\textbf{Software} & \shortstack{\textbf{Environment}\\\textbf{version}} & \shortstack{\textbf{Resource}\\\textbf{version}} & \textbf{Format} & \textbf{Documentation source} \\
\midrule
\textbf{KAlgebra} & 24.11.70 & 20.12 & PDF & \url{https://docs.kde.org/stable_kf6/en/kalgebra/kalgebra/kalgebra.pdf} \\
\addlinespace[2pt]
\textbf{Celestia} & 1.7.0 & 1.6.1 & PDF & \url{https://drive.google.com/file/u/1/d/1dxMnahknnVjrckqWOyJhvuS8IA0BCT0l/view} \\
\addlinespace[2pt]
\textbf{ChimeraX} & 1.9 & 1.9.0 & HTML & \url{https://web.archive.org/web/20241215145343id_/https://www.cgl.ucsf.edu/chimerax/docs/user/index.html}; root prefix \nolinkurl{/chimerax/docs/user/} \\
\addlinespace[2pt]
\textbf{GRASS GIS} & 8.4.1 & 8.4.1 & HTML & \url{https://lira.no-ip.org:8443/doc/grass-doc/html/index.html} \\
\addlinespace[2pt]
\textbf{Lean 4} & 4.14.0 & 4.26.0 & HTML & \url{https://lean-lang.org/theorem_proving_in_lean4/}; root prefix \nolinkurl{/theorem_proving_in_lean4/} \\
\addlinespace[2pt]
\textbf{TeXstudio} & 4.8.6 & 4.9.8 & HTML & \url{https://texstudio-org.github.io/} \\
\bottomrule
\end{tabularx}
\endgroup
\end{table}

We used \texttt{--mode multimodal} for all six targets. PDF sources were supplied through \texttt{sources.\allowbreak pdf\_guide.\allowbreak url}, whereas HTML sources were supplied through \texttt{sources.html\_guide.\allowbreak root\_\allowbreak url}; otherwise, the same Stage~1 construction setting was used across targets.

\subsection{Reproduction Procedure}

For each target, we ran the standard Stage~1 construction pipeline. VisualSkill first built a category--topic taxonomy from the supplied documentation, then prepared topic-level textual and visual evidence from the corresponding PDF pages or HTML pages. It next generated a multimodal \texttt{guide.md} for each topic and produced a \texttt{use\_when} routing description. These topic-level artifacts were finally assembled into the top-level \texttt{SKILL.md}. The generated skills were then evaluated on ScienceBoard-Verified using the same downstream protocol as the other Resource-to-Skill methods.

\begingroup
\raggedbottom
\DefineVerbatimEnvironment{RtwoSPrompt}{Verbatim}{%
  fontsize=\fontsize{8.5}{9.5}\selectfont,baselinestretch=1,
  breaklines=true,breakanywhere=true,breakautoindent=true,
  breakindent=1em,breaksymbolleft={},breaksymbolright={},
  frame=lines,framesep=5pt,vspace=4pt,rulecolor=\color{black!25}}
\section{Resource-to-Skill Implementation Details}
\label{app:design}

This appendix provides implementation details for the Resource-to-Skill pipeline in Sec.~\ref{sec:resource_to_skill}. The main text focuses on the knowledge-construction logic; here we report the executable child-session contracts and deterministic validation. Lower-level parser corner cases, full tool JSON schemas, retry/error handling, and storage internals are left to the released implementation.

\subsection{DSH Child Session Contracts}
\label{app:dsh_child_sessions}

A DSH child session is defined by more than its prompt. For an Agent $a$, the complete executable contract is
\begin{equation}
    \mathfrak{C}_a=(P_a,I_a,T_a,B_a,O_a,V_a,\tau_a),
\end{equation}
where $P_a$ is the prompt, $I_a$ is the input scope, $T_a$ is the allowed tool surface, $B_a$ is the tool budget, $O_a$ is the structured output contract, $V_a$ is the Python validator, and $\tau_a$ is the termination protocol. A child session can observe only authorized objects exposed through $T_a$ and can modify pipeline state only through the designated structured submission tools.

The HTML and PDF Evidence Architect contracts are source-type-specific and mutually exclusive. The HTML child accepts only one normalized HTML root scope, comprising the root page and all eligible in-scope child pages; the PDF child accepts only one normalized PDF document. Neither child may be dispatched a source of the other type.

\begingroup
\small
\setlength{\tabcolsep}{5pt}
\renewcommand{\arraystretch}{1.15}
\begin{longtable}{@{}>{\raggedright\arraybackslash}p{0.23\linewidth}>{\raggedright\arraybackslash}p{\dimexpr0.77\linewidth-2\tabcolsep\relax}@{}}
\caption{Executable contracts for the Resource-to-Skill child sessions. Budgets refer to the default production configuration; protocol-reserved calls are listed separately where applicable.}\label{tab:child_contracts}\\
\toprule
\textbf{Stage} & \textbf{Executable contract} \\
\midrule
\endfirsthead
\multicolumn{2}{@{}c@{}}{\small\itshape Table \thetable{} (continued)}\\
\toprule
\textbf{Stage} & \textbf{Executable contract} \\
\midrule
\endhead
\midrule
\endfoot
\bottomrule
\endlastfoot
Evidence Architect (HTML) & \textbf{Input.} One normalized HTML root scope (root page plus all eligible in-scope child pages); root-authored structure; candidate Evidence skeleton\par\textbf{Tools.} \nolinkurl{inspect_source}, \nolinkurl{get_root_structure}, \nolinkurl{submit_evidence_revision}\par\textbf{Output / budget.} \nolinkurl{EvidenceRevision}; exactly 3 calls in the default workflow \\
\addlinespace[4pt]
Evidence Architect (PDF) & \textbf{Input.} One normalized PDF document; selected navigation tree; content assignments; residual content; candidate skeleton\par\textbf{Tools.} \nolinkurl{inspect_source}, \nolinkurl{get_pdf_document_structure}, bounded \nolinkurl{read_pdf_content} and \nolinkurl{read_evidence_node}, \nolinkurl{submit_evidence_revision}\par\textbf{Output / budget.} \nolinkurl{EvidenceRevision}; at most 6 calls \\
\addlinespace[4pt]
Reference Build & \textbf{Input.} Frozen CorpusSnapshot and Python-generated review manifest\par\textbf{Tools.} Candidate/catalog inspection, bounded Evidence reading, \nolinkurl{submit_reference_review}\par\textbf{Output / budget.} Reference review decisions; at most 8 calls \\
\addlinespace[4pt]
Task Discovery & \textbf{Input.} Frozen CorpusSnapshot\par\textbf{Tools.} Evidence inventory, search/read/image inspection, Task batch submission\par\textbf{Output / budget.} \nolinkurl{TaskDraft}; exploration budget $\max(24,\min(40,3\,\texttt{max\_steps}))$ \\
\addlinespace[4pt]
Trajectory Inference & \textbf{Input.} One validated Task plus Corpus/Knowledge snapshots\par\textbf{Tools.} Evidence, Task, Reference and Reference-entry reads; trajectory submission\par\textbf{Output / budget.} \nolinkurl{TrajectoryDraft}; $\max(24,2\,\texttt{max\_steps})$ plus 2 protocol calls \\
\addlinespace[4pt]
Skill Induction & \textbf{Input.} A batch of validated Trajectories plus KnowledgeSnapshot\par\textbf{Tools.} Upstream-object reads, Skill submission, coverage inspection/review\par\textbf{Output / budget.} \nolinkurl{SkillDraft}; $\max(40,3\,\texttt{max\_steps})$ plus 6 protocol calls \\
\end{longtable}
\endgroup

\subsection{Deterministic Validation and Control}
\label{app:validation}

All Agent outputs are validated by Python before entering a downstream stage. A textual citation must contain a quote that exactly matches the corresponding Evidence text; a visual claim must reference an authorized and inspected image artifact. Citations cannot point to nonexistent or out-of-scope Evidence, and downstream objects can reference only upstream identifiers present in the current frozen snapshot.

Stage-specific validation additionally checks schema constraints, duplicate candidates, field lengths, step ordering, and cross-object consistency. For Trajectories, Python verifies that the submitted \nolinkurl{task_id} matches the dispatched Task, that exactly one acceptable trajectory is returned, that step ordinals are consecutive, and that linked References exist and satisfy the \nolinkurl{reference_guided} dependency rules. For Skills, Python validates Task/Trajectory/Reference relations, step structure, exact citations, and the coverage protocol over every dispatched Trajectory.

Pipeline state changes only through structured submission tools. Ordinary natural-language replies do not commit business results. Key stages freeze immutable CorpusSnapshot and KnowledgeSnapshot objects so that downstream child sessions read stable inputs. Attempt/Receipt bookkeeping, idempotent retry, recovery, and storage details are implementation mechanisms and are documented in the released code rather than expanded here.

\endgroup

\section{ScienceBoard-Verified: Detailed Repairs}
\label{app:scienceboard_verified_repairs}

\begingroup
\makeatletter
\renewcommand{\subsection}{\@startsection{subsection}{2}{\z@}
  {-1.8ex plus -0.5ex minus -.2ex}{0.8ex plus .2ex}
  {\normalfont\normalsize\bfseries\raggedright}}
\renewcommand{\paragraph}{\@startsection{paragraph}{4}{\z@}
  {1.5ex plus 0.5ex minus .2ex}{-1em}
  {\normalfont\normalsize\mdseries}}
\makeatother

This appendix documents the 37 task revisions and one task removal made when constructing ScienceBoard-Verified, corresponding to the 38 revised or removed tasks summarized in Figure~\ref{fig:scienceboard_ve}. A task may involve multiple issues; we assign it to one category to avoid double counting and describe secondary changes in the same entry. The first three categories contain 15, 12, and 10 revised tasks, respectively. The fourth contains the removed task and shared evaluation-environment repairs, which are not counted as additional task revisions.

\subsection{Instruction Ambiguity}
\label{app:verified_instructions}

These revisions clarify the intended target, information source, or completion requirements so that the instructions explicitly state what the evaluator expects.

\paragraph{1. Celestia A-02: Time direction and rate.}
The original instruction, ``-20x slower,'' could be interpreted as running backward at 20 times normal speed, whereas the evaluator expects a time scale of 0.05. The revised instruction explicitly requests forward playback at 0.05x, or 20 times slower than real time.

\paragraph{2. Celestia C-06: Ambiguous G0 star selection.}
The instruction asks for the name of the ``only'' G0 star, but the software lists multiple stars of this spectral type. We specify the closest G0 star to identify the intended object unambiguously.

\paragraph{3. Celestia C-10: Unspecified distance convention.}
Asking for the globular cluster ``farthest from the Milky Way'' could prompt agents to change the distance reference. The revision instead specifies the cluster with the largest \texttt{Distance (ly)} value in Celestia's Deep Sky Objects browser and requests its radius in light-years to two decimal places.

\paragraph{4. TeXstudio A-01: Angle insertion.}
The task asks the agent to measure an angle between specified atoms in ChimeraX and replace a placeholder in the article. The evaluator checks the saved file and compilation result, but these requirements were omitted from the instruction. We explicitly require saving the TeX file and ensuring successful compilation.

\paragraph{5. TeXstudio A-02: Residue-name insertion.}
The task asks for the name of residue \texttt{/A:6} to be inserted into a table. We add the previously unstated requirements to save the TeX file and ensure that it compiles successfully.

\paragraph{6. TeXstudio A-03: DNA sequence correction.}
The task asks the agent to correct the Chain A DNA sequence using ChimeraX. We explicitly require saving the corrected TeX file and successful compilation, matching the conditions checked by the evaluator.

\paragraph{7. TeXstudio A-04: Fibonacci value correction.}
The task asks the agent to obtain the Fibonacci value $f(24)$ in KAlgebra and replace an incorrect value in the article. The revision adds saving and successful compilation as explicit completion requirements.

\paragraph{8. TeXstudio B-01: Lean 4 citation.}
The task asks the agent to fill a citation in \texttt{main.tex} using the existing Lean 4 entry in \texttt{refs.bib}, without modifying the bibliography file. We preserve these constraints and add the requirements to save the TeX file and ensure successful compilation.

\paragraph{9. TeXstudio B-02: COSPAR identifier and citation.}
The task asks for Gemini 9A's COSPAR identifier followed immediately by a Celestia citation. We clarify that the completed TeX file must be saved and compile successfully.

\paragraph{10. TeXstudio B-03: Province count and citation.}
The task asks the agent to insert the number of provinces in Piemonte and a GRASS GIS citation into the article. We add the missing requirements to save the TeX file and ensure successful compilation.

\paragraph{11. TeXstudio C-01: Atom-count correction.}
The task asks the agent to correct a C, O, or N atom count using ChimeraX. We explicitly require saving the corrected TeX file and successful compilation.

\paragraph{12. TeXstudio C-02: Single-letter correction.}
The task asks the agent to correct a typo in the deduction rules of the $\lambda C$ type system by changing exactly one letter. We retain this restriction and add saving and successful compilation to the instruction.

\paragraph{13. TeXstudio C-03: Figure-path repair.}
The task asks the agent to repair the incomplete figure path \texttt{figs/lat}. We add the requirements to save the TeX file and verify successful compilation.

\paragraph{14. TeXstudio C-04: Figure width and completion requirements.}
The original English instruction misspells ``width'' as ``with.'' We correct the typo, retain the restriction to change only the width of \texttt{figs/envelope} to 0.48, and explicitly require saving and successful compilation.

\paragraph{15. TeXstudio C-05: Lean code correction.}
The task asks the agent to correct Lean 4 code in the article by changing exactly one word, with the supplied VS Code available for debugging. We correct ``exact one word'' to ``exactly one word'' and add the requirements to save the TeX file and ensure successful compilation.

\subsection{Task \& Answer Errors}
\label{app:verified_content}

These repairs correct task statements, reference answers, labeled test points, and expected results encoded in evaluators, including mathematical statements with insufficient assumptions.

\paragraph{16. KAlgebra C-07: Evaluating the wrong derivative.}
The task asks for the derivative of $f(x)=\ln(x+1)+x^2-x$, but the original evaluator uses labels corresponding to the derivative of $\ln(x^2+1)+x^2-x$. We rebuild the test points using
\[
    f'(x)=\frac{1}{x+1}+2x-1,
\]
with positive examples inside the original function's domain.

\paragraph{17. KAlgebra C-12: Incorrect positive points for a distance-product curve.}
The product of the Euclidean distances to $(1,0)$ and $(-1,0)$ must equal 16, giving the implicit equation
\[
    \bigl((x-1)^2+y^2\bigr)\bigl((x+1)^2+y^2\bigr)=256.
\]
The original positive examples do not lie on this curve. We recompute their coordinates according to the stated condition.

\paragraph{18. KAlgebra C-14: Confusing diameter with radius.}
The stated diameter is $2\sqrt{2}$, so the radius is $\sqrt{2}$ and the sphere satisfies $x^2+y^2+z^2=2$. The original labels instead correspond to $x^2+y^2+z^2=8$, treating the diameter as the radius. We replace the test points to distinguish the requested sphere from the incorrect one.

\paragraph{19. Celestia C-09: Apparent-magnitude answer inconsistent with the software.}
The original reference answer is 4.47. In the benchmark's initial environment, sorting the galaxy list by distance and excluding the Milky Way selects Sagittarius dSph, whose displayed apparent magnitude is 3.84. We update the answer to 3.84 and specify that the answer must come from Celestia's Celestial Browser, excluding the Milky Way itself.

\paragraph{20. Celestia D-02: Eclipse date and answer format.}
The original answer gives December 21 for the last lunar eclipse of 2056; we correct it to December 22. The revised instruction also specifies the UTC date at maximum eclipse as seen from Earth and requests only the numeric day of December, removing ambiguity about the event time and submission format.

\paragraph{21. ChimeraX C-04: Incorrect hydrogen count.}
The task asks for the number of hydrogen atoms in \texttt{/A:1}. We correct the reference answer from 12 to 0 and update the evaluator accordingly.

\paragraph{22. Lean 4 A-02: Overly strong Ackermann inequality.}
The original statement requires $\operatorname{ack}(x,y+1)<\operatorname{ack}(x+1,y)$. Equality can occur when $y=0$, so the strict inequality is false. We replace $<$ with $\leq$.

\paragraph{23. Lean 4 C-05: Incorrect set-containment assumptions.}
The original assumptions $A\subseteq Y$ and $Y\subseteq A^{\mathsf c}$ do not imply that $Y$ is compact. For example, when $A=\emptyset$, they permit noncompact closed sets $Y$. We replace these containment conditions with $Y\subseteq A$, retaining compactness of $A$ and closedness of $Y$, so that the statement expresses compactness of a closed subset of a compact set.

\paragraph{24. Lean 4 D-02: Missing convexity assumption.}
The original statement tries to derive uniform continuity on $S$ from a bounded derivative on $S$, without conditions connecting its points. We add \texttt{Convex} $\mathbb{R}\;S$, allowing the derivative bound to control function differences through a mean-value inequality and support the uniform-continuity conclusion.

\paragraph{25. Lean 4 D-07: Moment condition does not express integrability.}
The original statement binds $X$ twice and uses an equality between an expectation and some real number $M$ without explicitly requiring the necessary integrability. We remove the duplicate binding and assume that $\omega\mapsto |X(\omega)|^p$ is integrable with respect to $\mathbb P$, supporting the tail conclusion
\[
    x^p\,\mathbb P\bigl(|X|>x\bigr)\longrightarrow 0
    \qquad (x\to+\infty).
\]

\paragraph{26. Lean 4 E-03: Hamiltonian endpoints and edge bound.}
The original hypothesis requires a Hamiltonian walk between every pair of vertices, including identical endpoints. We restrict it to distinct vertices. The extra constant term $+1$ in the original conclusion is also invalid: $K_4$ satisfies the revised hypothesis but has $2e=3n$. We therefore correct the conclusion to $2e\geq3n$ for $n\geq4$.

\paragraph{27. GRASS GIS D-01: Incorrect expected schema type.}
The task requests a \texttt{COLOR varchar} column, which the current GRASS GIS and SQLite environment reports as \texttt{TEXT} through \texttt{v.db.select}. The original evaluator instead expects \texttt{CHARACTER}. We retain the requested \texttt{varchar} declaration and change the expected schema type to \texttt{TEXT}. The evaluator already supports both GUI and Console execution; this repair corrects the expected result.

\subsection{Evaluator Bias}
\label{app:verified_evaluators}

These repairs remove unnecessary evaluator restrictions on execution paths, output representations, and output locations while preserving the task requirements. For GRASS GIS B-03--B-10, we verified that both GUI operations and Console commands achieve the required outcomes. However, the original evaluators rely on GUI property caches and reject equivalent Console executions when those caches are not populated. We extend the checks to recognize both paths. TeXstudio B-04 addresses overly strict matching of BibTeX title capitalization and formatting. GRASS GIS B-12 extends output checking to a separate statistics window. Tasks that explicitly require a GUI state change retain checks of that state.

\paragraph{28. GRASS GIS B-03: Category and topology display.}
The original evaluator reads \texttt{display} from the layer-properties cache. An equivalent \texttt{d.vect} command can produce the requested display without populating that cache. We inspect either GUI parameters or layer command state for both \texttt{cat} and \texttt{topo}, excluding states marked hidden or unchecked.

\paragraph{29. GRASS GIS B-04: Area fill color.}
The task requires the fill color $(191,191,191)$. In addition to GUI parameters, we inspect \texttt{fill\_color} or \texttt{fcolor} in the layer's \texttt{d.vect} command. The evaluator accepts RGB notation and the equivalent RGBA notation with an opaque alpha value.

\paragraph{30. GRASS GIS B-05: Road line color.}
The task requires white lines for \texttt{roadsmajor}. We inspect GUI parameters or layer command state and accept \texttt{white}, \texttt{255:255:255}, and the equivalent opaque RGBA representation.

\paragraph{31. GRASS GIS B-06: Railway line width.}
A Console command can set the \texttt{railroads} line width to 2 while leaving the GUI parameter cache empty. We add checks of layer command state and accept both \texttt{2} and \texttt{2.0}.

\paragraph{32. GRASS GIS B-07: Fire-station symbol.}
The task requires the \texttt{basic/marker} symbol for \texttt{firestations}. We check the \texttt{icon} parameter in either GUI properties or the layer's \texttt{d.vect} command, allowing an equivalent Console result to be recognized.

\paragraph{33. GRASS GIS B-08: Hospital symbol size.}
The task requires symbol size 10 for \texttt{hospitals}. We additionally inspect the command's \texttt{size} parameter and accept both \texttt{10} and \texttt{10.0}, removing dependence on editing the GUI property panel.

\paragraph{34. GRASS GIS B-09: School filtering condition.}
The task requires displaying schools satisfying the wheelchair-access condition. We inspect \texttt{where} in either GUI parameters or layer command state and remove whitespace before comparison with \texttt{wheelchair='yes'}.

\paragraph{35. GRASS GIS B-10: Label configuration.}
The task requires the \texttt{boundary\_region} label field \texttt{NOME}, font \texttt{Ubuntu:Bold}, and horizontal alignment \texttt{center}. Console execution can produce the requested display while leaving the GUI cache empty. We check all three settings in either GUI parameters or the same layer command, recognizing both completion paths.

\paragraph{36. GRASS GIS B-12: Statistics in a separate output window.}
The original evaluator searches only \texttt{dump['output']} for \texttt{n: 263975}. We retain that check and add a fallback for a separate \texttt{r.univar} window. The fallback uses the accessibility tree to identify the target map and selected Command output tab, then checks the corresponding image region for rendered output, covering cases where the output is absent from the original field.

\paragraph{37. TeXstudio B-04: Overly strict BibTeX title matching.}
The task asks the agent to obtain a paper's BibTeX from Google Scholar and write it to \texttt{refs.bib}. The original evaluator requires exact title capitalization and spacing around the equals sign, rejecting valid entries for the same paper with equivalent textual representations. We make title matching case-insensitive and allow whitespace around the equals sign. The instruction also explicitly requires saving \texttt{refs.bib}.

\subsection{Environment Repairs}
\label{app:verified_environment}

This category covers repairs to the evaluation environment, including application-usage prompts, answer-submission protocols, action-response handling, compilation, and initialization settings. It also records the removal of one task requiring an unavailable external account. These shared repairs are not counted as additional task revisions.

\paragraph{38. ChimeraX D-03: External account required for AlphaFold.}
The task asks for an AlphaFold structure prediction from a supplied amino-acid sequence. Completion requires an external Google account that is not provided by the benchmark environment. We remove this task from the final evaluation set.

\paragraph{Answer-submission protocol (32 tasks).}
We observed correct answers being rejected because their response format was not correctly handled by the framework. The original prompt did not sufficiently specify the code-block structure required for \texttt{ANS} submissions. This protocol issue affects all 32 answer-submission tasks. We explicitly require \texttt{ANS} followed by the actual answer inside a separate fenced code block, with neither the answer on the fence line nor additional text outside the block.

\paragraph{Application-usage prompts.}
The original \texttt{LEAN\_VM\_USAGE} incorrectly contains TeXstudio document-editing instructions, while \texttt{TEXSTUDIO\_VM\_USAGE} is undefined. We provide the correct Lean 4 instructions: complete the target proof or code in VS Code, save the file, and use Lean feedback to check for errors, warnings caused by \texttt{sorry}, and remaining goals. We also restore the TeXstudio-specific document-editing and compilation instructions.

\paragraph{Malformed action responses.}
The original framework does not validate unsupported tool or scratchpad control markers, so responses containing duplicated actions inside such envelopes can be parsed as ordinary action text. We reject these responses before emitting any actions and allow one retry with the same observation. The recovery prompt asks the agent to follow the existing action format without unsupported envelopes or duplicated draft and final content.

\paragraph{TeXstudio compilation environment.}
We repair the TeXstudio compilation environment so that compilation checks can run normally, preventing an environment failure from blocking task completion and evaluation.

\paragraph{GRASS GIS initial pointer position.}
At initialization, the mouse pointer originally obscured the Console area, interfering with observation and evaluation. We reposition it near the center of the interface so that the Console remains unobstructed.

\paragraph{Update notifications.}
An update dialog appeared after every initialization. Dismissing it consumed interaction steps and materially interfered with evaluation by using part of the task's step budget on an unrelated action. We disable updates to prevent these recurring dialogs and their associated evaluation overhead.
\par
\endgroup

\section{Task Construction and Deduplication}
\label{app:task_deduplication}

We construct 108 additional tasks across the six scientific applications for skill self-evolution. These tasks use the same types of software operations as the test tasks to support the transfer of operational experience, while varying the target objects, task parameters, or specific requirements to form distinct tasks. We manually compare each additional task against the ScienceBoard-Verified tasks, examining its goal, initial state, required output, and evaluation criteria to confirm that it does not duplicate a test task. Researchers in the corresponding scientific domains who are familiar with the respective software manually review and execute all additional tasks to verify that their requirements are clear, their reference answers or target states are correct, and they can be completed in the specified software environments.

\section{Execution-Guided Skill Self-Evolution: Additional Details}
\label{app:self_evolution}

This appendix specifies the execution protocol used in Section~\ref{sec:evolution}. Evolution uses only $E=\mathrm{Expanded}\setminus\mathrm{Verified}$ and the original resources $\mathcal{D}$. For new-knowledge creation, $\mathcal{D}$ is accessed through bounded live browsing that starts from the configured official documentation root for the current application and follows only in-scope links discovered from fetched official pages. It does not use a search engine, BM25 retrieval, a saved Resource-to-Skill snapshot, or model knowledge as source evidence. The execution model is fixed, and $V=\mathrm{Verified}$ is reserved for final evaluation.

\subsection{Frozen Initial Library and Per-Task State}
\label{app:evolution_library}

All evolution tasks start from the same frozen initial library $\mathcal{L}_0=(\mathcal{S}_0,\mathcal{R}_0)$. A task never installs a provisional update into the library used by another task. Instead, it maintains a private execution history and may contribute at most one independently validated candidate to a global candidate pool $\mathcal{P}$. Candidates are consolidated only after all tasks terminate.

For each task $q$, its first task exploration selects a fixed routed set $\mathcal{F}_q$ of Skills and References from $\mathcal{L}_0$. Later task explorations use the same routed names and initial versions, even after the task has produced a failed candidate. This separates behavioral exploration from candidate installation: rejected knowledge remains evidence, not active guidance.

Each candidate contains one complete target entry relative to $\mathcal{L}_0$. A candidate produced after a failed exploration may revise one entry in $\mathcal{F}_q$ or add one Skill or Reference supported by $\mathcal{D}$. A candidate distilled after a later successful exploration may only revise one existing routed entry. A rejected candidate is never used as the textual base of a later candidate.

\subsection{Complete Execution History and Input Separation}
\label{app:evolution_history}

Each task has an indexed history archive. For every task exploration and candidate validation, the archive records the task instruction, execution type, exact execution-time prompt, raw action--observation trajectory, evaluator outcome, routed Skill and Reference names, their complete execution-time contents and version hashes, and every screenshot mapped to its trajectory step. When applicable, it also records the diagnosis and the complete candidate entry used in validation. Screenshots are interpreted as pre-action observations: the effect of an action is checked in the following screenshot.

The first task exploration receives no history. Every later task exploration receives all earlier task explorations and candidate validations for that task, including their exact prompts, raw trajectories, diagnoses, candidates, outcomes, and screenshots. The records are explicitly labeled so that the agent can distinguish exploration from validation, identify which knowledge version was loaded, and map each screenshot to its execution and step. All historical screenshots are placed in the first model turn and marked as evidence; the final live screenshot remains the state on which the agent must act. The prompt instructs the agent to inspect the complete history, avoid repeating failed action sequences, and not follow initial Skills or References mechanically when current GUI evidence suggests a version mismatch.

Diagnosis and knowledge writing receive the task, loaded entries, exact prompt, raw trajectory, evaluator result, and complete screenshot sequence. The analysis agent must inspect every screenshot before submitting its result. Candidate validation is deliberately separated: it receives no exploration history, successful trajectory, diagnosis, or candidate rationale. This prevents the validation agent from succeeding through transient trajectory memory rather than through the proposed external knowledge.

\subsection{Task Exploration, Failure Diagnosis, and Candidate Validation}
\label{app:evolution_diagnosis}

Let $i\in\{1,\ldots,M_{\mathrm{explore}}\}$ index task explorations for task $q$. Each exploration begins from the standard initial environment in a fresh session:
\begin{equation}
    \tau_q^{(i)}
    =\operatorname{Execute}_{\theta}
      \bigl(q;\mathcal{L}_0,\mathcal{F}_q,\mathcal{H}_q^{(i)}\bigr),
    \label{eq:evolution_exploration}
\end{equation}
where $\mathcal{H}_q^{(1)}=\emptyset$ and $\mathcal{H}_q^{(i)}$ for $i>1$ contains all earlier exploration and candidate-validation records for $q$. Routing is performed normally only for $i=1$; its result defines $\mathcal{F}_q$ for subsequent exploration and validation.

If $\tau_q^{(1)}$ succeeds, the task exits without proposing an update. A successful exploration with $i>1$ enters the submission process in Appendix~\ref{app:evolution_submission}. Each failed exploration is analyzed for one primary root cause. Exhausting the action budget, stopping early, or failing the evaluator describes the observed outcome rather than the underlying cause. The diagnosis uses exactly three categories:
\begin{enumerate}
    \item \textbf{Grounding failure} (\texttt{grounding\_failure}). Incorrect interface localization, targeting, state reading, or manipulation is the principal obstacle. Repeated grounding mistakes that consume the action budget can qualify. The counterfactual criterion is whether correcting those grounding mistakes would remove the principal obstacle; if an independent knowledge deficiency would remain, a knowledge category must be selected instead.
    \item \textbf{Modify existing knowledge} (\texttt{modify\_existing}). One loaded Skill or Reference is relevant but contains an erroneous or incomplete procedure, omitted condition, inappropriate parameter, unsupported syntax, or version-dependent instruction. The target must be one entry in $\mathcal{F}_q$.
    \item \textbf{Add new knowledge} (\texttt{add\_new\_knowledge}). The routed entries do not contain the capability or operational knowledge needed for the task. This category is reserved for a genuine coverage gap rather than a minor defect in a relevant entry. The new Skill or Reference must be supported by task-relevant content in the original resources $\mathcal{D}$; without sufficient source support, no candidate is produced.
\end{enumerate}

Diagnosis and conditional candidate writing occur in the same analysis session. The diagnosis step fixes the category, evidence, and proposed repair. For \texttt{modify\_existing}, it also fixes the type and name of one routed target, after which the next step writes its complete replacement while retaining access to the same trajectory and screenshots. For \texttt{add\_new\_knowledge}, the diagnosis deliberately leaves the target type and name unspecified and describes only the missing reusable knowledge scope. Only after this diagnosis is accepted does the same session unlock bounded live access to $\mathcal{D}$; the research step determines whether the supported artifact is a Skill or Reference, chooses its stable title, and writes one complete new entry. If the fetched official material is insufficient, no candidate is produced. A grounding diagnosis likewise produces no candidate and therefore requires no candidate validation. Either no-candidate outcome consumes the current exploration attempt. If another exploration remains, the task continues with its accumulated history. Even after the final allowed exploration fails, its diagnosis and, when applicable, its one candidate validation are still completed.

\subsubsection{Independent Candidate Validation}
\label{app:evolution_validation}

For a candidate $\Delta_q^{(i)}$, validation constructs a temporary library
\begin{equation}
    \widetilde{\mathcal{L}}_q^{(i)}
    =\mathcal{L}_0\oplus\Delta_q^{(i)},
    \label{eq:evolution_temporary_library}
\end{equation}
where $\oplus$ replaces the target's initial version for a revision or inserts one source-supported entry for an addition. The validation agent starts from the task's standard initial environment in a fresh session. It force-loads the original entries named by $\mathcal{F}_q$; for a revision, the candidate replaces its corresponding original entry, and for an addition, the candidate is force-loaded in addition to all original routed entries. Thus every attempt contains at most one modified or new target.

The validation trajectory is
\begin{equation}
    \widetilde{\tau}_q^{(i)}
    =\operatorname{Execute}_{\theta}
      \bigl(q;\widetilde{\mathcal{L}}_q^{(i)},\widetilde{\mathcal{F}}_q^{(i)},\emptyset\bigr),
    \label{eq:evolution_validation}
\end{equation}
where the empty history excludes all prior trajectories, screenshots, diagnoses, and rationales. The task evaluator alone determines success. A successful candidate is added to $\mathcal{P}$ and the task terminates. A failed candidate, its complete contents, and its validation record are archived; if another task exploration remains, that exploration again uses $\mathcal{L}_0$ and $\mathcal{F}_q$, with the failed validation supplied only as history.

\subsection{Submission After History-Guided Success}
\label{app:evolution_submission}

A task exploration that succeeds at $i>1$ may have relied on failure history unavailable at deployment. The submission agent therefore contrasts the successful exploration with all earlier failed explorations, candidate validations, loaded knowledge versions, and candidate contents. In one response it either concludes that no knowledge change is warranted or writes a complete revision to one existing entry in $\mathcal{F}_q$. This process cannot create a new Skill or Reference.

At most $M_{\mathrm{submit}}$ submission attempts are allowed. Every revised candidate is generated from the target's version in $\mathcal{L}_0$, not by editing the preceding rejected submission candidate. On later attempts, the submission agent additionally receives all previous submission candidates and their validation trajectories so that it can avoid repeating an ineffective revision. Each candidate is validated through the history-free protocol in Equation~\ref{eq:evolution_validation}. The first successful candidate enters $\mathcal{P}$ and ends the task.

When the submission agent attributes a failed validation to grounding, it may retain the same candidate and validate it again; the unchanged retry consumes one submission attempt. If no candidate has yet been proposed, it may instead validate the unchanged initial routed knowledge. Success of this unchanged validation ends the task without adding a candidate. If all $M_{\mathrm{submit}}$ attempts fail, the task terminates without an update and does not return to task exploration.

\subsection{Candidate Consolidation and Final Evaluation}
\label{app:evolution_consolidation}

After every evolution task terminates, accepted candidates are grouped by target Skill or Reference. Revisions are grouped by the identity of their original entry. For additions, a conservative Grok semantic-deduplication pass groups candidates only when they have the same knowledge type and express the same reusable capability at substantially the same granularity, scope, and intended use; Skills and References are never grouped together, and uncertain candidates remain separate. Each group adopts the title of one independently validated member rather than inventing a new title. If this auxiliary grouping call is technically unavailable or violates its structured-output contract, the pipeline fails open to exact type-and-title grouping, thereby avoiding an unsupported merge. Different targets are consolidated independently and are never mixed merely because they came from the same task.

If a target has exactly one accepted candidate, that already validated candidate is adopted directly without another task execution. If a target has multiple candidates, the consolidation model receives every accepted candidate and either the target's complete initial content for a revision or an explicit new-entry baseline for an addition. It then produces one complete entry that preserves the independently validated, non-conflicting improvements of all candidates. The merged text is not allowed to discard a candidate-specific capability merely to simplify the entry.

Because merging can change individually validated content, a multi-candidate consolidation must pass regression certification. Let $Q_g$ contain all tasks that contributed candidates to consolidated target $g$. The same consolidated content and hash are installed for every replay, and each $q\in Q_g$ is executed from its standard initial environment in a fresh, history-free session with its first-exploration routing fixed. The initial full replay over $Q_g$ does not consume the reflection budget.

If every replay succeeds, the target is certified. Otherwise, a reflection agent receives the initial entry, all accepted candidates, the current consolidation, and every regression prompt, raw trajectory, evaluator outcome, and screenshot. It chooses between two actions:
\begin{enumerate}
    \item \textbf{Revise.} If the consolidated content caused or contributed to any failure, a complete replacement is regenerated from the initial entry and all accepted candidates rather than edited cumulatively from the failed consolidation. Previously passed results are invalidated, and all tasks in $Q_g$ are replayed.
    \item \textbf{Keep.} The current consolidation may be retained only when every current failure is attributable to grounding and the content itself requires no change. Passed tasks remain certified, and only the failed tasks are replayed.
\end{enumerate}

Each reflection followed by its required replay consumes one of at most $M_{\mathrm{consolidation}}$ cycles. Infrastructure or API failures are handled separately and are not treated as task failures. If certification still fails after the allowance is exhausted, target $g$ is rejected: a revised target retains its initial version, while an uncertified addition is omitted. After all target groups are resolved, $\mathcal{L}_{\mathrm{final}}=\operatorname{Consolidate}(\mathcal{L}_0,\mathcal{P})$ is frozen and evaluated on $V$ with standard initial environments, fresh sessions, normal routing, and no evolution history. No additional global regression is performed.

\subsection{Stopping Conditions and Budgets}
\label{app:evolution_stopping}

The default configuration uses
\begin{equation}
    M_{\mathrm{explore}}=3,\qquad
    M_{\mathrm{submit}}=3,\qquad
    M_{\mathrm{consolidation}}=3.
    \label{eq:evolution_budgets}
\end{equation}
Only task explorations consume $M_{\mathrm{explore}}$; their diagnosis, candidate writing, and one independent candidate validation do not. Only submission validations consume $M_{\mathrm{submit}}$. The initial regression of a multi-candidate consolidation does not consume $M_{\mathrm{consolidation}}$; each subsequent reflection-and-replay cycle does.

Excluding consolidation replay, a conservative execution bound is
\begin{equation}
    N_{\mathrm{exec}}
    \leq |E|\bigl(2M_{\mathrm{explore}}+M_{\mathrm{submit}}\bigr),
    \label{eq:evolution_budget}
\end{equation}
which counts at most one candidate validation after each task exploration and at most $M_{\mathrm{submit}}$ submission validations. This is intentionally loose because a task that enters submission cannot also fail all task explorations. For a multi-candidate consolidation group $g$, regression adds at most $(1+M_{\mathrm{consolidation}})|Q_g|$ task executions; keep decisions can reduce this by retaining already passed tasks. These bounds exclude final evaluation, diagnosis and writing calls, source retrieval, routing, and separately limited infrastructure retries.

Evolution outcomes and consolidation replays are process measurements and are not substituted for held-out evaluation results.

\clearpage
\section{Case Study: Complementary Roles of Skills and References in KAlgebra}
\label{app:case-kalgebra-c05}

KAlgebra task C-05 illustrates how procedural guidance and shared expression knowledge support different parts of one task. We compare three executions with autonomous routing: Skills and References together, Skills only, and References only.

\subsection{Task, Routing, and Selected Knowledge}

\paragraph{Task and initial state.} The task instruction is: \emph{\textquotedblleft Load script from \texttt{\textasciitilde/script.kal} where vector $a$ are defined. Let $b=(\sqrt{3},2)$; calculate their inner product in KAlgebra and store the result in variable \texttt{p}.\textquotedblright} The script contains \texttt{a := vector\{-1, root(3, 3)\}}. Loading it defines $a=(-1,\sqrt[3]{3})$, so the required scalar result is
\[
    p=a\cdot b=-\sqrt{3}+2\sqrt[3]{3}\approx 1.15244833305.
\]
Initially, $a$, $b$, and $p$ are undefined.

\paragraph{Run settings and routing.} All runs use Grok-4.6 (xhigh), the same initial environment, and five interaction turns; a turn may include multiple GUI actions or expression submissions. The model autonomously routes over both resource types in the joint condition. Each ablation disables one type while retaining autonomous routing over the other. The selected entries are listed below; the Skill-only run selects two additional Skills beyond S1.

\begin{center}
\small
\renewcommand{\arraystretch}{1.1}
\begin{tabularx}{\linewidth}{@{}lX@{}}
\toprule
\textbf{Condition} & \textbf{Autonomously selected resources} \\
\midrule
Skill + Reference & S1: script workflow; R1: function catalog; R2: expression syntax \\
Skill only & S1: script workflow; S2: execution mode; S3: latest-result reuse \\
Reference only & R1: function catalog; R2: expression syntax \\
\bottomrule
\end{tabularx}
\end{center}

\paragraph{S1: Save and rerun Calculator instruction sequences.} The relevant procedural instruction is:
\begin{quote}
\small
\textquotedblleft To execute the saved sequence later, choose Calculator $\rightarrow$ Load Script... (Ctrl+L) and select the script file. This executes the file's instructions sequentially.\textquotedblright
\end{quote}

\paragraph{R1: KAlgebra Built-in Function Catalog.} The catalog specifies \texttt{root(par1, par2)}, illustrated by \texttt{x->root(x, 2)}. It also lists the lowercase function name \texttt{scalarproduct}. These entries supply the application-specific names and call forms used by the joint run.

\paragraph{R2: KAlgebra Expression Syntax and Operators.} This Reference covers function calls, arithmetic, and variable definitions, including \texttt{:=} for assignment. Yet Skill-only uses \texttt{:=} correctly although none of its three selected Skills explains this syntax. The model already uses it without R2; these trajectories therefore do not establish an independent incremental contribution from R2.

\paragraph{Additional Skills in the Skill-only run.} S2, \emph{Choose KAlgebra's calculator execution mode}, explains selecting Calculator $\rightarrow$ Evaluate or Calculate before entering an expression. S3, \emph{Reuse the latest Calculator result with ans}, describes reusing the latest result through \texttt{ans}. Neither supplies the function catalog used by the joint run.

\newcommand{\CFiveShot}[2]{%
\begin{minipage}[t]{0.485\linewidth}
\vspace{0pt}
\centering
\includegraphics[width=\linewidth]{Figure/kalgebra_c05/#1}\par\vspace{4pt}
\raggedright\small #2
\end{minipage}}
\newcommand{\CFivePair}[4]{%
\par\smallskip\noindent\CFiveShot{#1}{#2}\hfill\CFiveShot{#3}{#4}\par\medskip}

\clearpage
\subsection{Single-Resource Trajectories: Different Execution Bottlenecks}

\paragraph{Reference only: the loading workflow remains unresolved within the budget.} The selected References provide expression knowledge but omit the script-loading workflow. The model first searches Session, then tries Calculator and Ctrl+O. Only after reopening Calculator does it locate Load Script. Its last turn clicks this command without selecting or loading the file, leaving $a$, $b$, and $p$ undefined.

\CFivePair{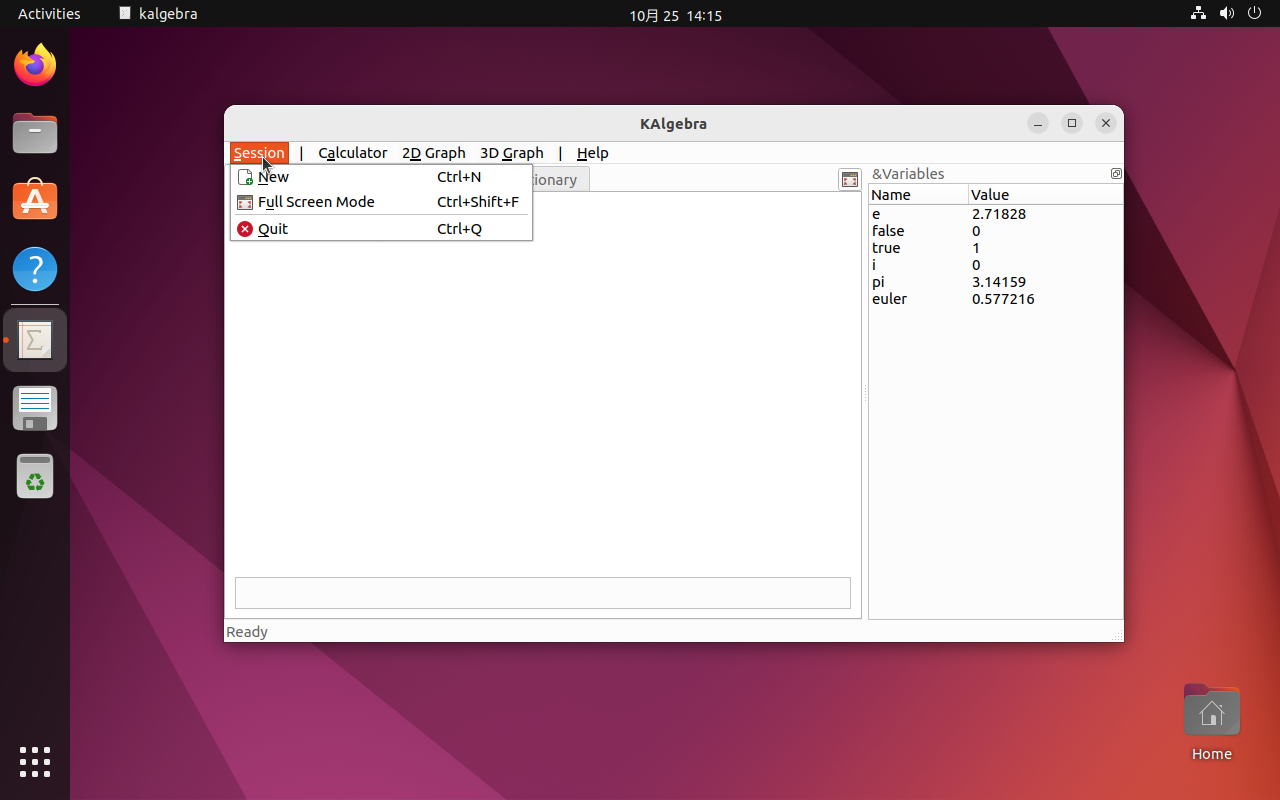}{\textbf{After turn 1.} Session has no script-loading command. The model next checks Calculator to find the loading entry.}{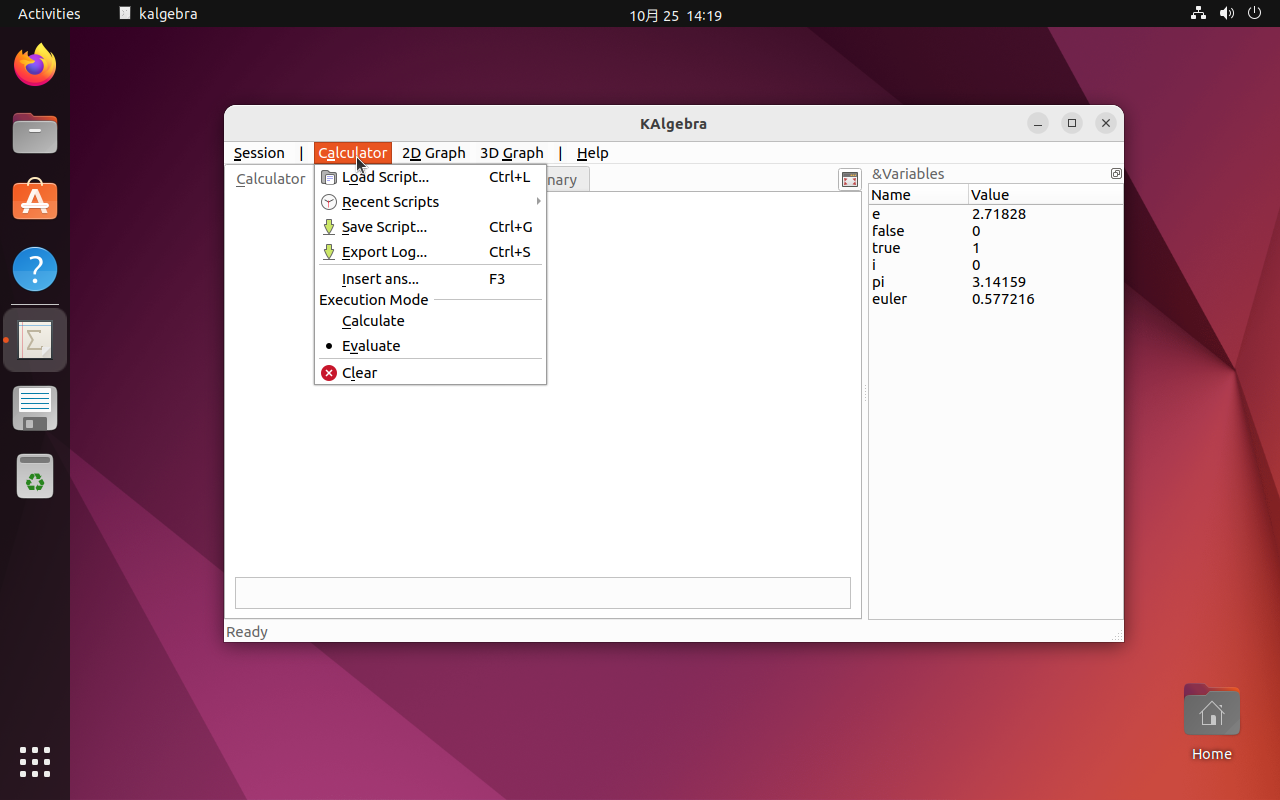}{\textbf{After turn 4.} Load Script is now visible. The model selects it in the last turn, but does not select or load the script within the budget.}

\paragraph{Skill only: loading succeeds, but the expressions remain symbolic.} With S1 available, the model follows Calculator $\rightarrow$ Load Script and loads $a$ by turn~3. It then enters \texttt{b:=vector\{sqrt(3), 2\}} and \texttt{p:=scalarProduct(a, b)}. Both \texttt{sqrt(3)} and \texttt{scalarProduct(a, b)} remain symbolic in the observed state. The model attributes this to Evaluate mode and attempts to select Calculate in its last turn, without resubmitting the expression. The final $p$ is still \texttt{scalarProduct(a, b)}, so evaluation fails.

\CFivePair{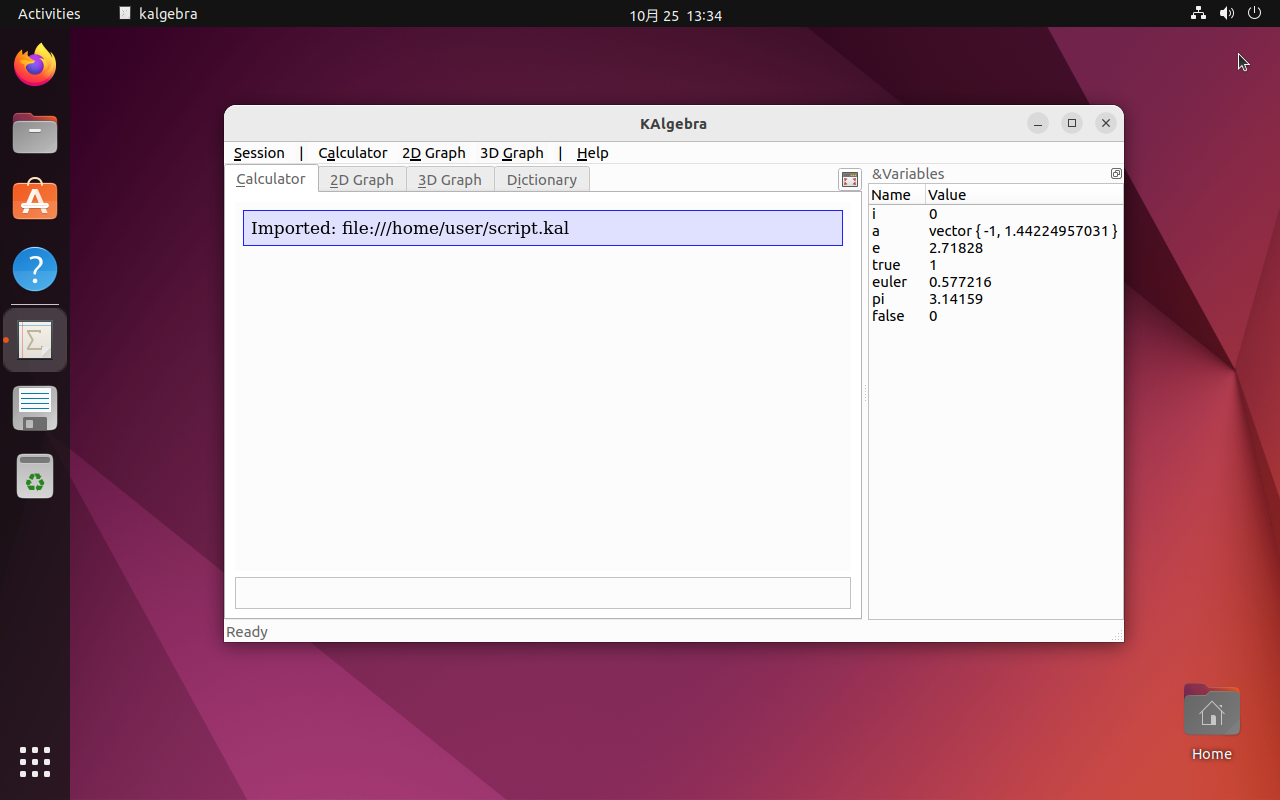}{\textbf{After turn 3.} The script has loaded and $a$ is available. The model next tries to define $b$ with \texttt{sqrt(3)} and compute $p$ with \texttt{scalarProduct(a, b)}.}{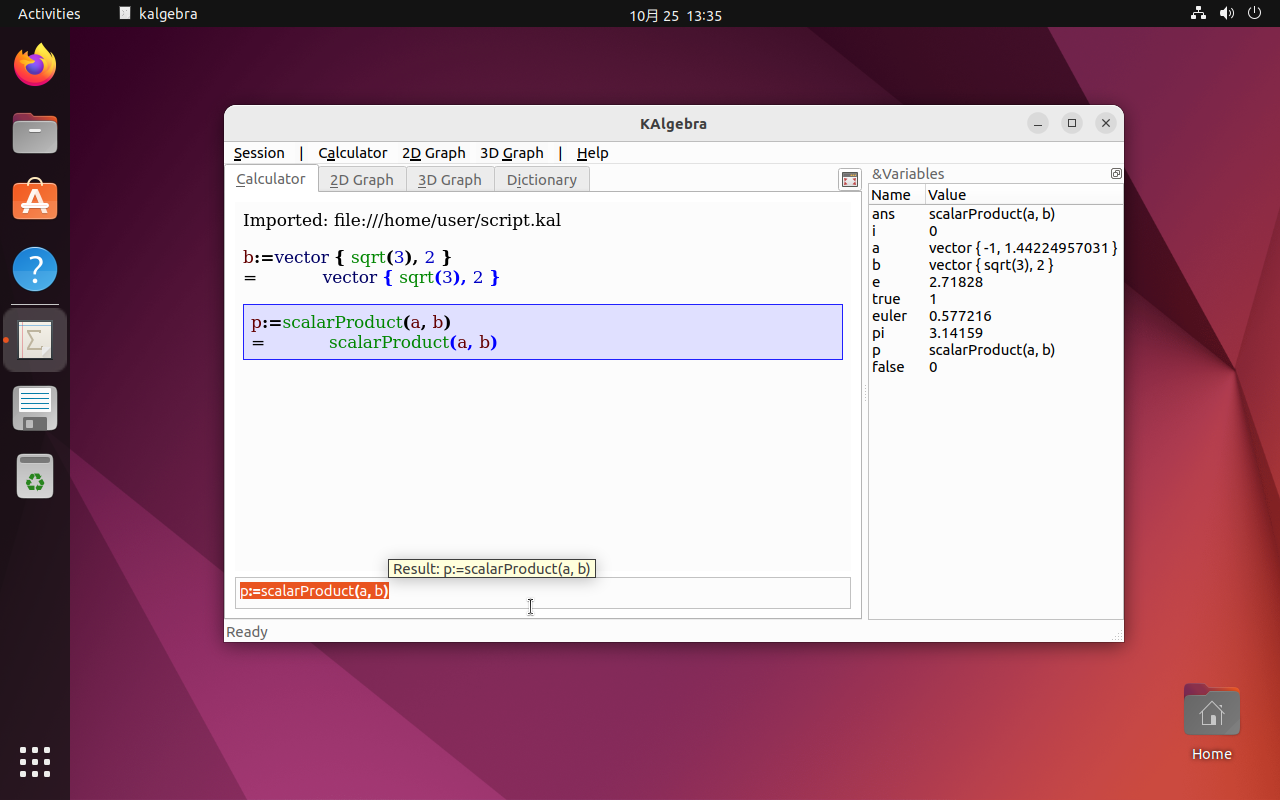}{\textbf{After turn 4.} The expressions remain symbolic. The model tries to address this by switching execution mode, rather than correcting the function calls.}

\clearpage
\subsection{Joint Trajectory: Combining the Workflow and Expression Knowledge}

The joint run selects S1, R1, and R2. Its script-loading steps follow S1, and the script is loaded by turn~3. In turn~4, it enters \texttt{b:=vector\{root(3, 2), 2\}}, producing $b=(1.73205080757,2)$, and attempts the inner product with \texttt{p:=scalarproduct(a, b)}. The numerical square root and lowercase function name match R1's function catalog.

\CFivePair{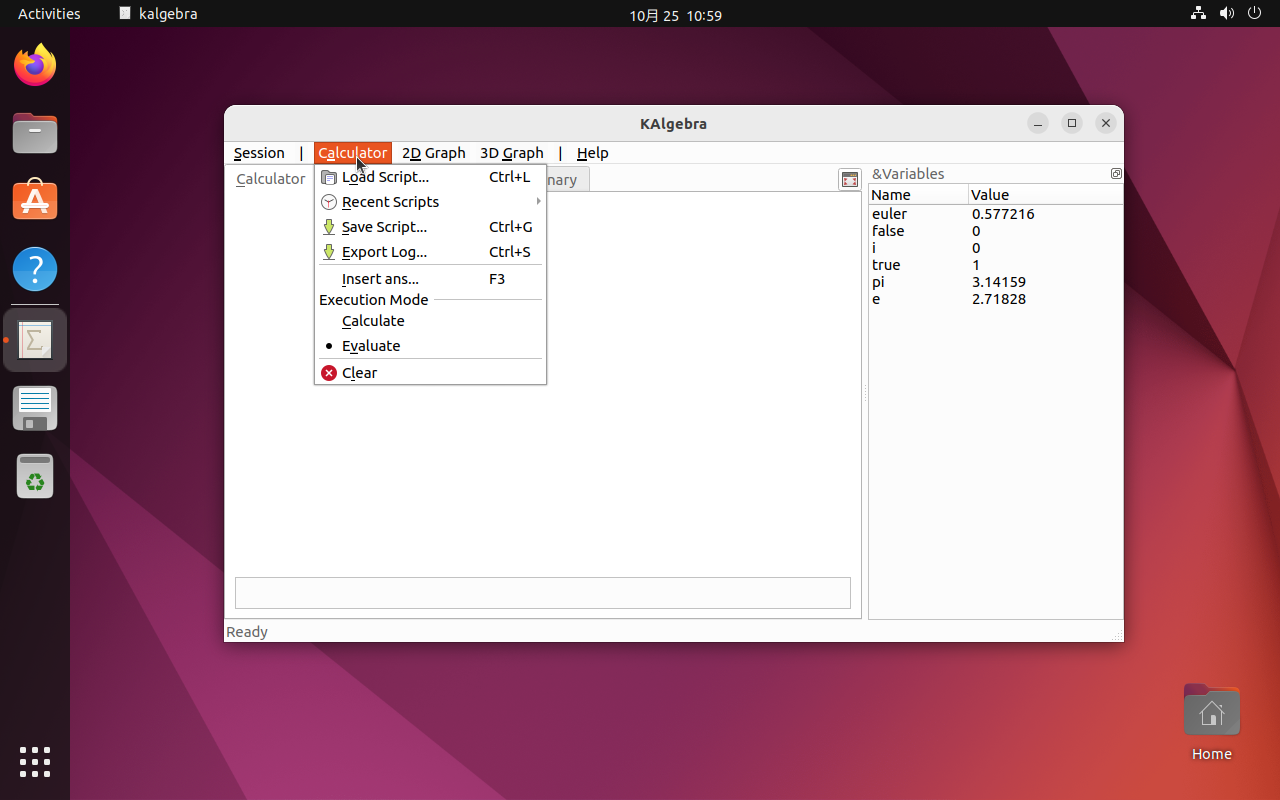}{\textbf{After turn 1.} Calculator exposes the Load Script command documented by S1. The model selects it to open the file-selection dialog.}{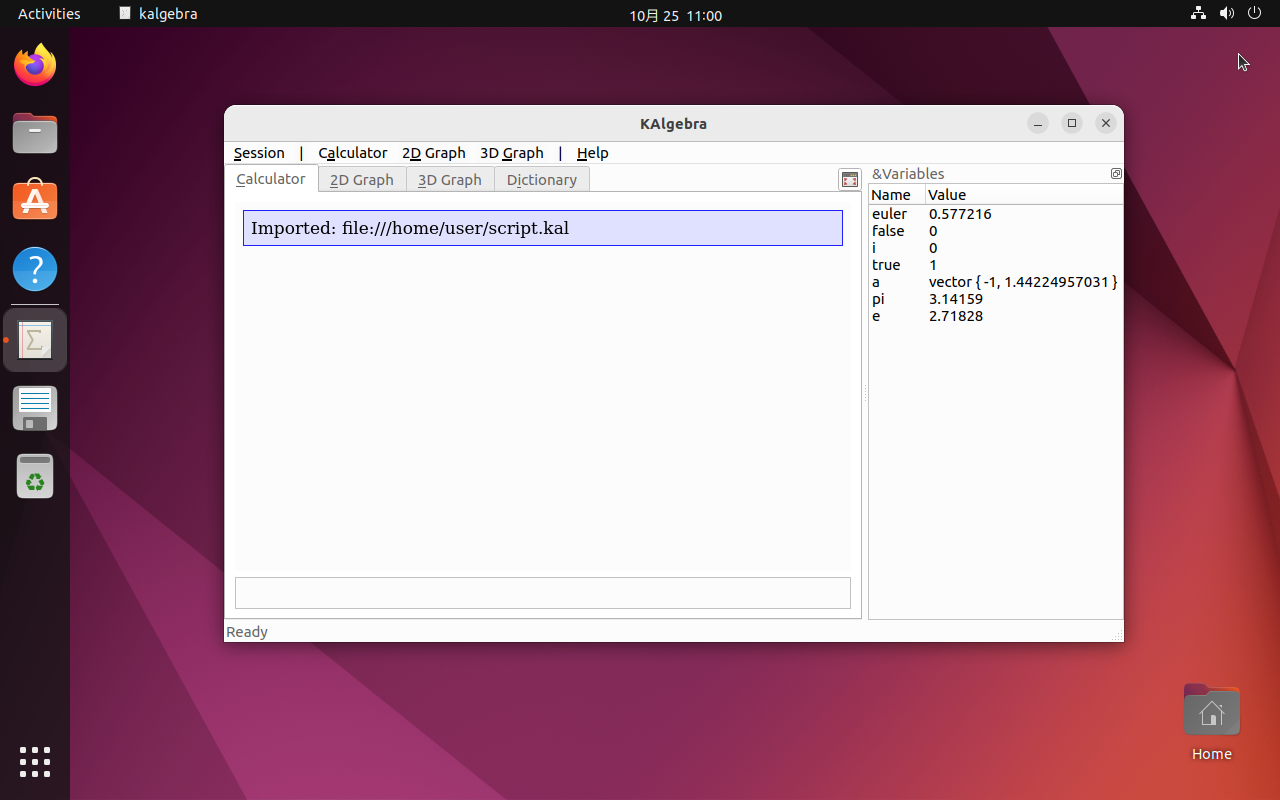}{\textbf{After turn 3.} The script has loaded and $a$ is available. The model next uses \texttt{root(3, 2)} and \texttt{scalarproduct} to define $b$ and attempt the inner product.}

\par\smallskip\noindent
\begin{minipage}[t]{0.485\linewidth}
\vspace{0pt}
\includegraphics[width=\linewidth]{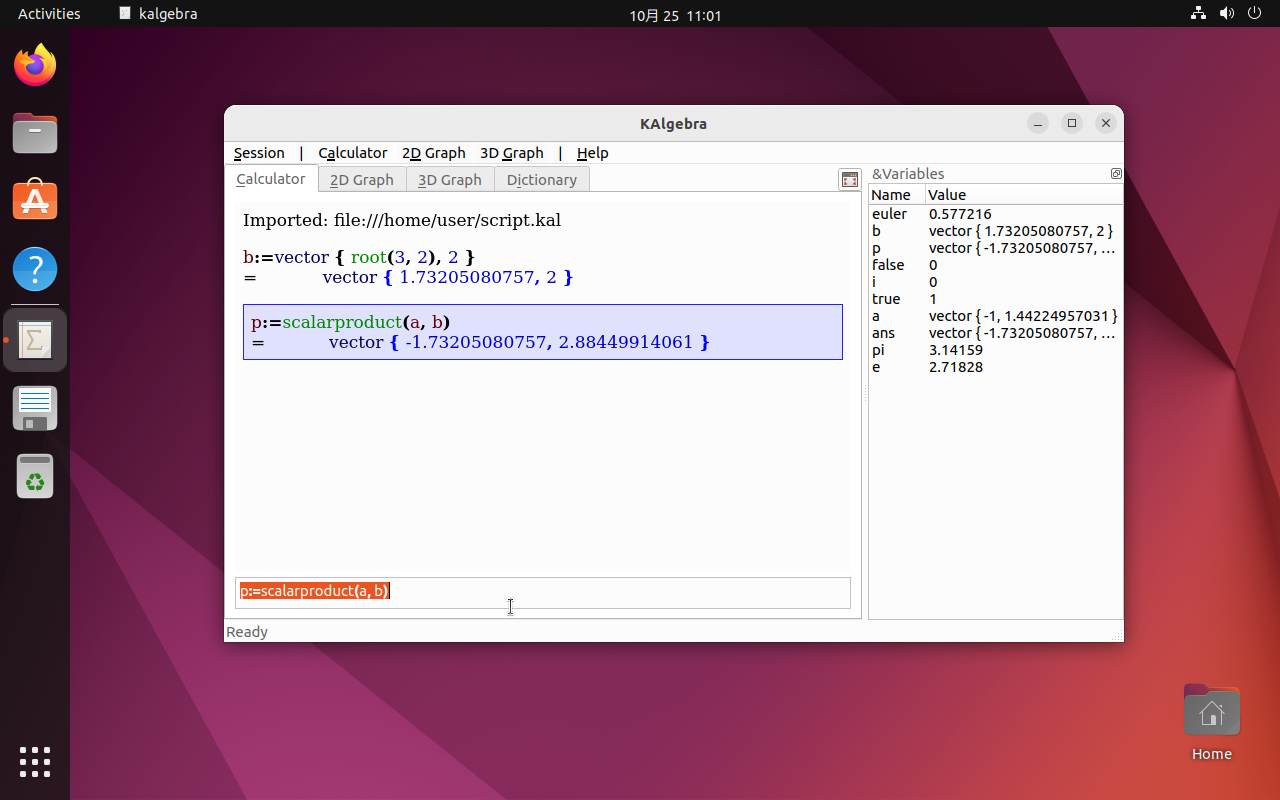}
\end{minipage}\hfill
\begin{minipage}[t]{0.485\linewidth}
\vspace{0pt}\small\raggedright
\textbf{Feedback correction.}\par\smallskip
\textbf{After turn 4.} $b$ is numerical, but $p=(-1.73205080757,2.88449914061)$ is a vector. The model recognizes the mismatch and, in turn~5, sums component products:
\par\smallskip
\texttt{p:=a[1]*b[1]+a[2]*b[2]}.
\par\smallskip
The resulting $p=1.15244833305$ passes evaluation. This correction follows execution feedback; the References do not explicitly prescribe it. The screenshot precedes this final correction.
\end{minipage}\par\medskip

\paragraph{Final outcomes.} Values below are recorded after turn~5; screenshots show earlier observations.
\begin{center}
\small
\renewcommand{\arraystretch}{1.15}
\begin{tabularx}{\linewidth}{@{}lXl@{}}
\toprule
\textbf{Condition} & \textbf{Final state} & \textbf{Evaluation} \\
\midrule
Reference only & $a$, $b$, and $p$ are undefined & Fail \\
Skill only & \texttt{b=vector\{sqrt(3),2\}}; \texttt{p=scalarProduct(a,b)} & Fail \\
Skill + Reference & $b=(1.73205080757,2)$; $p=1.15244833305$ & Pass \\
\bottomrule
\end{tabularx}
\end{center}

\paragraph{Complementary contributions.} The single-resource runs encounter different bottlenecks: completing script loading and obtaining the required numerical result. The observable complementarity in the joint run lies primarily in S1's loading procedure and R1's application-specific function knowledge, followed by execution feedback for the final correction. These trajectories are consistent with the aggregate ablation results in Section~\ref{sec:rq2}.
\clearpage

\section{Case Study: How Self-Evolve Repairs Resource--Environment Mismatches and Improves Task Performance}
\label{app:case-self-evolve-kalgebra}

We use KAlgebra surface-plotting tasks to examine how a mismatch between available documentation and the deployed software affects the initial Skills and References, and how Self-Evolve addresses it through execution feedback. Tasks B-14 and C-19 in Expanded--Verified expose the initial failures and support Skill revision. We then examine whether the frozen updated Skill helps previously failed tasks in the disjoint Verified set succeed. All runs use Grok 4.6 (xhigh).

\newcommand{\EvoImageScale}{0.90}
\newcommand{\EvoShot}[2]{%
\begin{minipage}[t]{0.485\linewidth}
\centering
\includegraphics[width=\EvoImageScale\linewidth]{Figure/kalgebra_self_evolve/#1}\par\vspace{1pt}
\raggedright\scriptsize #2
\end{minipage}}
\newcommand{\EvoPair}[4]{%
\par\smallskip\noindent\EvoShot{#1}{#2}\hfill\EvoShot{#3}{#4}\par\vspace{3pt}}
\newcommand{\EvoSingle}[2]{%
\par\smallskip\noindent\EvoShot{#1}{#2}\par\vspace{3pt}}

\subsection{Resource--Environment Mismatch: Why Self-Evolve Is Needed}
\label{app:case-self-evolve-mismatch}

Resource-to-Skill organizes documented procedures into callable Skills and References, but their applicability depends on whether the documentation describes the deployed environment. During resource collection, we could not retrieve documentation matching the deployed software version and used the relevant documentation available to us. The resulting resources therefore describe a different software version.

The collected KAlgebra handbook states: \emph{``Z cannot be defined yet. For the moment KAlgebra only supports 3D graphs explicitly dependent only on the x and y.''} The initial Skill, \emph{Plot an explicit x-and-y 3D surface}, incorporates this restriction as a precondition, while the associated 3D Graph Reference retains the documented guidance. As the later executions demonstrate, the deployed software also accepts three-variable mappings and implicit equations involving $z$.

\par\smallskip
\noindent\begin{minipage}{\linewidth}
\centering
\includegraphics[width=0.72\linewidth]{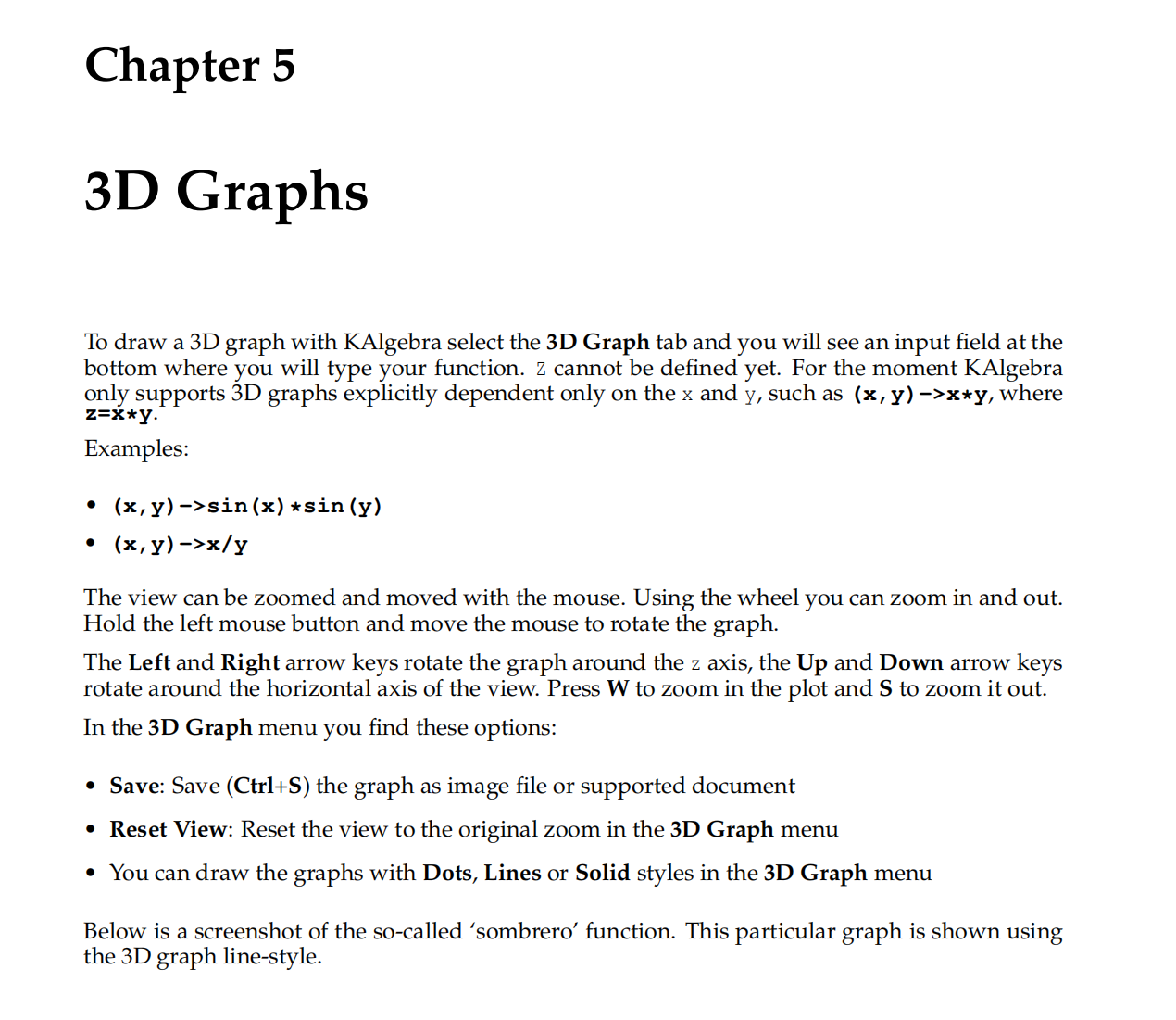}\par\smallskip
\raggedright\scriptsize\textbf{Source documentation.} The collected handbook restricts 3D plotting to explicit dependence on $x$ and $y$. This restriction is inherited by the initial resources but does not describe the deployed environment's full input capabilities.
\end{minipage}\par\smallskip

\paragraph{B-14: initial execution under the routed resources.} The task is to plot $2x^2+3y^2-z^2=1$. The model initially routes to the 3D-plotting Skill and two References: the function catalog and the 3D Graph reference. Following the explicit-only guidance, the first exploration enters \texttt{(x,y)->root(2*x\^{}2+3*y\^{}2-1,2)}. This selects only the positive root and plots one branch; the task fails. This initial exploration uses no failure history or revised Skill.

\EvoPair{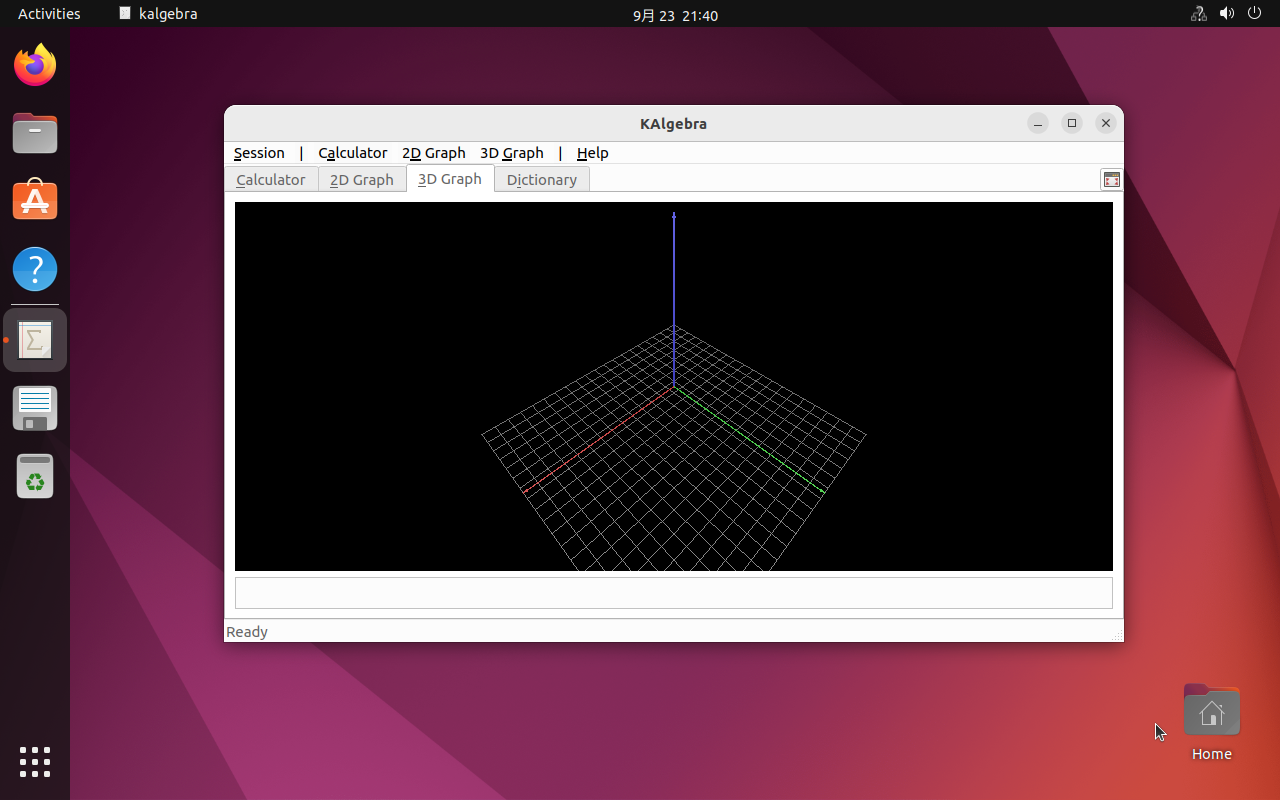}{\textbf{B-14, initial state.} The 3D Graph input is empty.}{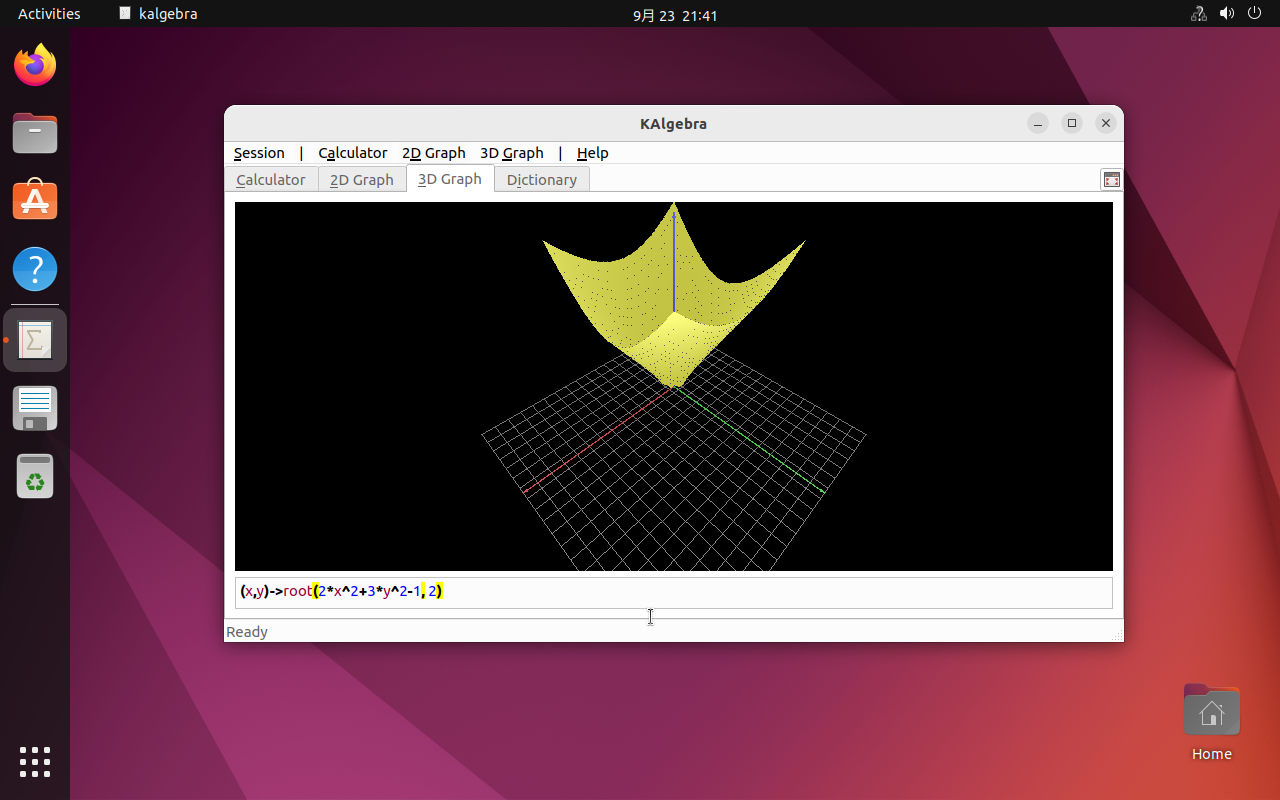}{\textbf{After entering the root mapping.} Only one branch is plotted; evaluation fails.}

\paragraph{C-19: initial execution under the routed resources.} The task is to plot a sphere with diameter endpoints $A(1,-2,3)$ and $B(5,2,-1)$. Its center is $(3,0,1)$ and its squared radius is $12$. The initial Skill and routed 3D Graph Reference likewise direct the first exploration toward an explicit mapping. The positive-root expression produces only the upper hemisphere. Zooming out does not change the incomplete surface, and the task fails. This exploration also precedes any failure-history input or Skill revision.

\EvoPair{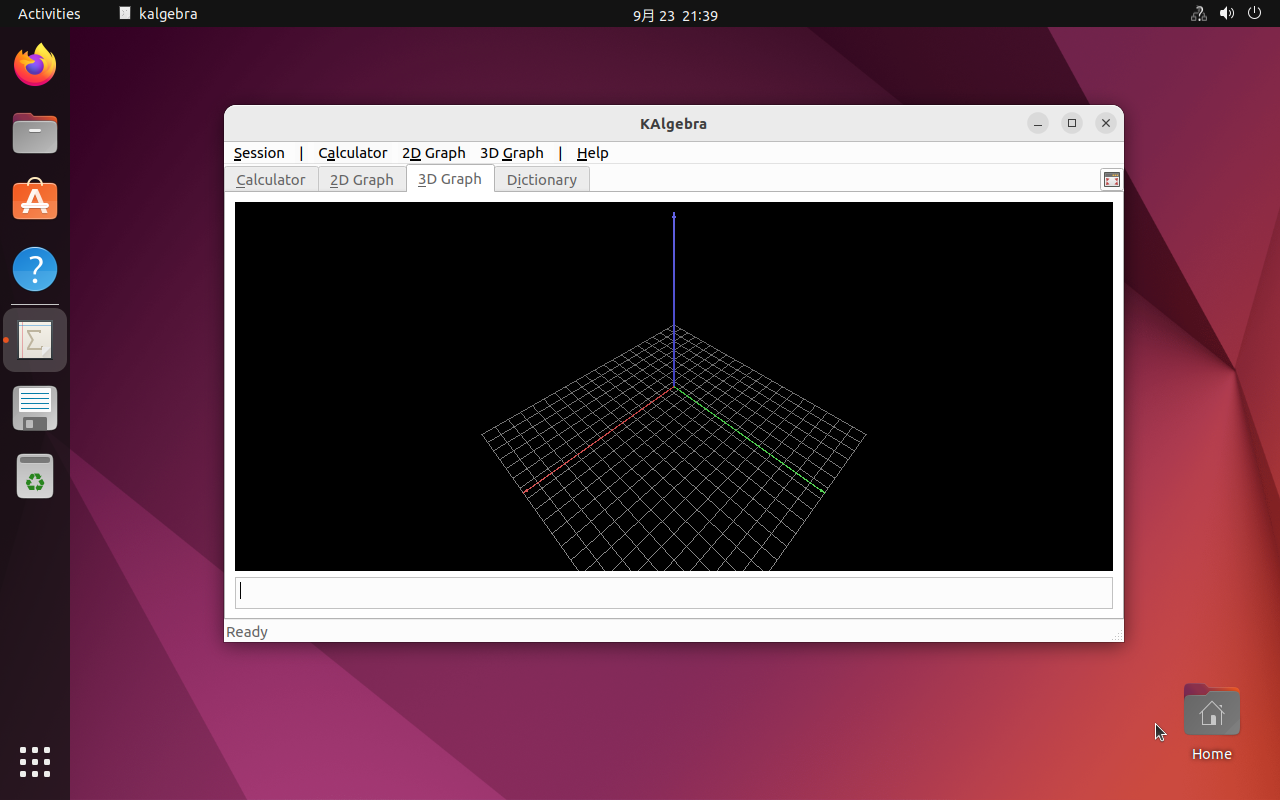}{\textbf{C-19, initial state.} The 3D Graph is empty.}{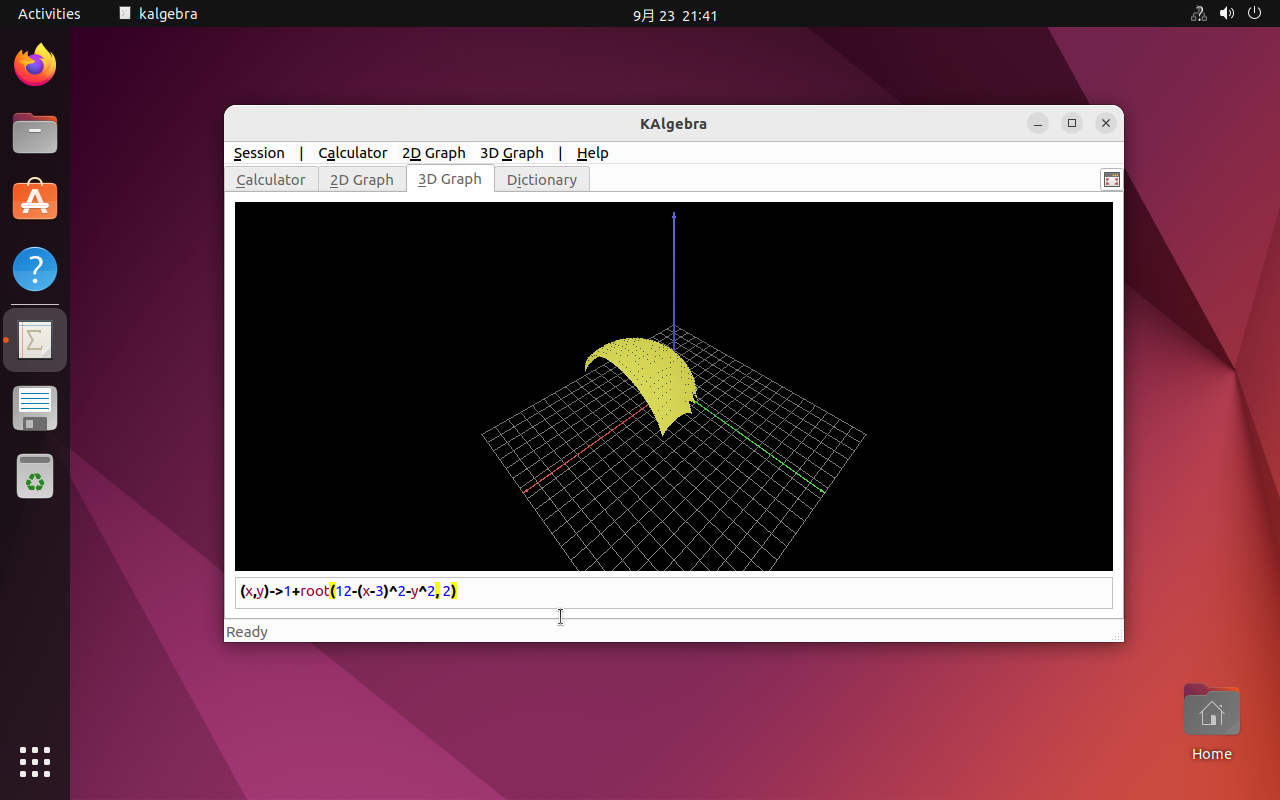}{\textbf{After entering the positive-root mapping.} Only the upper hemisphere is plotted.}
\EvoPair{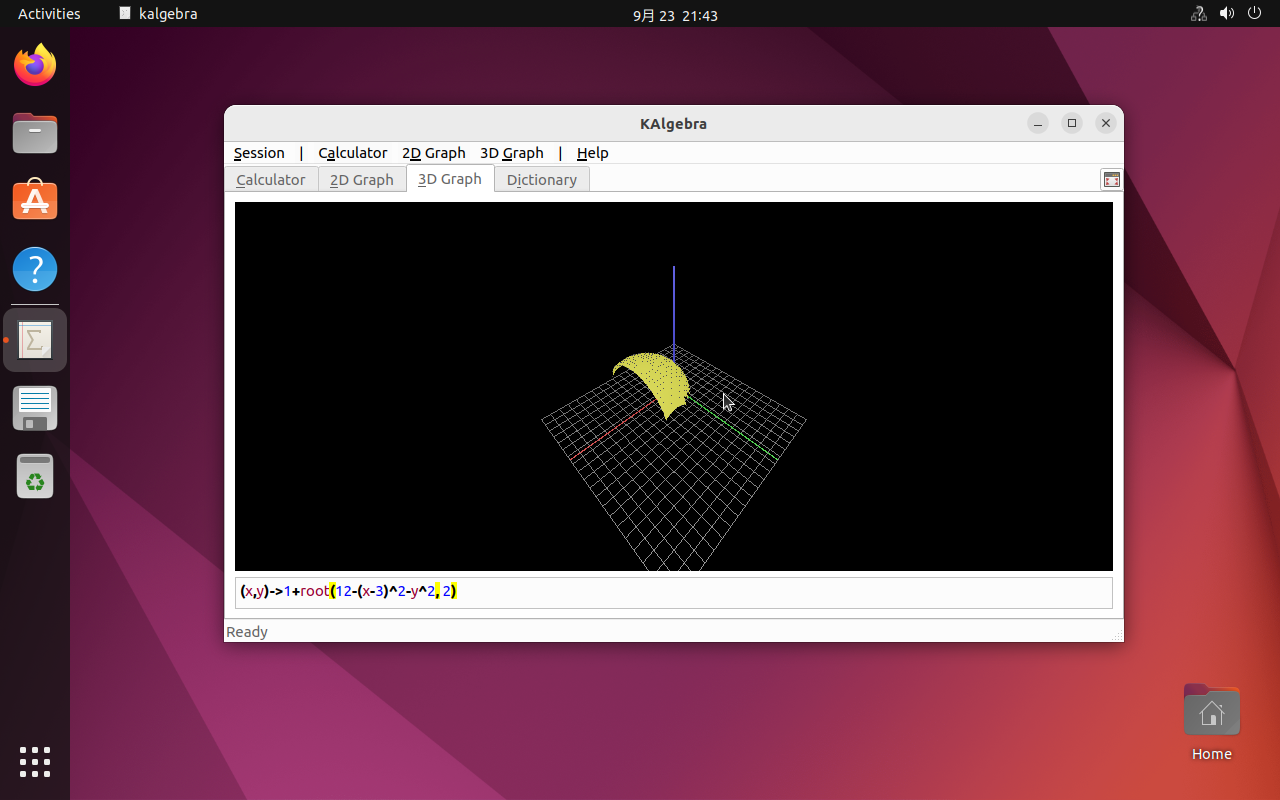}{\textbf{After zooming out.} The sphere remains incomplete.}{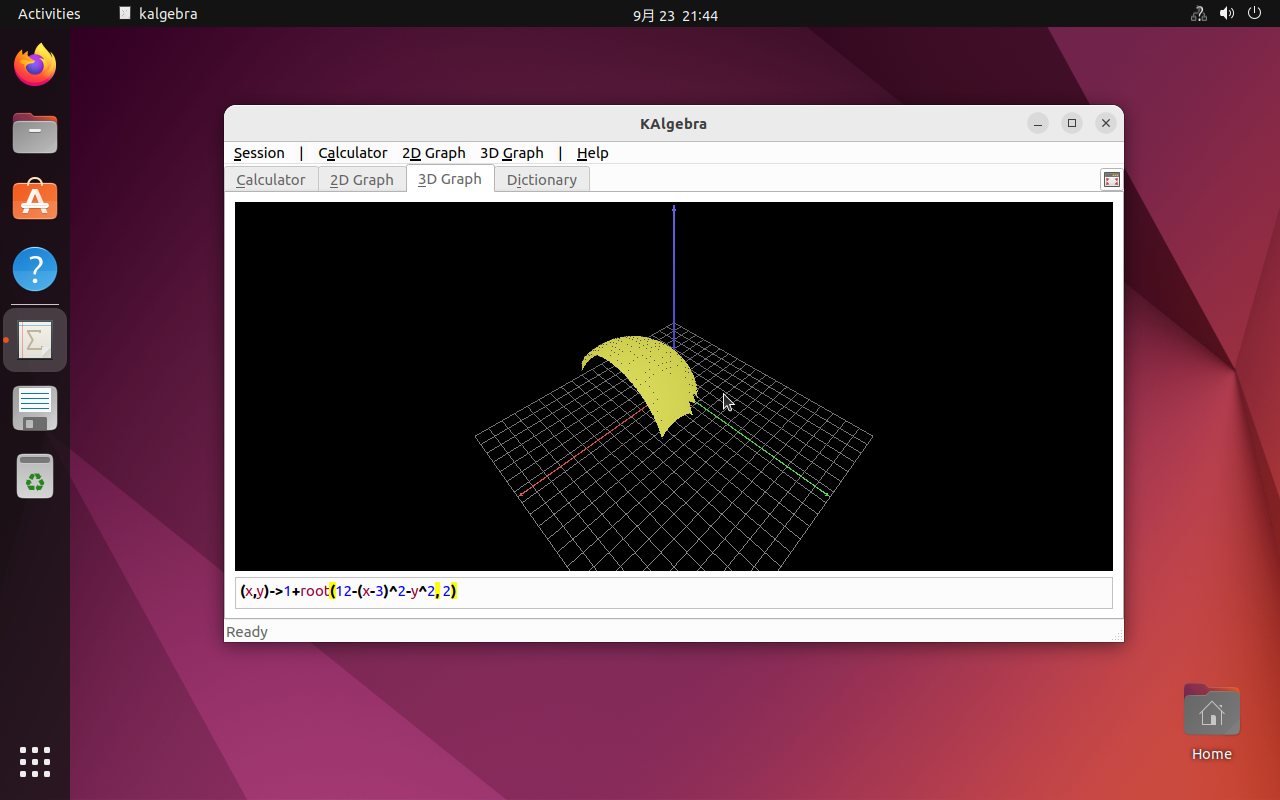}{\textbf{End-of-exploration checkpoint.} The incomplete sphere remains; evaluation fails.}

\paragraph{Implication for the initial resources.} Both trajectories show how the documented restriction narrows the agent's solution path to explicit $z=f(x,y)$ expressions. For these tasks, a single root branch covers only part of the required surface. An accepted expression and a visible plot therefore do not establish task completion. These failures motivate using actual execution to test and revise the inherited procedural constraints.

\subsection{From Execution Failure to Skill Revision}
\label{app:case-self-evolve-history}

B-14 and C-19 belong to the Expanded--Verified evolution set. Throughout exploration, the initial library and each task's first routed resource set remain fixed. After failure, the system retains the execution trajectory, screenshots, evaluator result, diagnosis, and candidate-validation records. Later explorations receive the accumulated history for that task, with historical executions explicitly identified and their mapped screenshots included in the prompt. They can therefore reconsider unsuccessful actions while still starting from the original routed resources.

Candidate revisions are tested separately in fresh environments without exploration history. After a successful exploration, the submission step uses that trajectory and the earlier failures to revise the original Skill, followed by another independent validation. The discovered procedure must thus be written into the Skill and reproduce success without access to prior trajectories or diagnoses. The following sections continue from the initial failures shown above.

\subsection{B-14: Revising the Single-Branch Procedure}
\label{app:case-self-evolve-b14}

\paragraph{First candidate: changing the square-root expression.} After the initial failure, a candidate proposes \texttt{sqrt(...)}. Independent validation reports that the function is not recognized. Falling back to \texttt{root(...)} again produces one branch, so the candidate fails.

\EvoPair{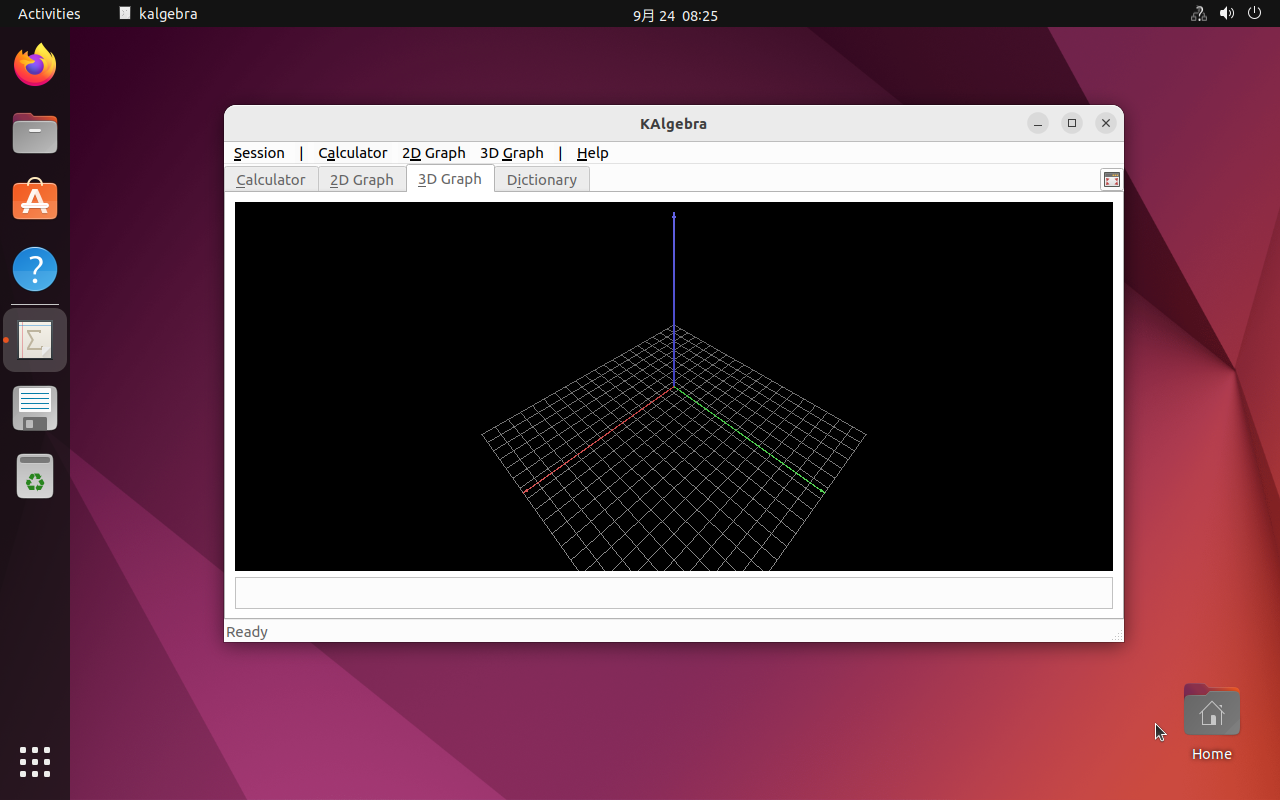}{\textbf{Candidate validation 1, initial state.} The proposed revision is tested without history.}{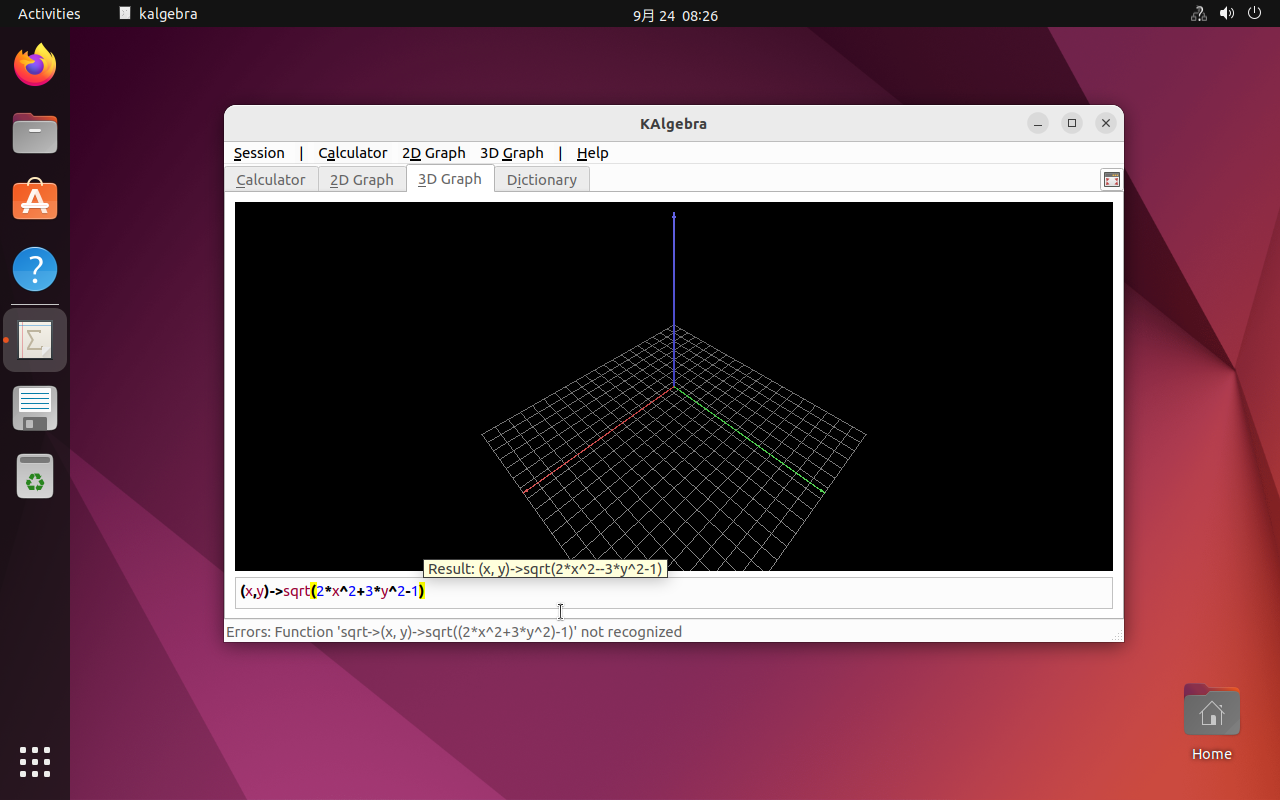}{\textbf{After entering \texttt{sqrt(...)}.} The function is not recognized.}
\EvoSingle{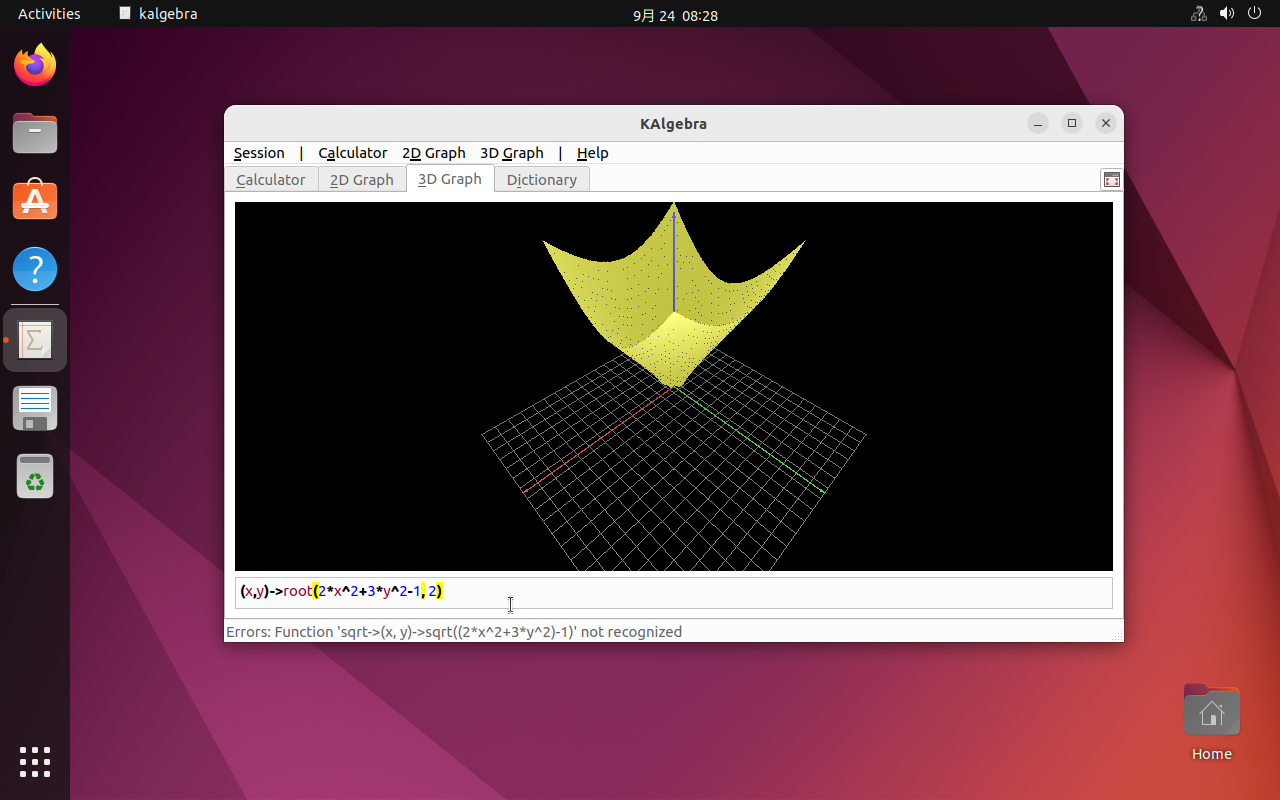}{\textbf{After fallback to \texttt{root(...)}.} Only one branch appears; validation fails.}

\paragraph{Second exploration: using the first failure records.} Exploration 2 receives exploration 1 and candidate validation 1: two executions with five screenshots. The agent consults the in-app Help for a usable 3D-plotting procedure, but the search does not establish a working solution.

\EvoPair{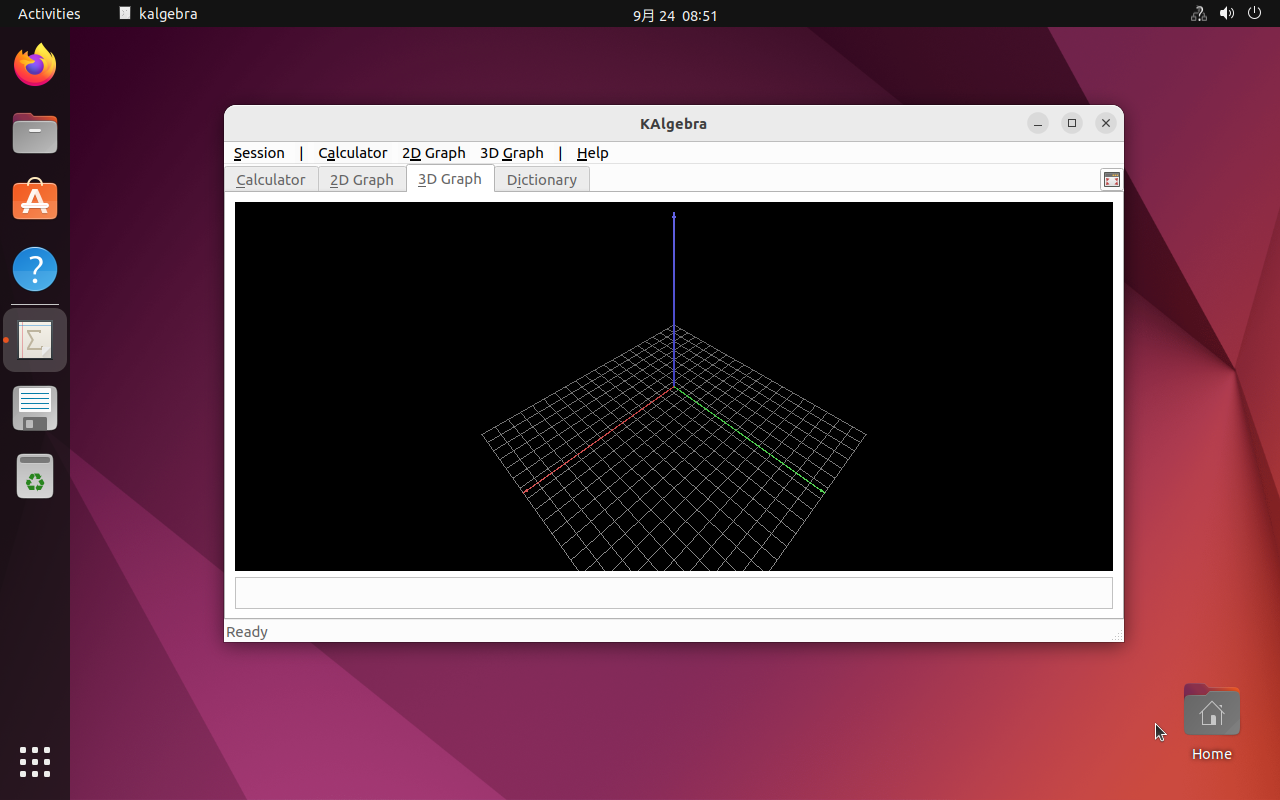}{\textbf{Exploration 2, initial state.} Earlier failure records are supplied for this run.}{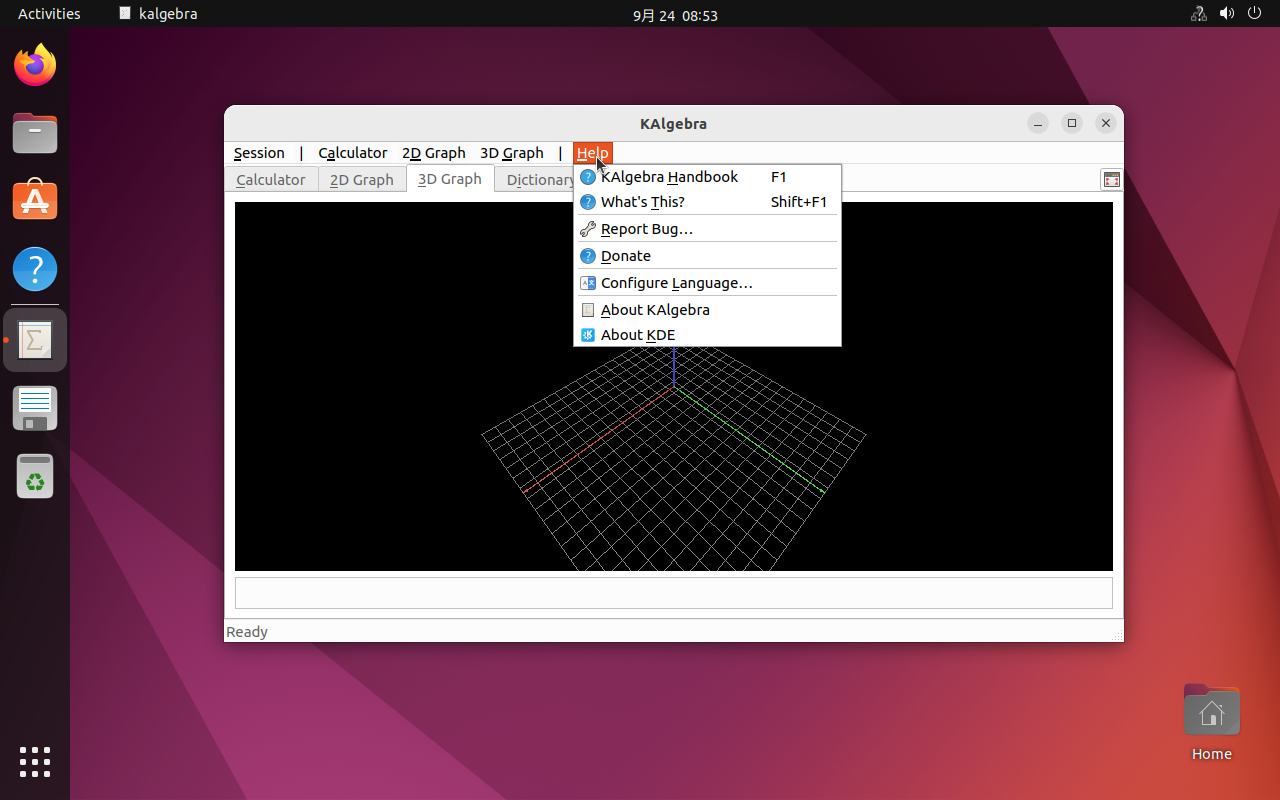}{\textbf{After opening Help.} The agent searches for plotting guidance.}
\EvoPair{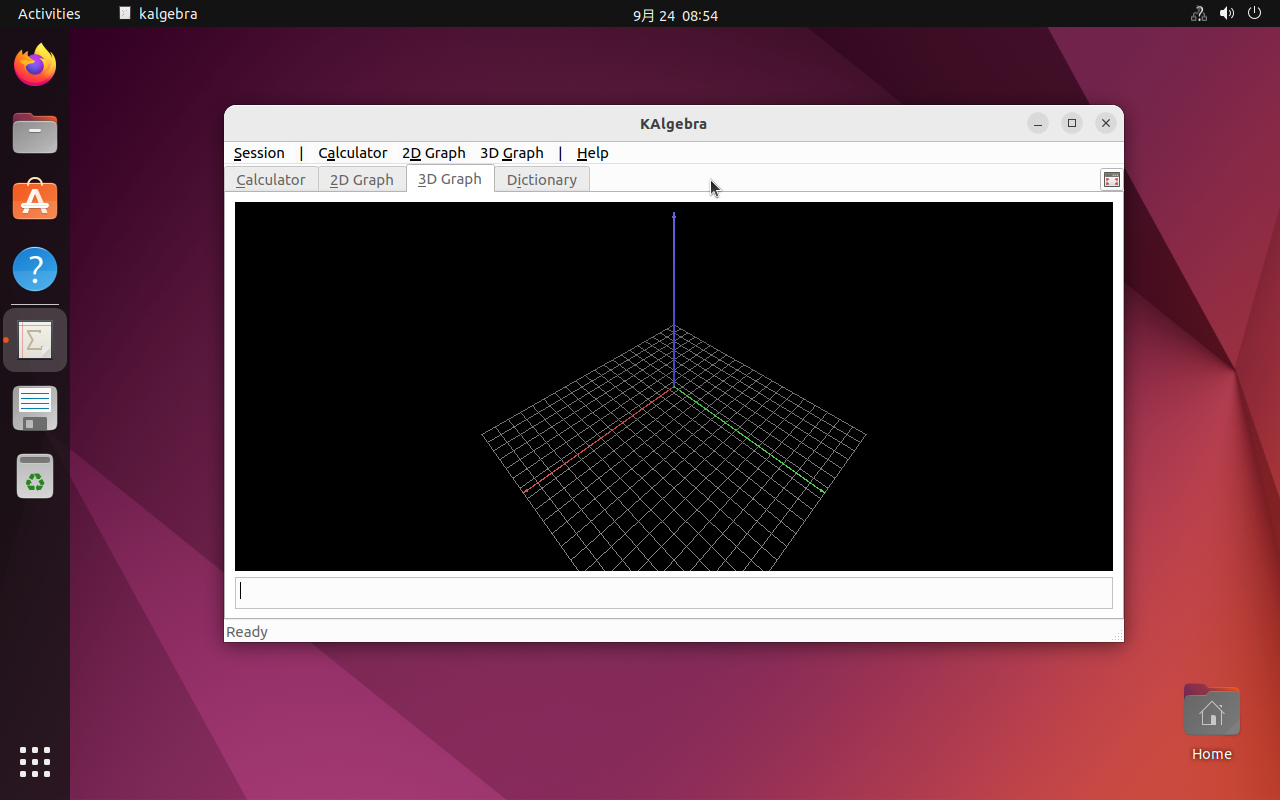}{\textbf{After F1.} The plotting window remains visible.}{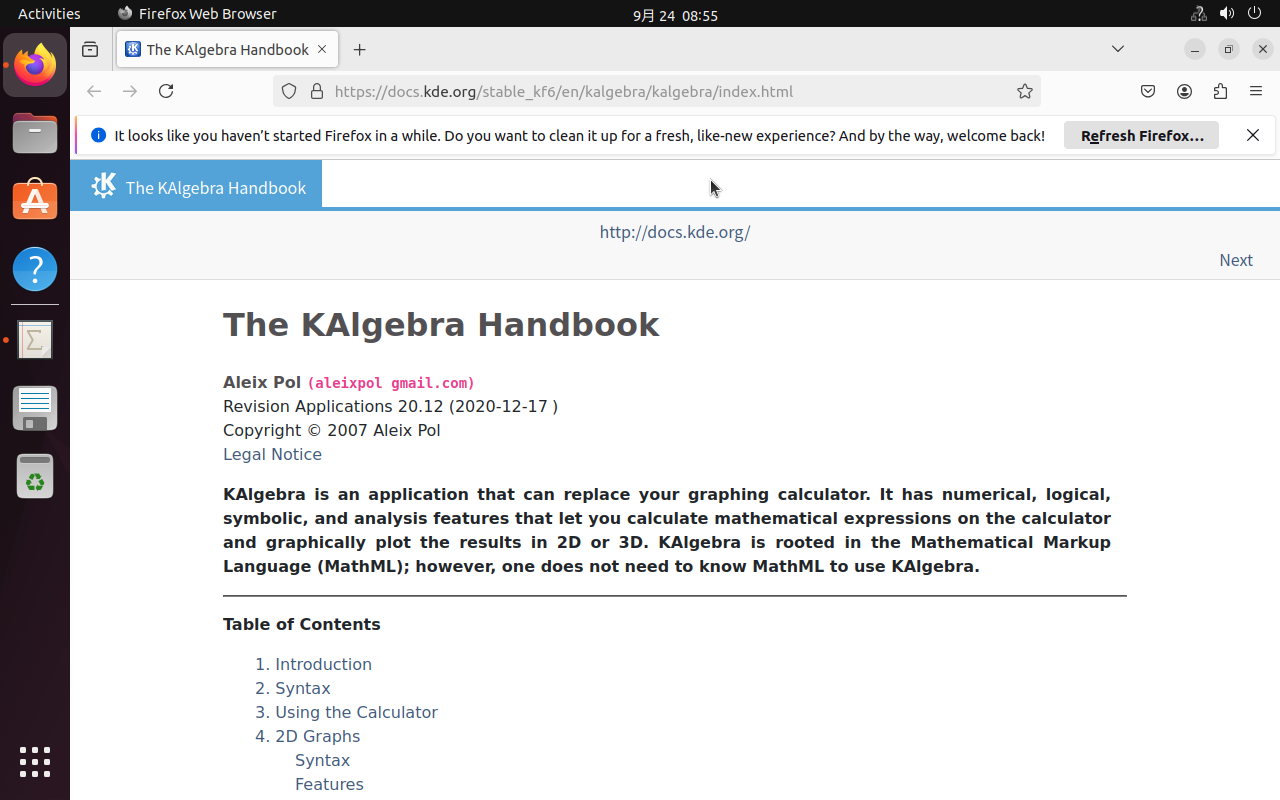}{\textbf{Next Help checkpoint.} The handbook is visible in a browser.}
\EvoSingle{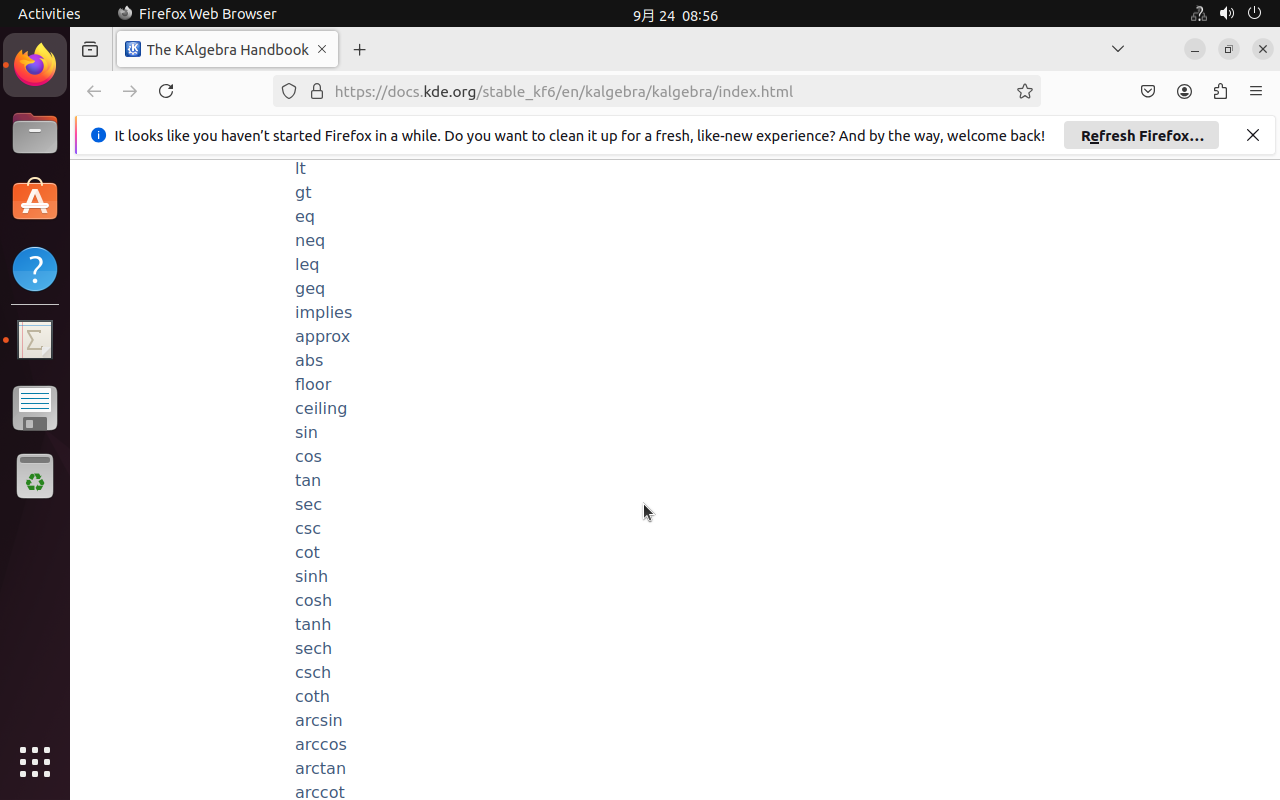}{\textbf{After searching for ``3D Graphs.''} The search does not identify a working full-surface procedure.}

A subsequent power-form candidate is also independently tested. Although the environment accepts the expression, it still plots only one branch. Changing the square-root notation has not resolved the incomplete geometry.

\EvoPair{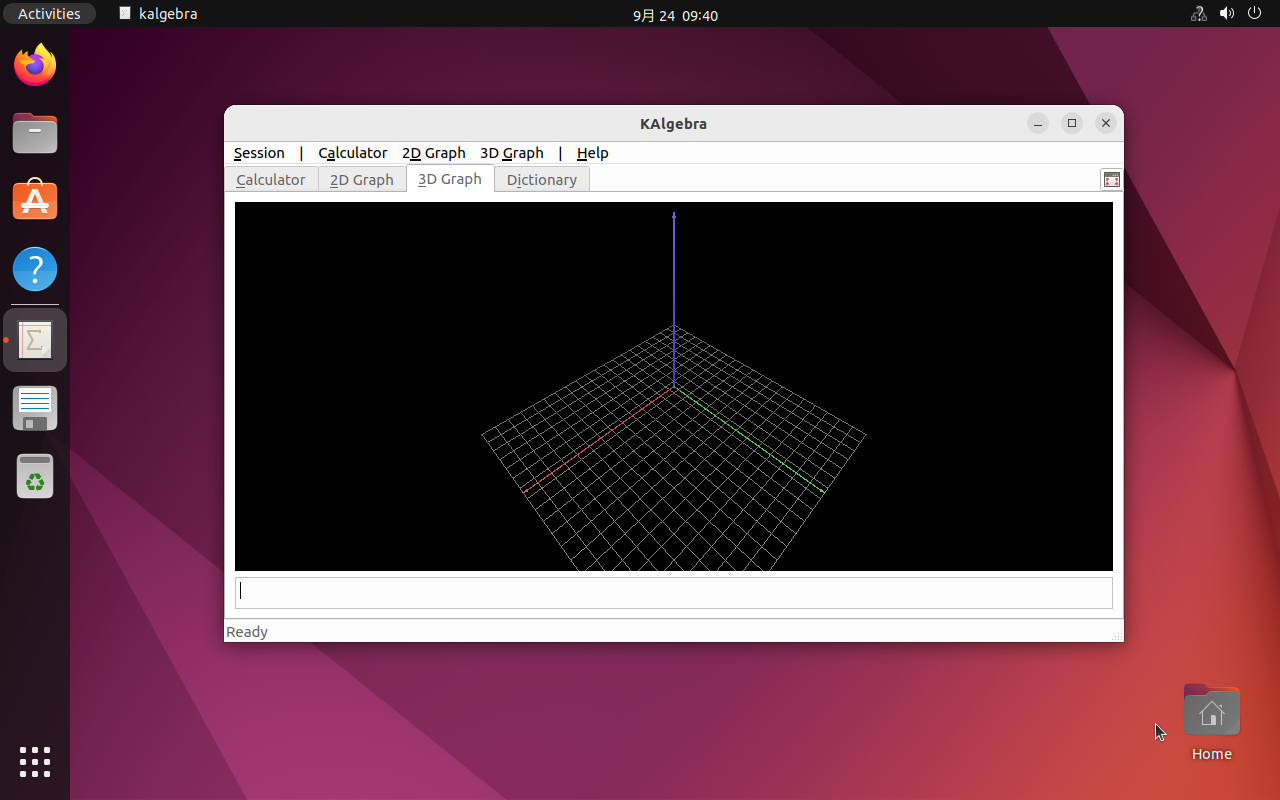}{\textbf{Candidate validation 2, initial state.} The power-form revision is tested without history.}{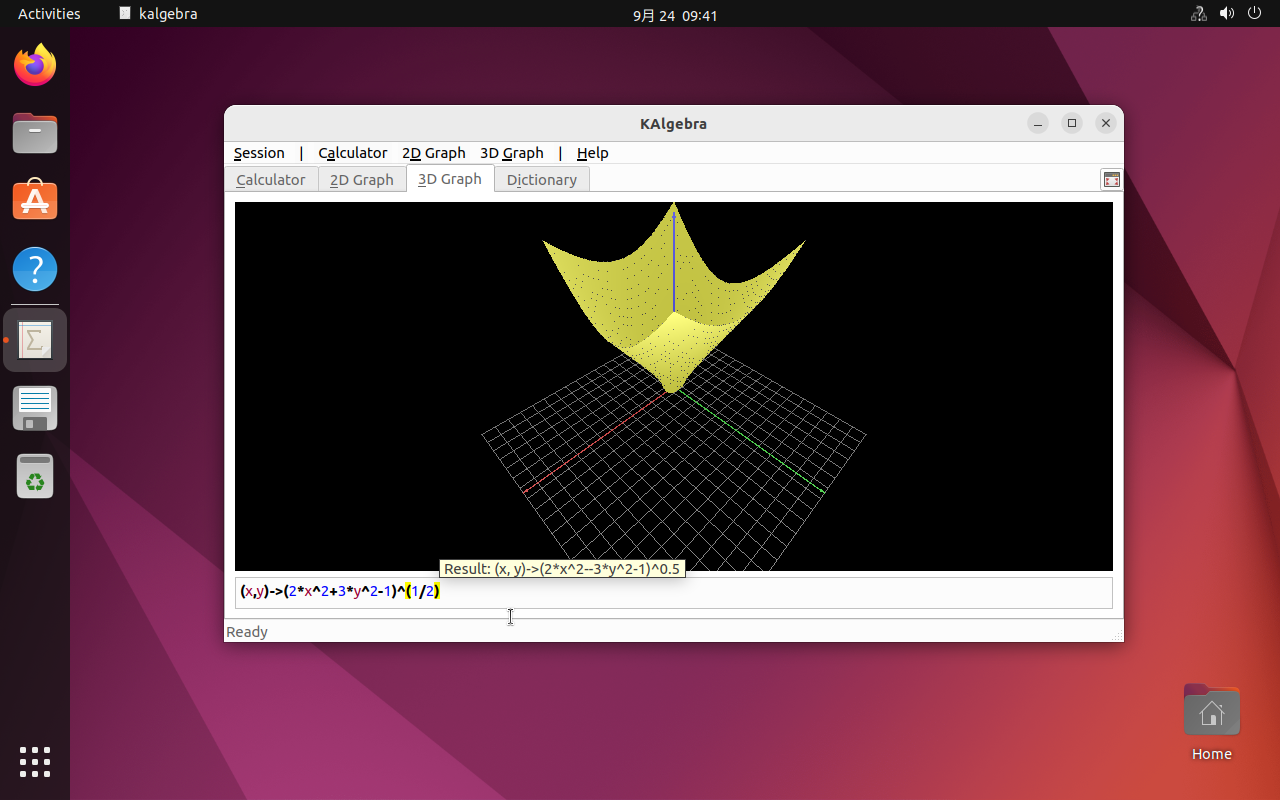}{\textbf{After entering the power mapping.} The expression is accepted, but the surface remains incomplete.}

\paragraph{Third exploration: using the accumulated evidence.} Exploration 3 receives both earlier explorations and both candidate validations: four executions with twelve screenshots. The agent then enters \texttt{(x,y,z)->2*x\^{}2+3*y\^{}2-z\^{}2-1}. The environment accepts this three-variable mapping and renders the complete one-sheet hyperboloid; evaluation passes. This execution directly contradicts the initial explicit-only restriction.

\EvoPair{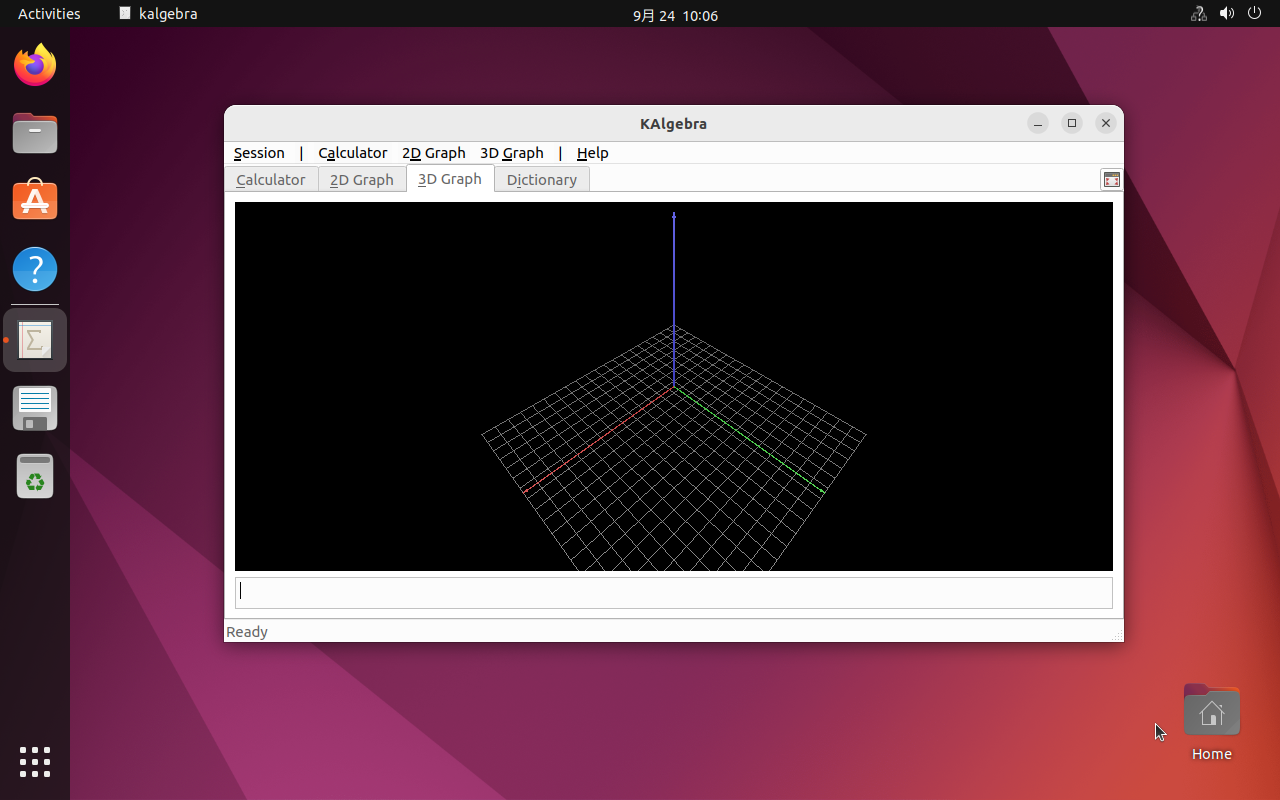}{\textbf{Exploration 3, initial state.} The accumulated history accompanies the original routed resources.}{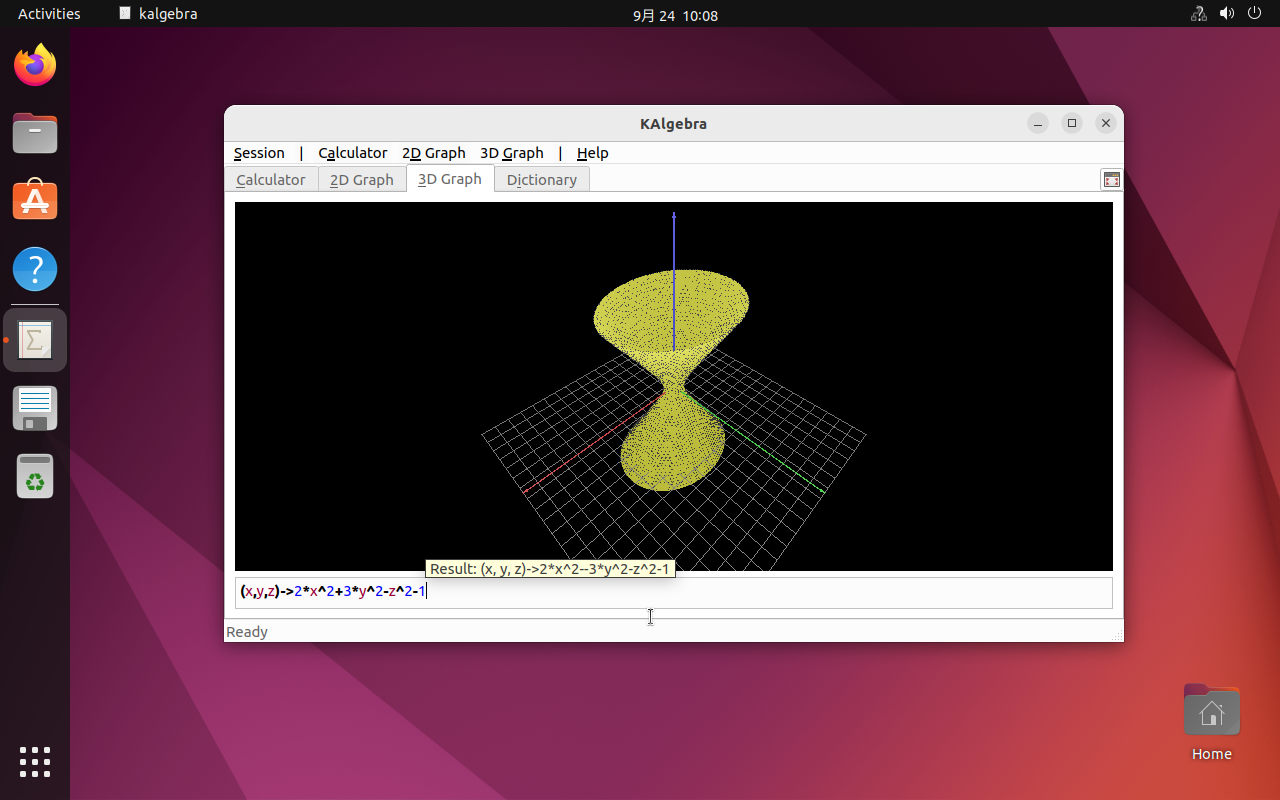}{\textbf{After entering the three-variable mapping.} The complete hyperboloid appears; evaluation passes.}

\paragraph{Encoding and independently validating the update.} The submitted revision removes the incorrect restriction on defining $z$, retains the explicit-surface workflow, and adds a conditional three-variable mapping path for implicit surfaces. It also explains that a single explicit root branch does not represent the complete surface. A fresh run using the revised Skill reproduces the full hyperboloid without exploration history, showing that the successful procedure has been captured in the Skill.

\EvoPair{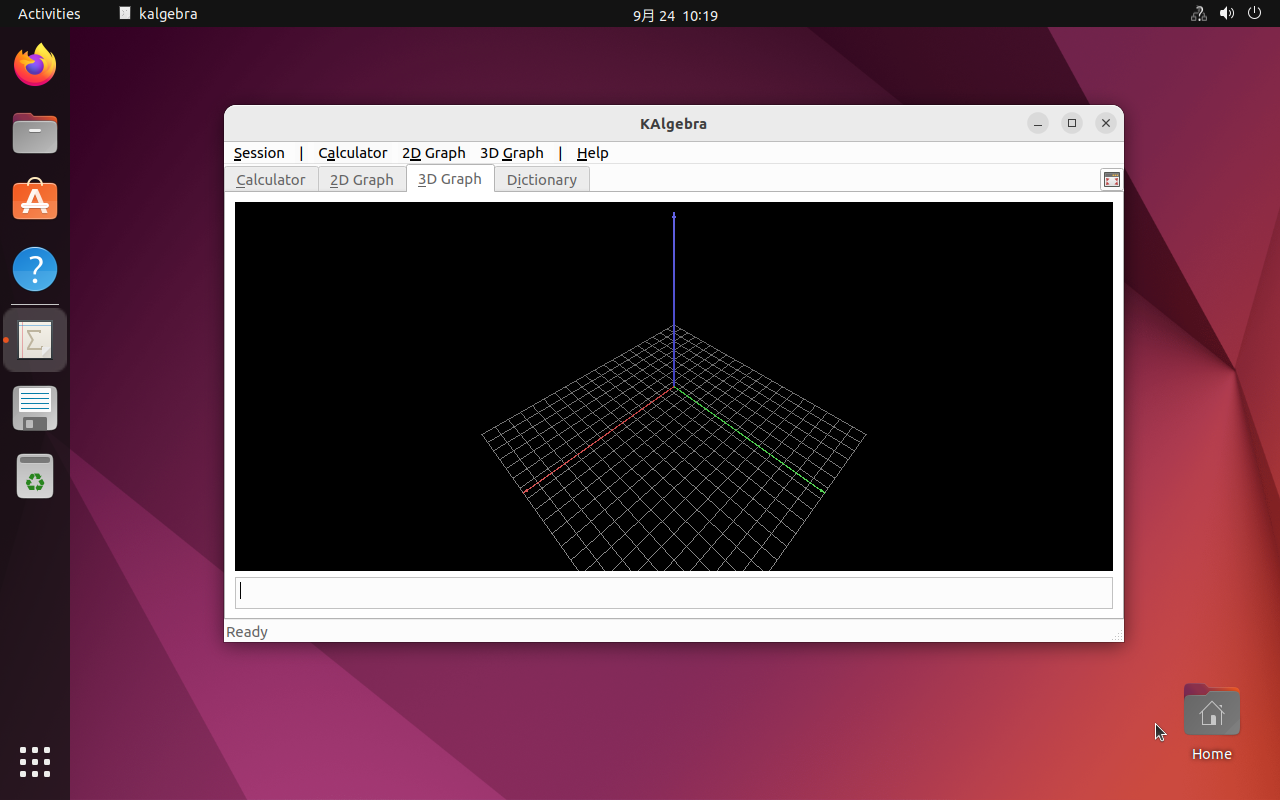}{\textbf{Independent validation, initial state.} The revised Skill is loaded without history.}{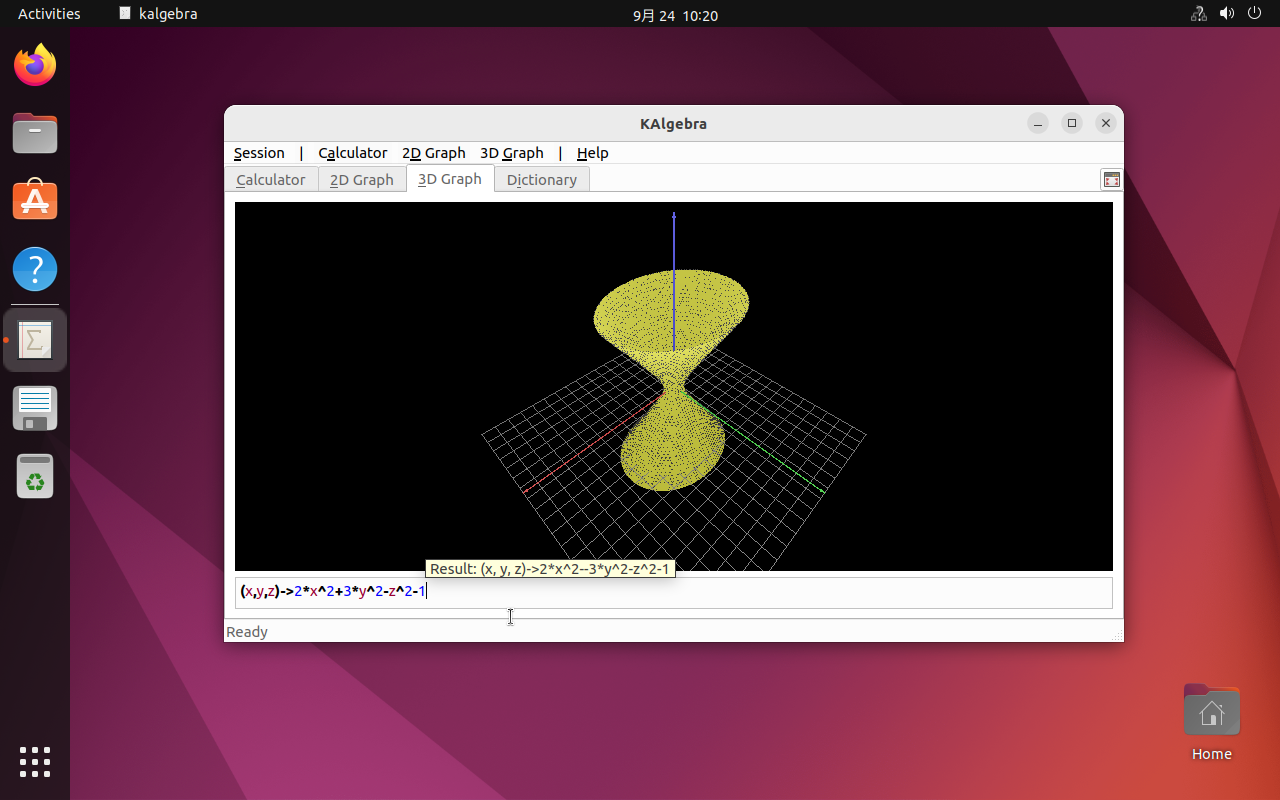}{\textbf{After following the revised mapping.} The complete hyperboloid is reproduced; validation passes.}

\subsection{C-19: Revising the Two-Hemisphere Procedure}
\label{app:case-self-evolve-c19}

\paragraph{First candidate: combining explicit branches.} Following the initial failure in Section~\ref{app:case-self-evolve-mismatch}, a candidate proposes plotting the positive and negative root branches to form the full sphere. Its independent validation does not produce the required combination: the visible result remains a single hemisphere.

\EvoPair{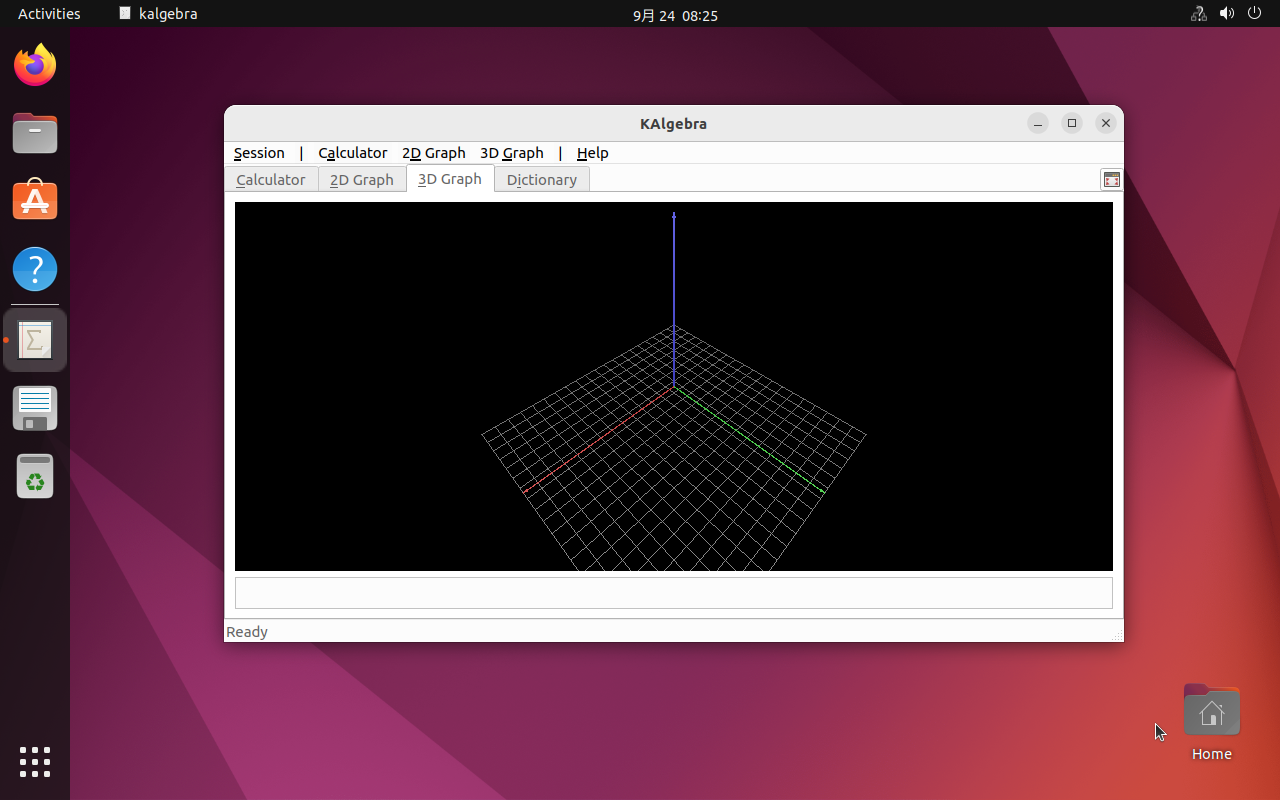}{\textbf{Candidate validation, initial state.} The two-branch revision is tested without history.}{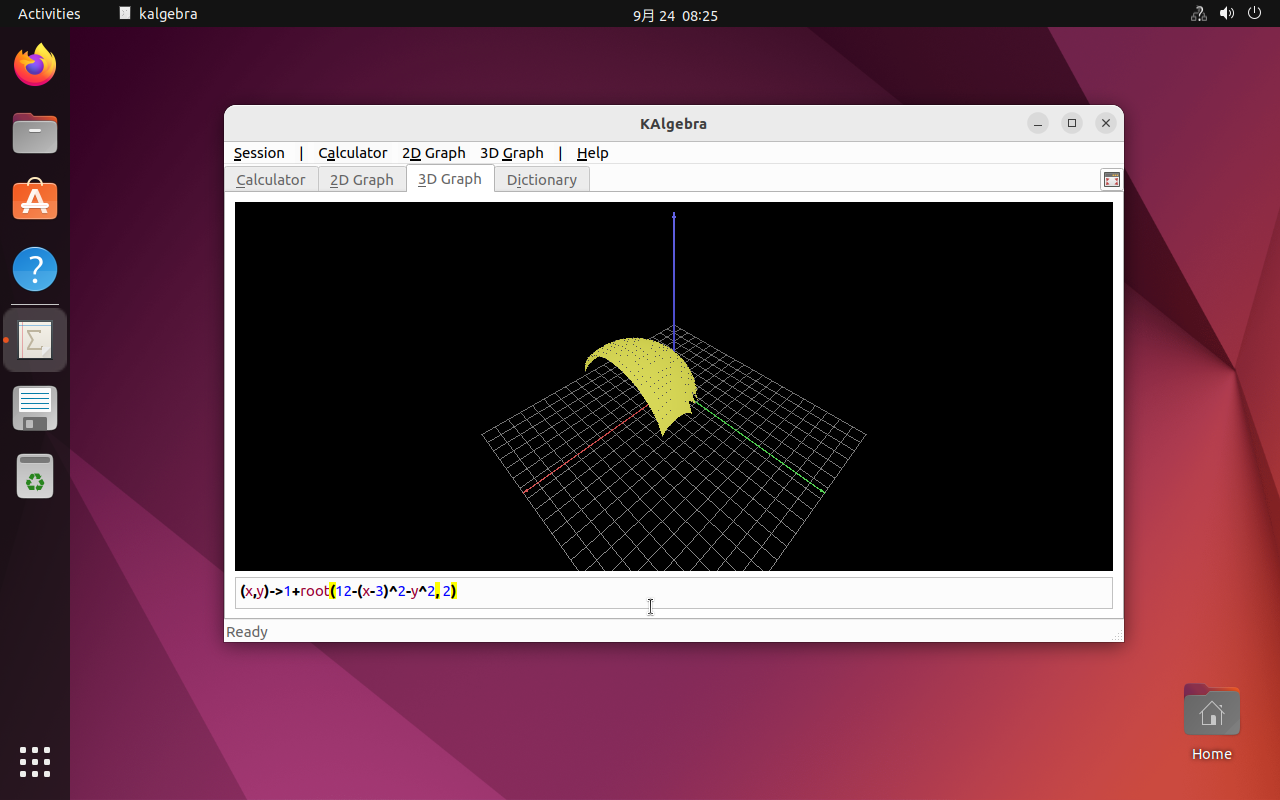}{\textbf{After plotting the positive branch.} Only the upper hemisphere is visible.}
\EvoPair{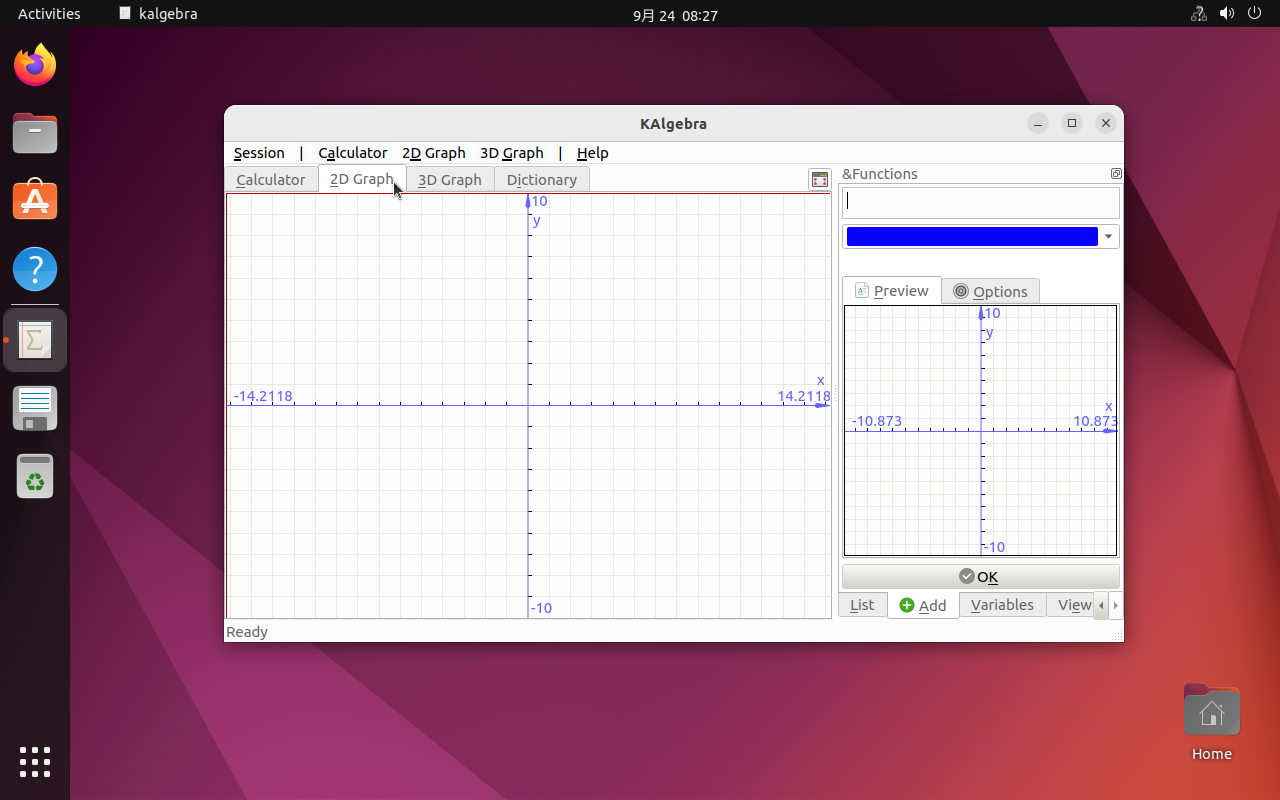}{\textbf{After switching tabs.} A second plotted branch has not been produced.}{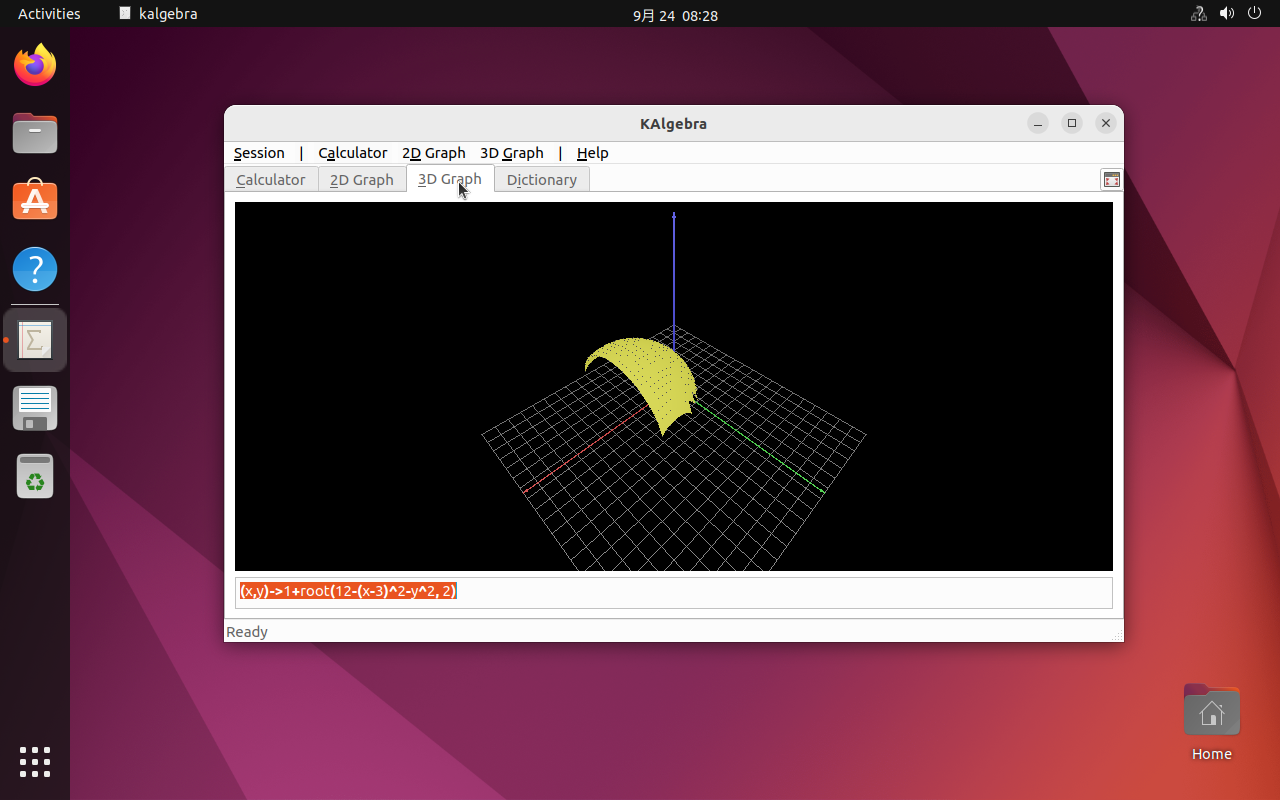}{\textbf{After returning to 3D Graph.} The input still contains the positive branch.}
\EvoSingle{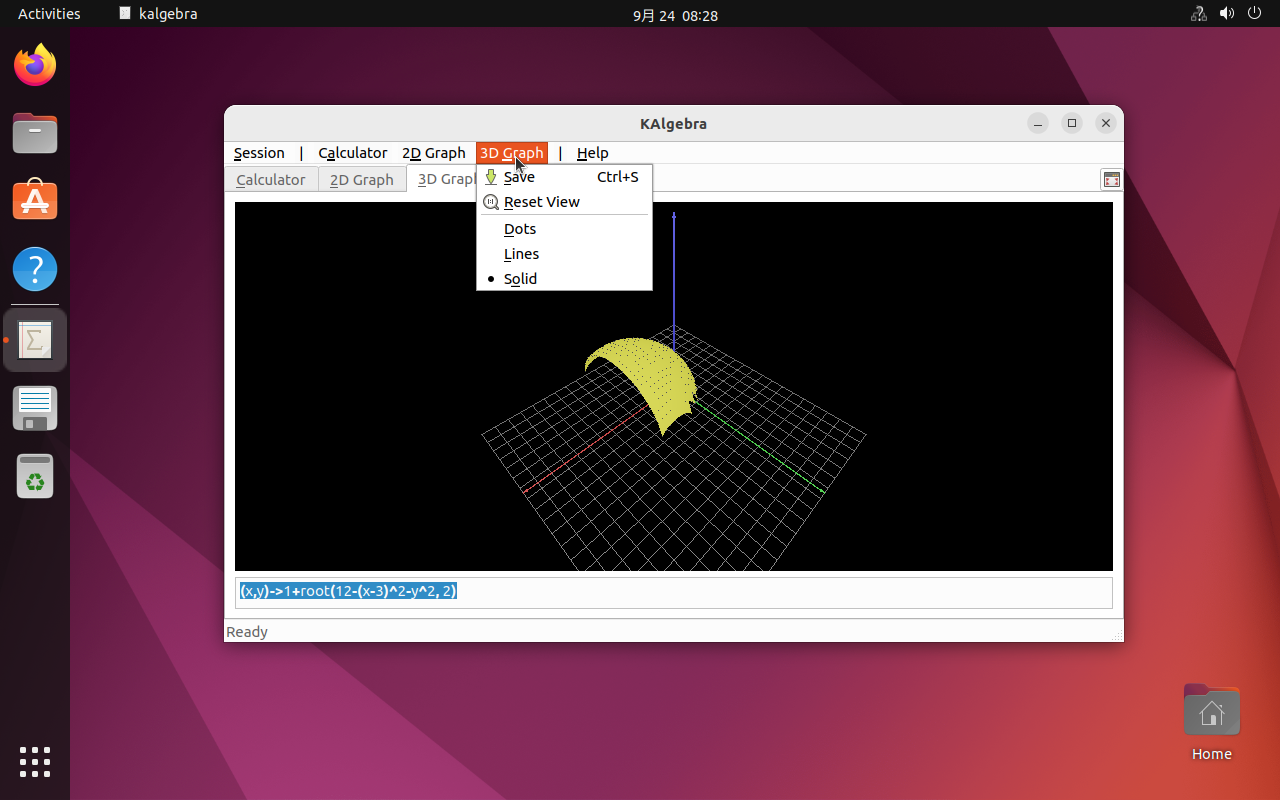}{\textbf{After opening the 3D Graph menu.} The sphere remains incomplete; validation fails.}

\paragraph{Second exploration: identifying replacement and testing an implicit equation.} Exploration 2 receives the initial exploration and failed candidate validation: two executions with nine screenshots. Entering the positive and then the negative root reveals that the latter replaces the former in the input field. The displayed surface changes from the upper to the lower hemisphere rather than combining them. The agent then enters \texttt{(x-3)\^{}2+y\^{}2+(z-1)\^{}2=12}. The complete sphere appears and evaluation passes.

\EvoPair{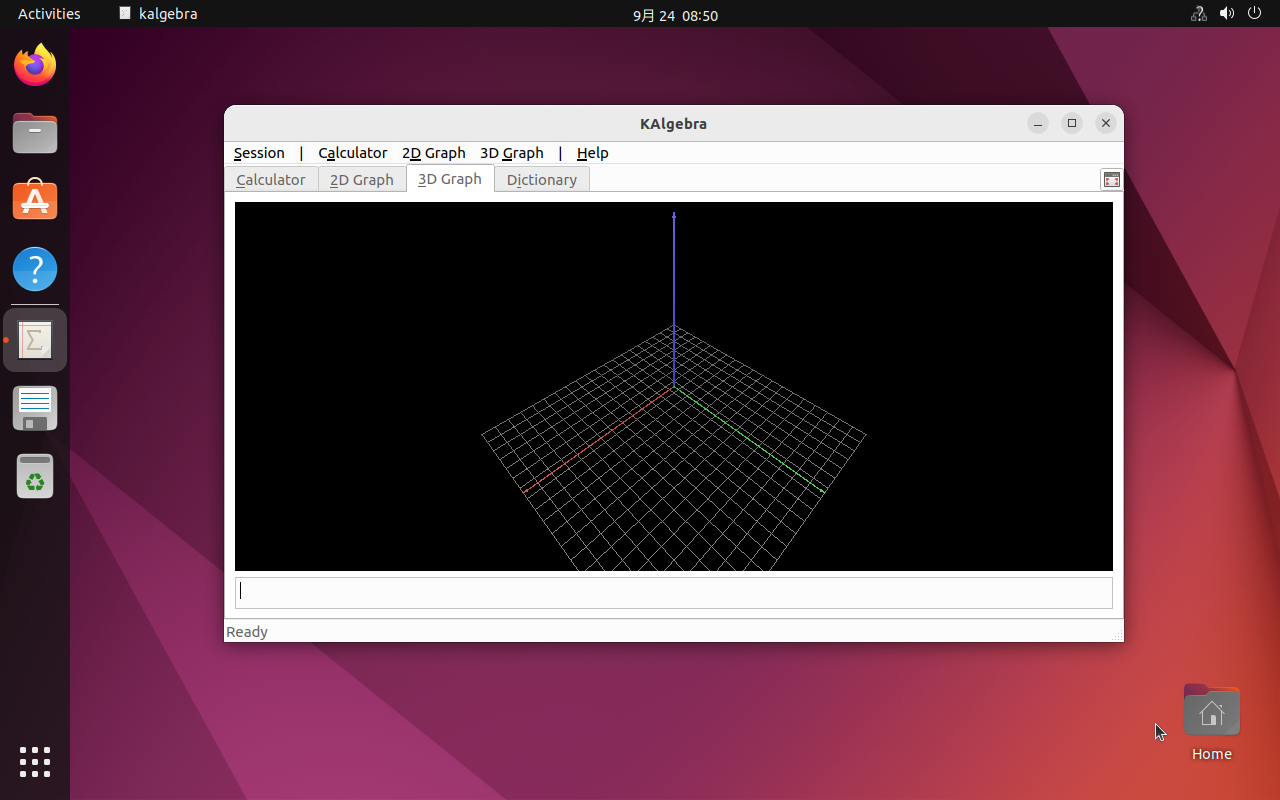}{\textbf{Exploration 2, initial state.} The earlier failure records are supplied.}{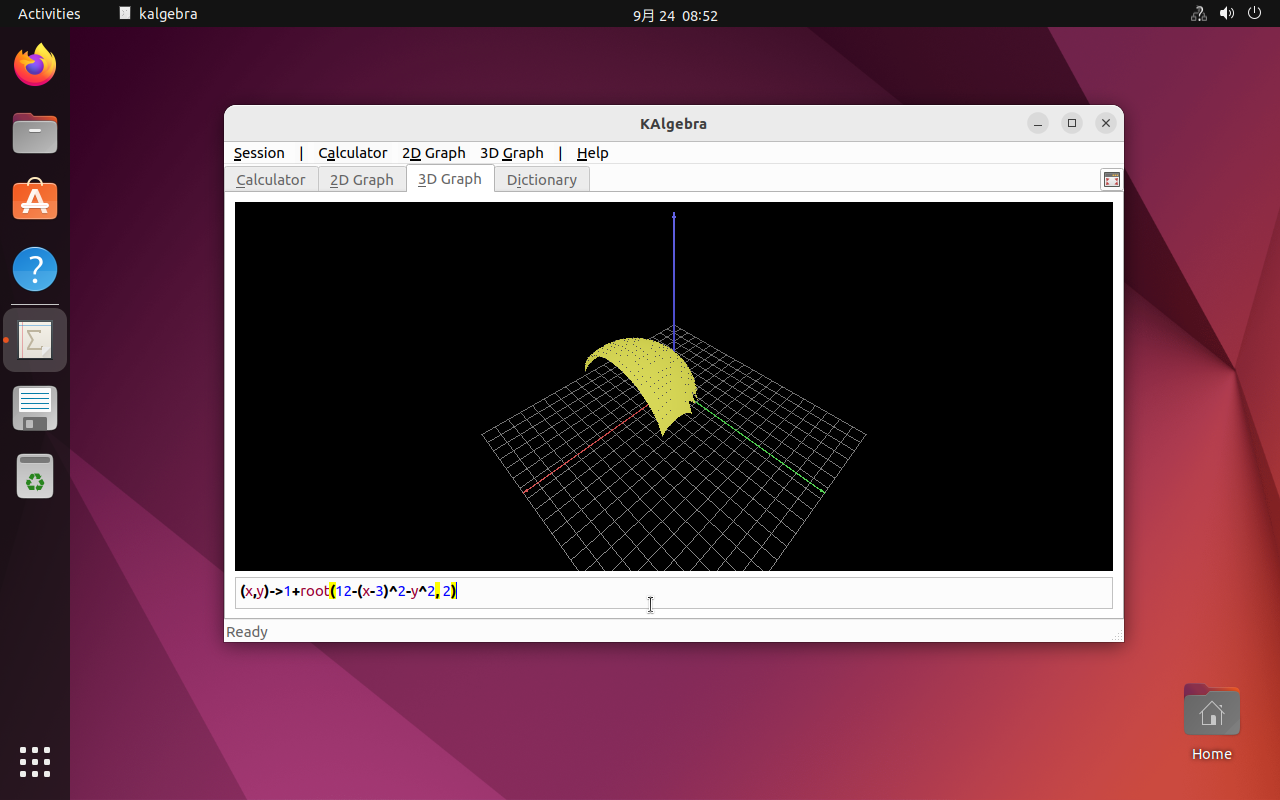}{\textbf{After entering the positive branch.} The upper hemisphere appears.}
\EvoPair{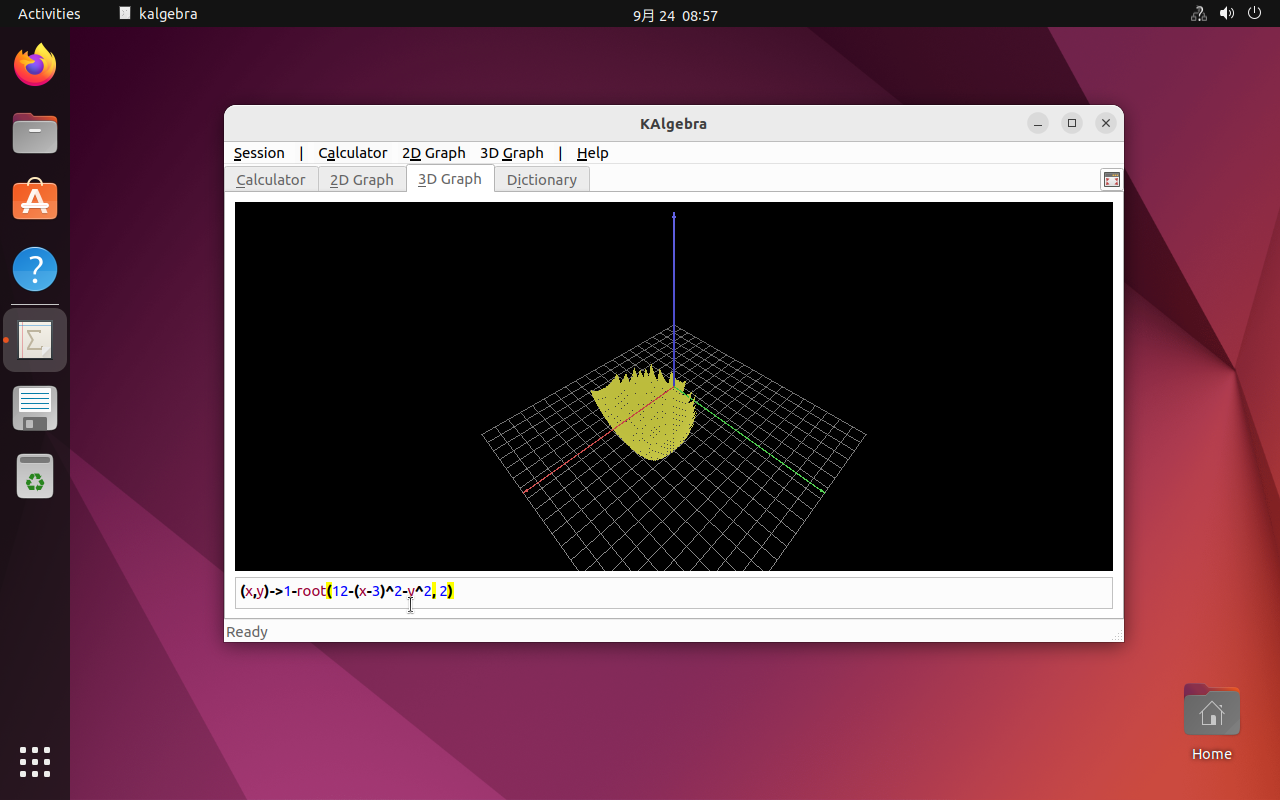}{\textbf{After entering the negative branch.} The lower hemisphere replaces the upper one.}{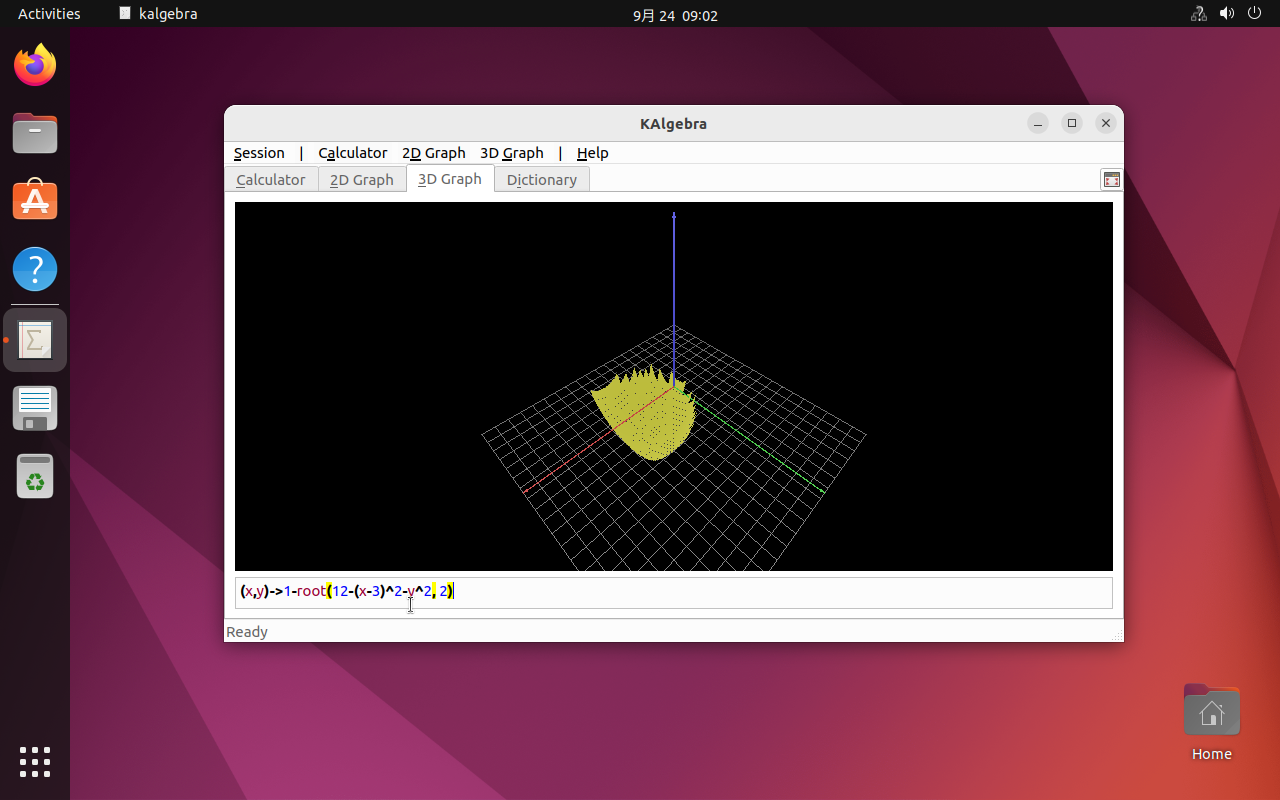}{\textbf{Checkpoint before the implicit equation.} Only the lower hemisphere remains.}
\EvoSingle{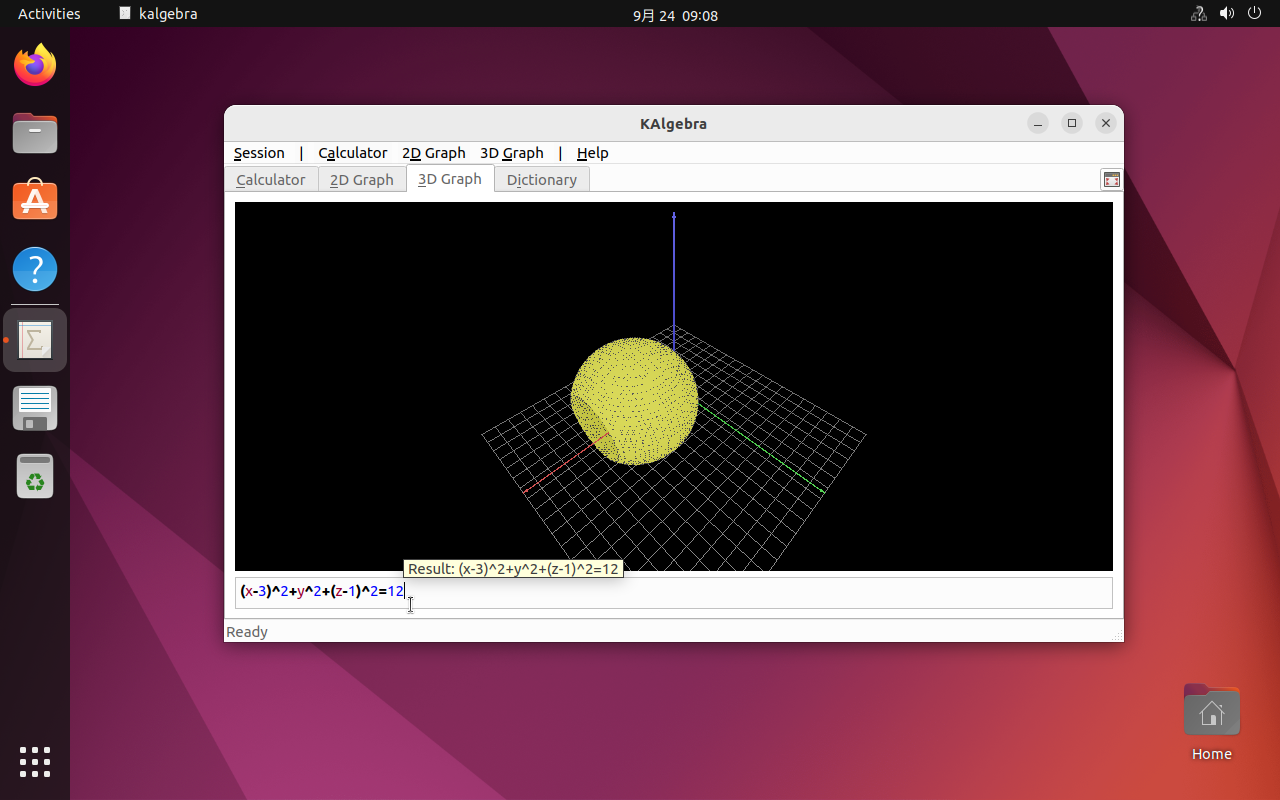}{\textbf{After entering the implicit equation.} The complete sphere appears; evaluation passes.}

\paragraph{Encoding and independently validating the update.} The revision retains the explicit-mapping procedure, corrects the version-dependent restriction, and adds the observed implicit-equation route. It also states that entering the two explicit hemisphere expressions successively replaces the input rather than combining the surfaces. Without exploration history, an independent run using the revised Skill again produces the complete sphere.

\EvoPair{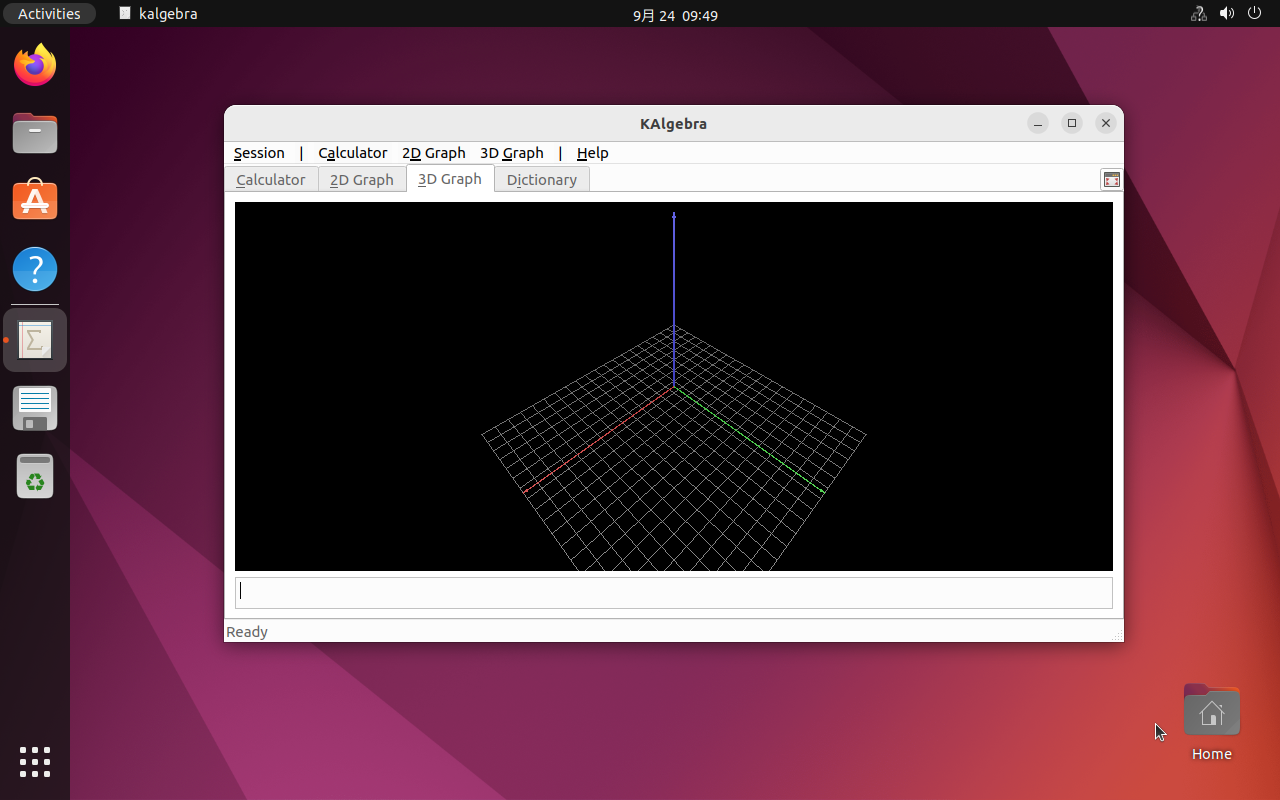}{\textbf{Independent validation, initial state.} The revised Skill is loaded without history.}{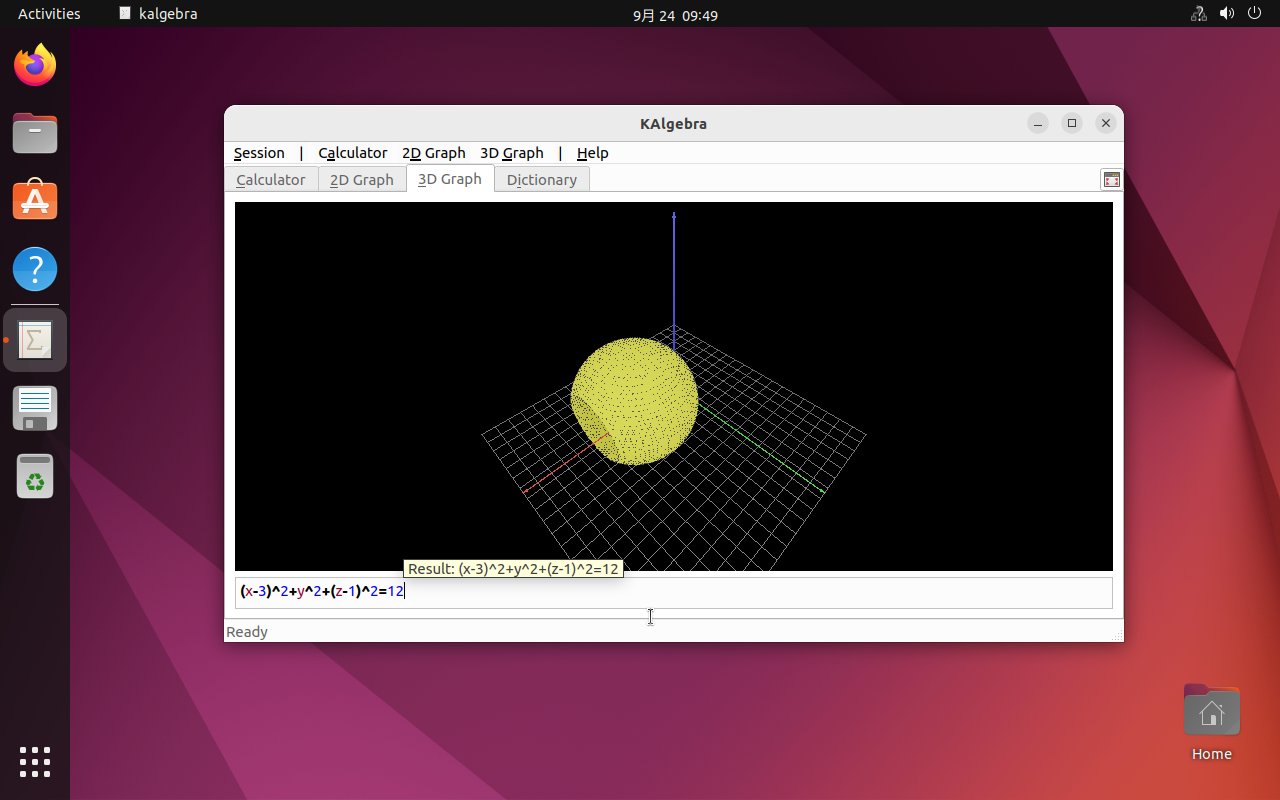}{\textbf{After entering the sphere equation.} The complete sphere is reproduced; validation passes.}

\subsection{Consolidation and Regression: Preserving Both Revisions}
\label{app:case-self-evolve-consolidation}

The two accepted candidates revise the same initial Skill. They are consolidated into one entry that preserves the separately demonstrated procedures: the three-variable mapping used in B-14 and the implicit equation used in C-19. The consolidated Skill is replayed on both contributing tasks without history. Both tasks succeed again, showing that consolidation retains the capabilities established by the two revisions.

\EvoPair{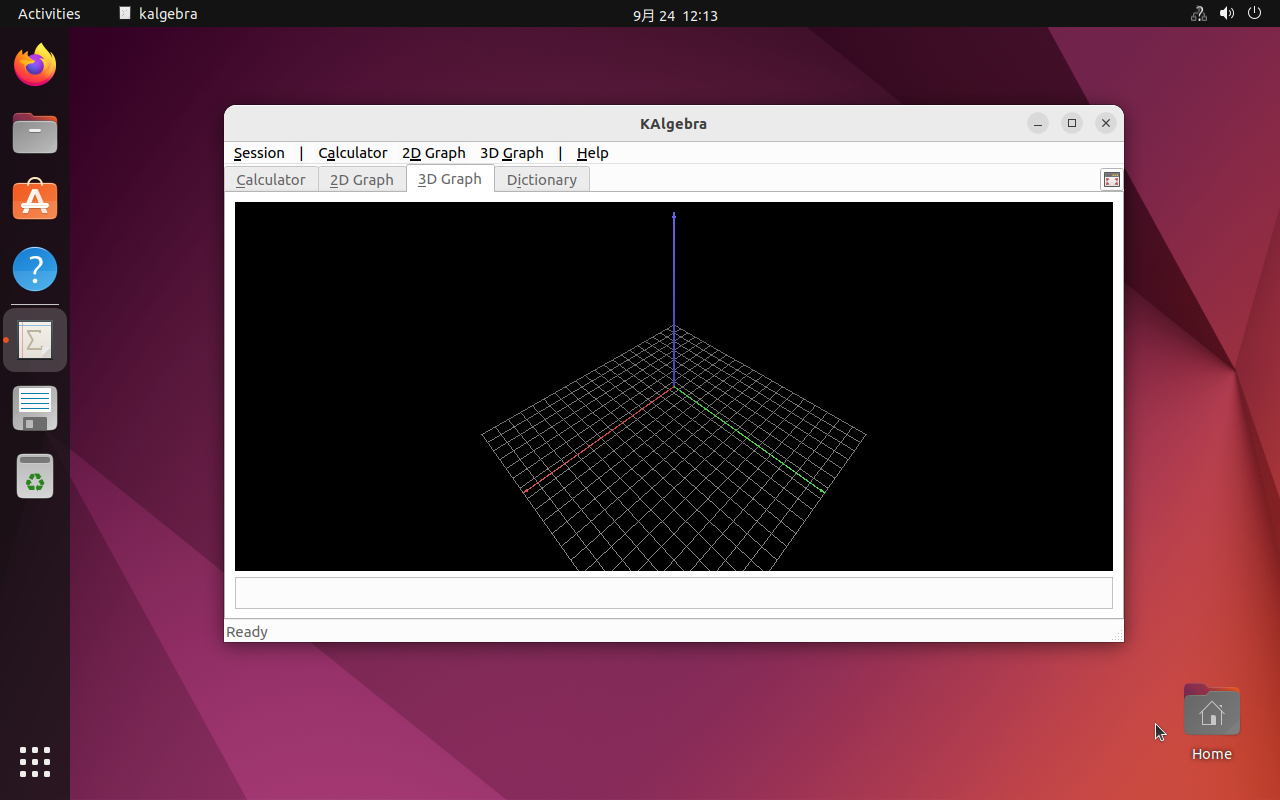}{\textbf{B-14 regression, initial state.} The consolidated Skill is tested without history.}{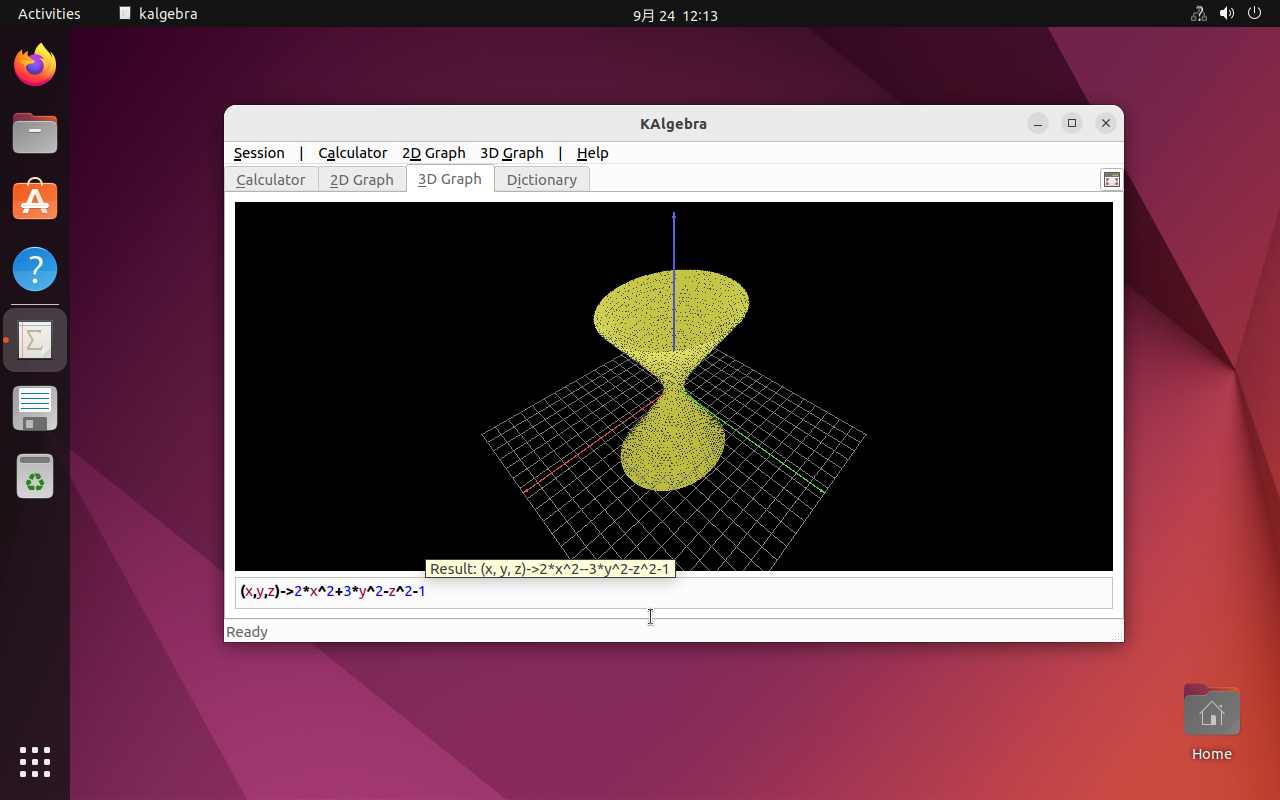}{\textbf{B-14 regression result.} The complete hyperboloid is reproduced.}
\EvoPair{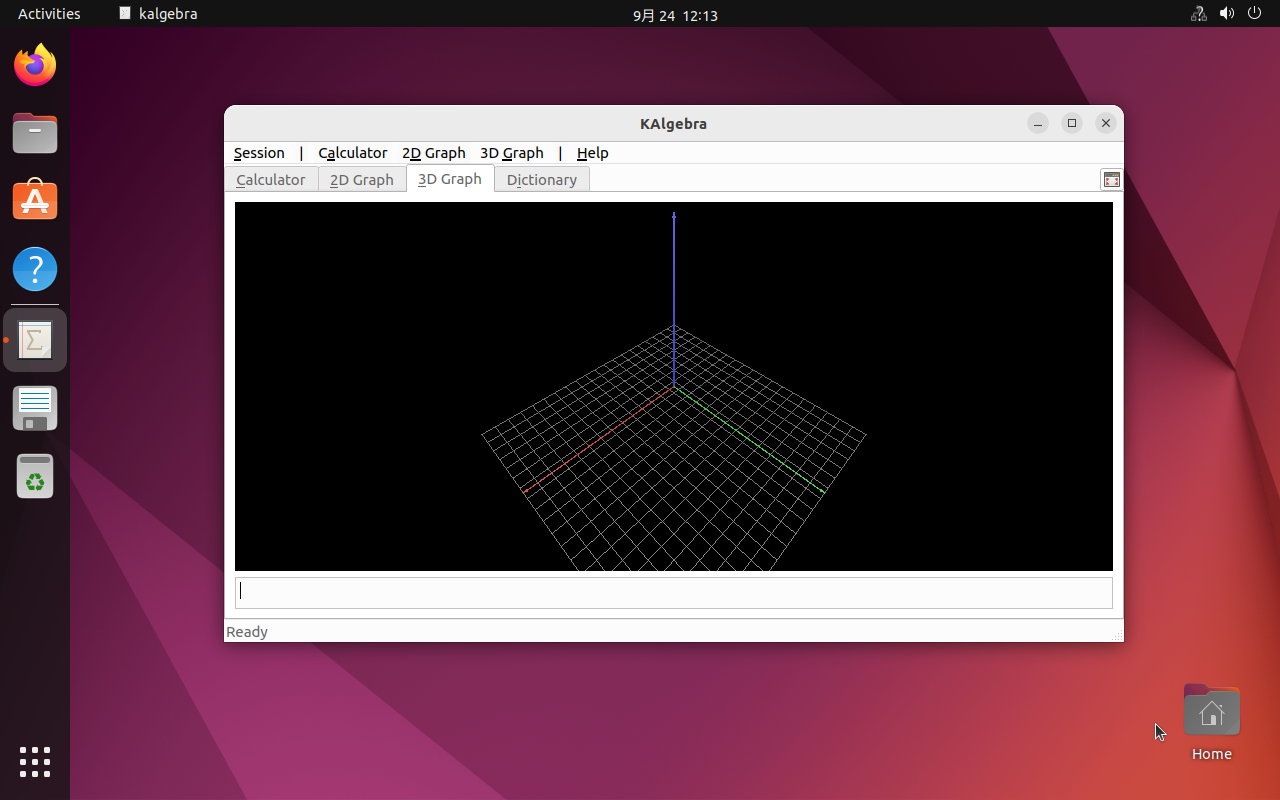}{\textbf{C-19 regression, initial state.} The consolidated Skill is tested without history.}{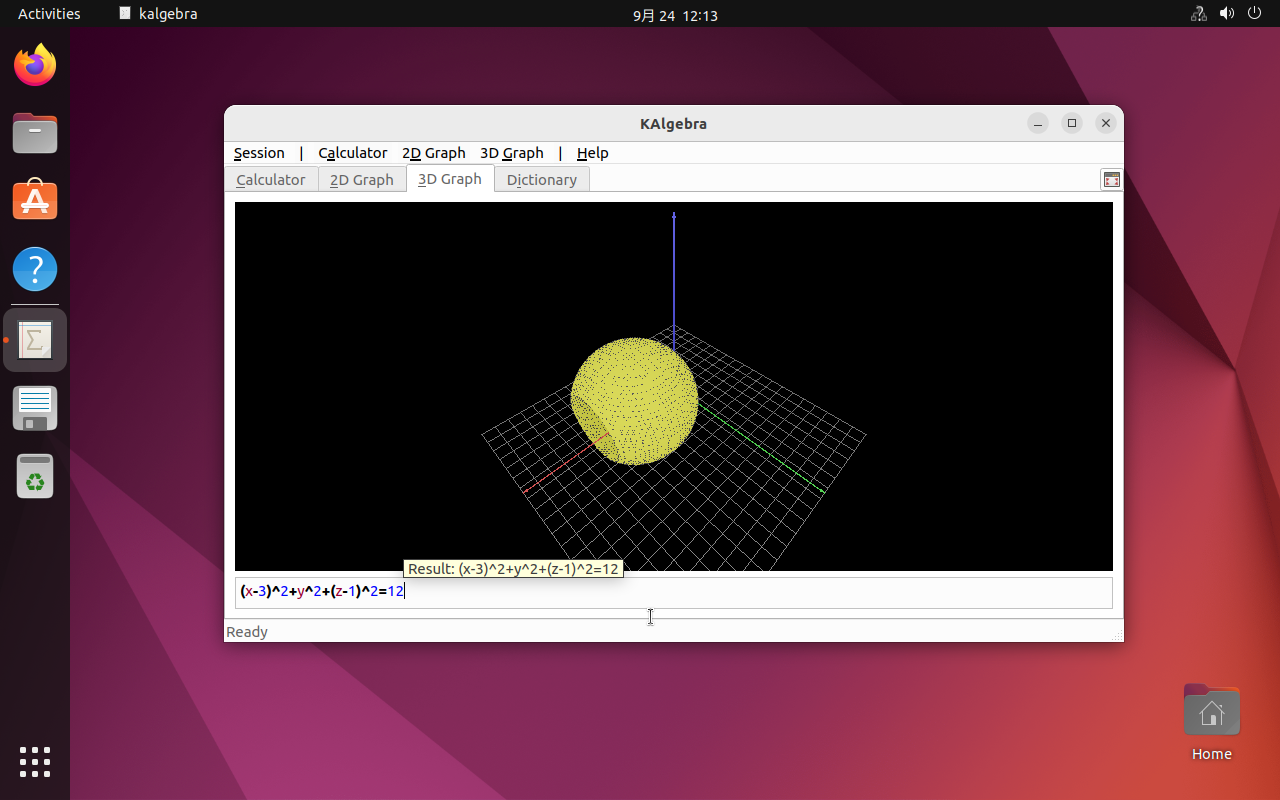}{\textbf{C-19 regression result.} The complete sphere is reproduced.}

The initial failures have now led to validated, reusable procedures through accumulated history, exploration, revision, and regression. We next test whether these procedures help tasks that did not contribute to the revisions.

\begingroup
\renewcommand{\EvoImageScale}{0.72}
\subsection{Transfer to Verified: Previously Failed Tasks Succeed}
\label{app:case-self-evolve-transfer}

After consolidation, the updated Skill is frozen. Tasks A-03, C-14, and C-17 belong to the disjoint Verified set and did not participate in the revisions above. All three failed with the initial library. Their subsequent runs route to the updated Skill without receiving the evolution histories of B-14 or C-19. The table lists the expressions entered in the successful runs.

\begin{center}
\small
\begin{tabular}{@{}p{0.12\linewidth}p{0.46\linewidth}p{0.16\linewidth}p{0.16\linewidth}@{}}
\hline
Task & Expression with updated Skill & Initial library & Updated Skill \\
\hline
A-03 & \texttt{(x,y,z)->x\^{}2+y\^{}2-z\^{}2-1} & Failed & Passed \\
C-14 & \texttt{x\^{}2+y\^{}2+z\^{}2=2} & Failed & Passed \\
C-17 & \texttt{(x,y,z)->(x+z)\^{}2+y\^{}2-1} & Failed & Passed \\
\hline
\end{tabular}
\end{center}

\paragraph{A-03.} The initial attempt renders only one branch. With the updated Skill, the agent uses a three-variable mapping to render the complete hyperboloid and passes evaluation.
\EvoPair{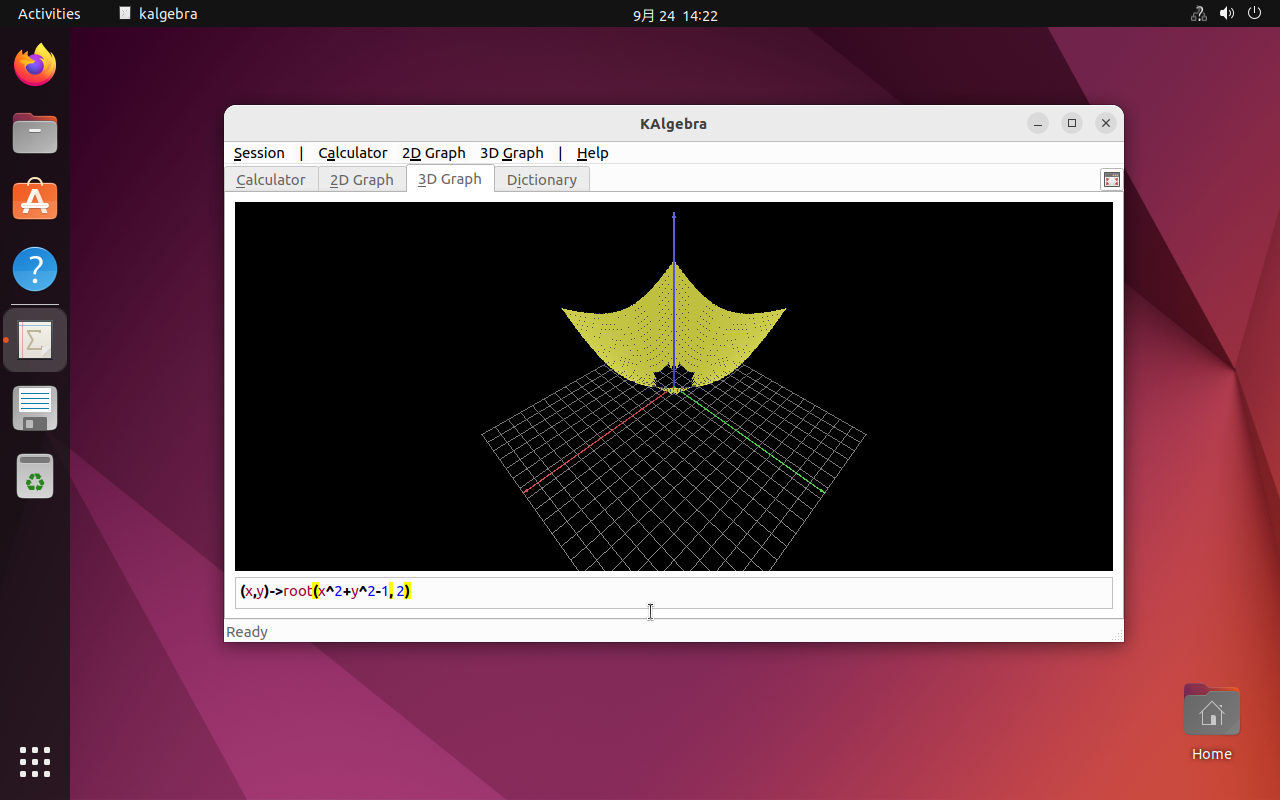}{\textbf{Initial library.} The single-branch result is incomplete.}{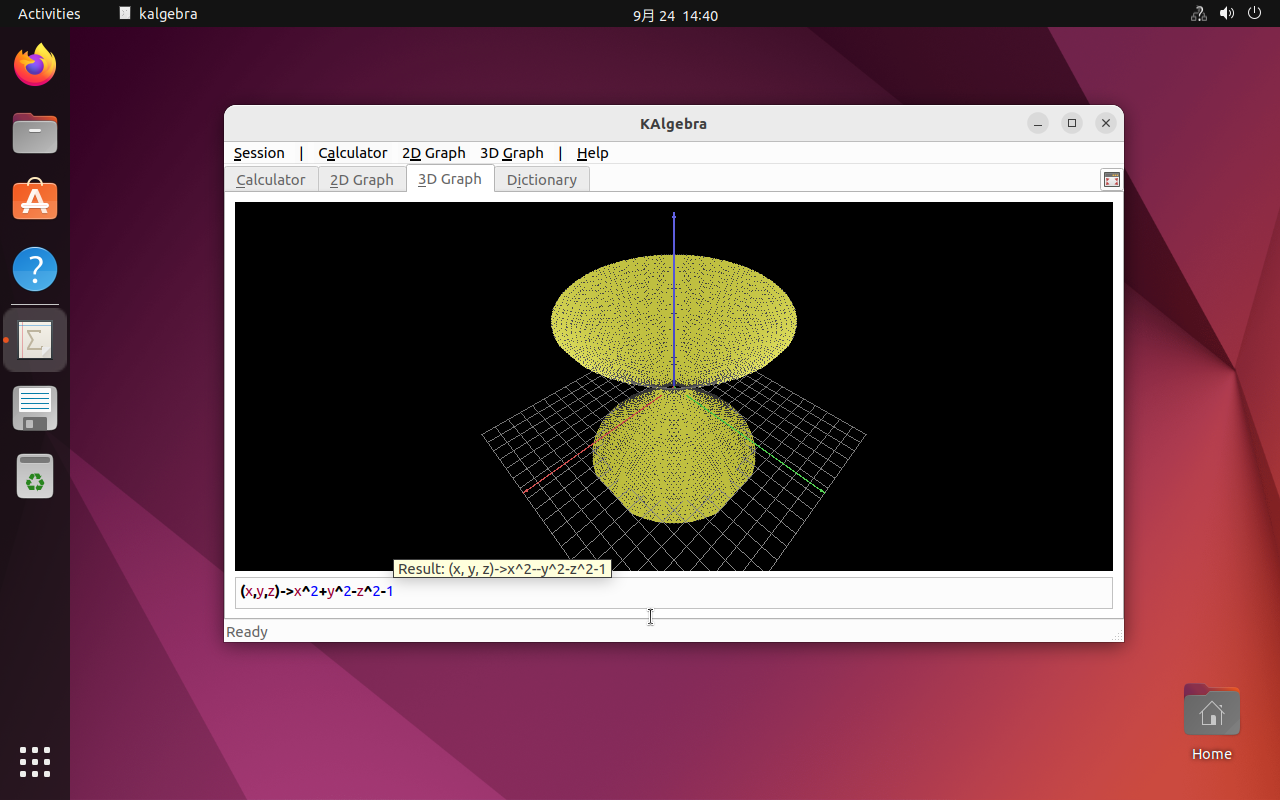}{\textbf{Updated Skill.} The complete hyperboloid is rendered; evaluation passes.}

\paragraph{C-14.} The initial explicit-branch attempt does not form the requested sphere. The updated Skill supports an implicit equation that produces the complete sphere.
\EvoPair{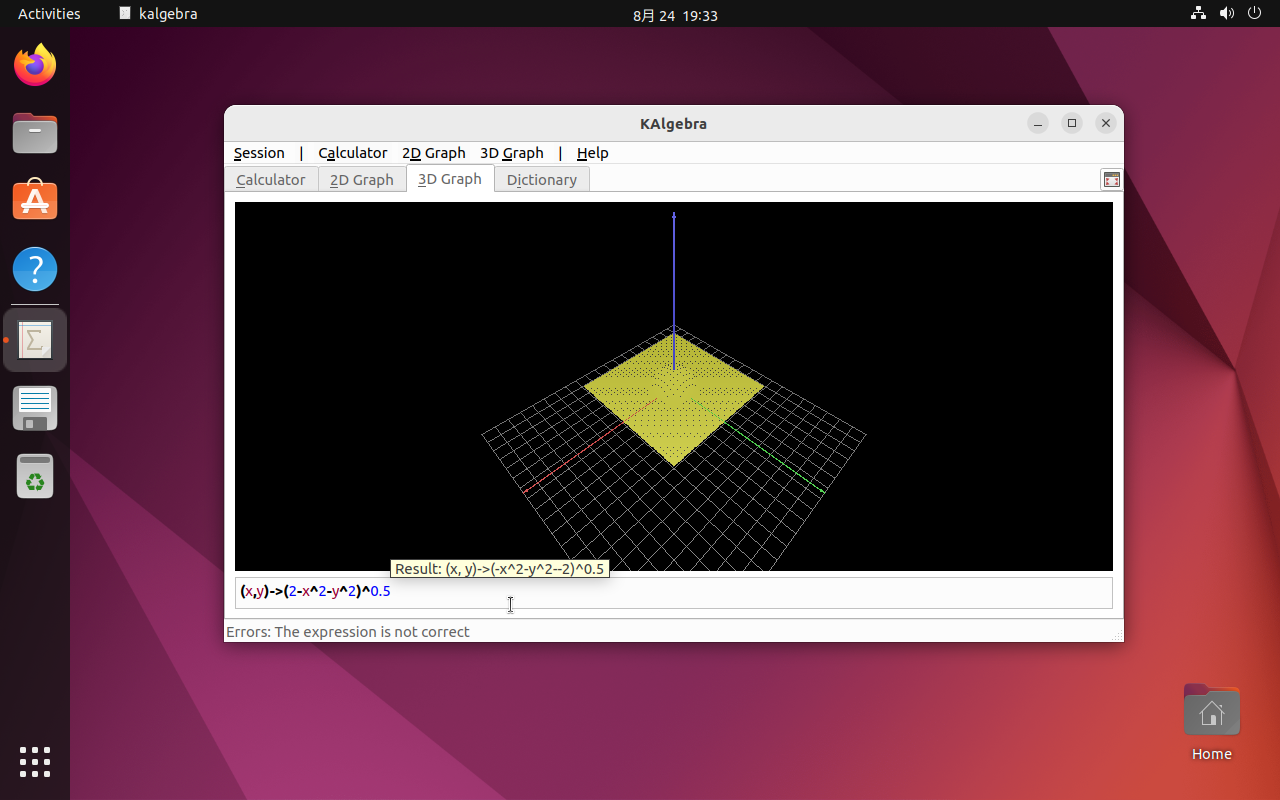}{\textbf{Initial library.} The requested complete sphere is not produced.}{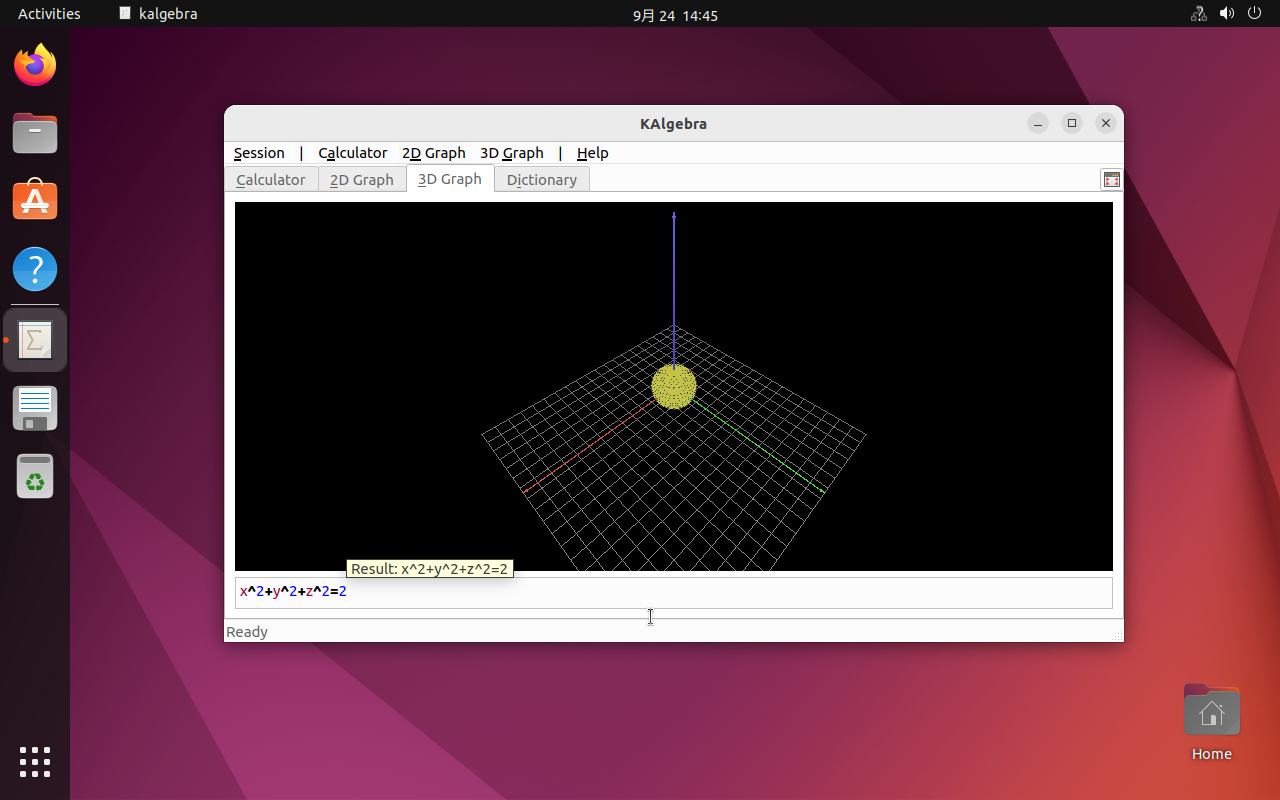}{\textbf{Updated Skill.} The implicit equation produces the sphere; evaluation passes.}

\paragraph{C-17.} The initial explicit-branch construction is incomplete. With the updated Skill, a three-variable mapping produces the required cylinder.
\EvoPair{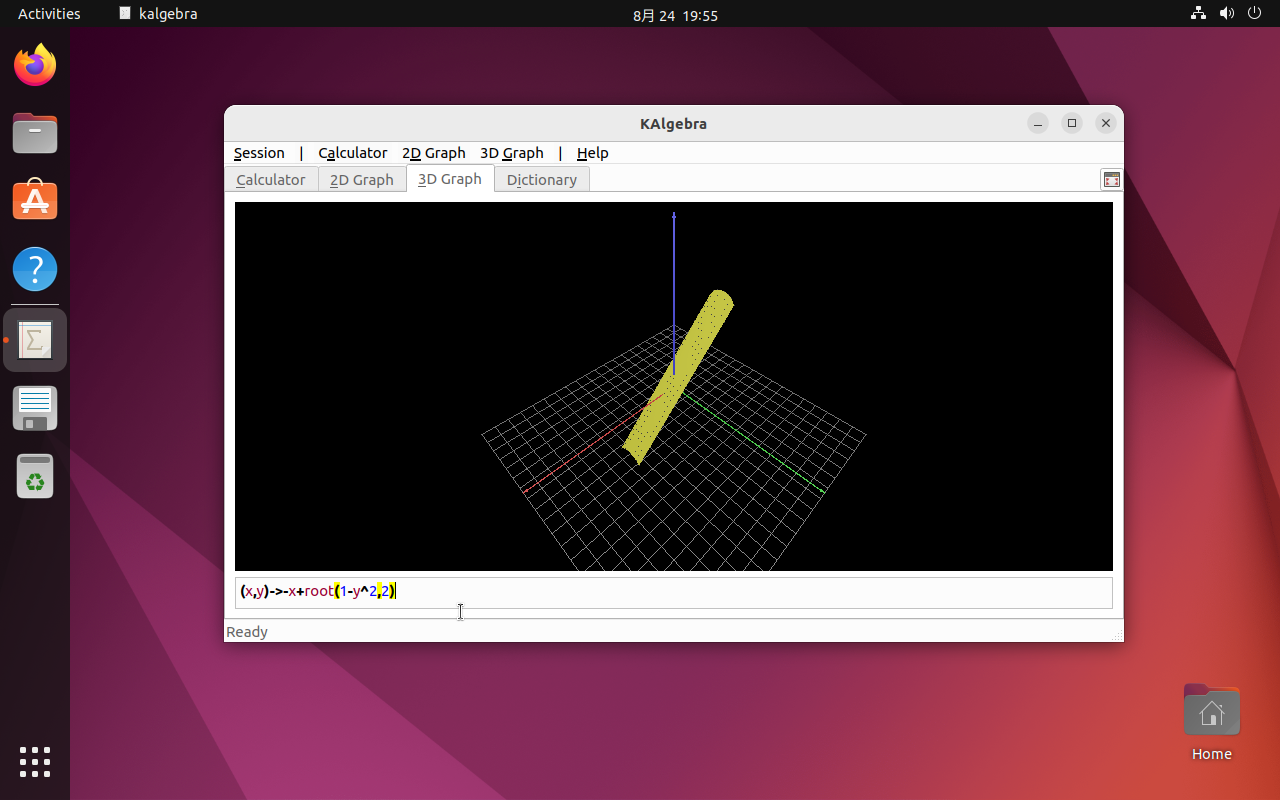}{\textbf{Initial library.} The explicit-branch construction is incomplete.}{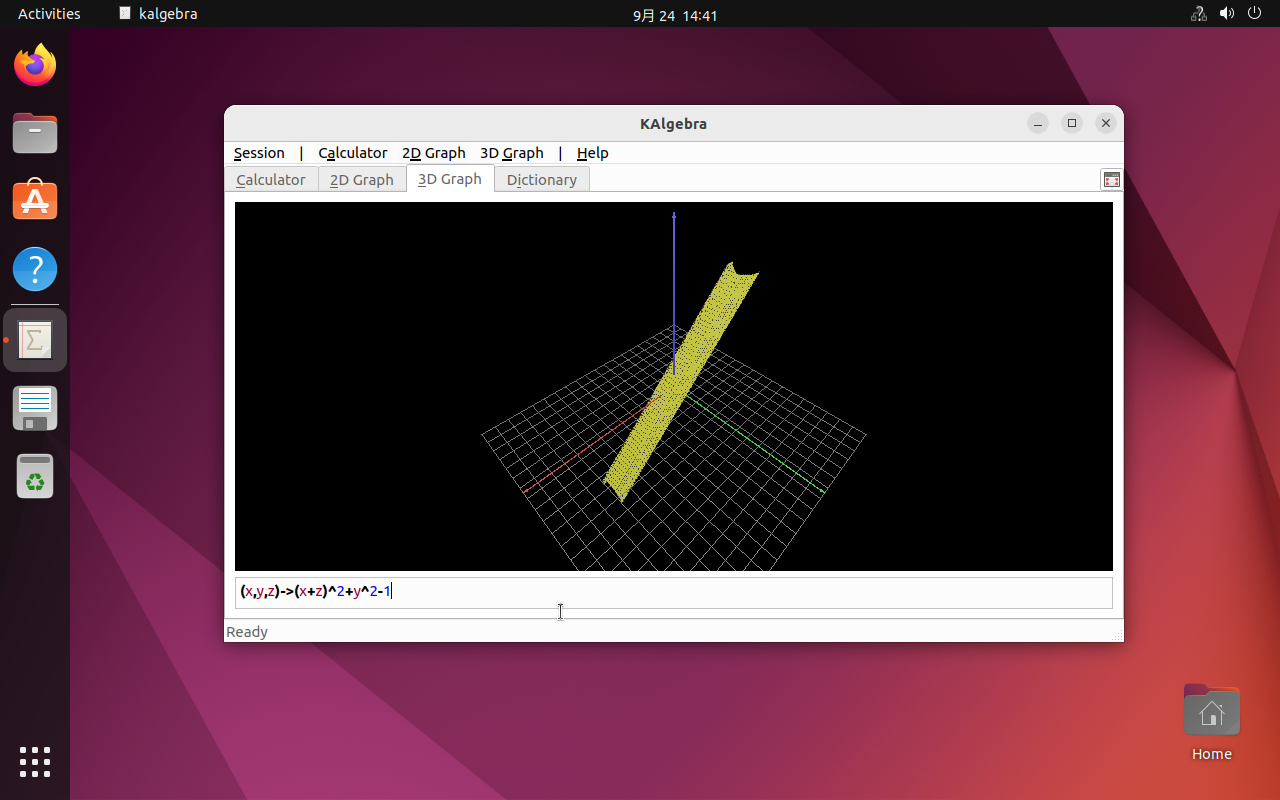}{\textbf{Updated Skill.} The required cylinder is produced; evaluation passes.}

These cases connect the resource mismatch to improvements on independent tasks. Outdated constraints first manifest as failures on Expanded--Verified tasks. Self-Evolve uses the failure history to discover procedures supported by the deployed environment and incorporates them into independently validated Skills. The updated Skill then helps three previously failed Verified tasks succeed, demonstrating both the need to revise initial resources through execution and the transfer of the resulting procedural knowledge.
\par
\endgroup

\section{Case Study: Adding a Module-Level Reference for GRASS GIS}
\label{app:case-grass-reference}

\paragraph{Why broad resource coverage leaves a task-relevant gap.} The GRASS GIS Reference Manual organizes software knowledge into twelve major sections. The benchmark tasks, however, are concentrated in a much smaller part of this space, primarily the Graphical User Interface (including Display), General, 3D raster processing, and Database sections. Resource-to-Skill runs before the downstream task distribution is known, so its initial construction aims to cover the software broadly. This coverage-oriented strategy can produce Skills and References for material that benchmark tasks rarely use, while leaving frequently needed leaf-level knowledge too general to execute. The \texttt{d.vect} command illustrates this mismatch: the initial Display catalog identifies the command but does not provide the syntax and parameter combinations needed by the recurring vector-styling tasks.

\begin{center}
\includegraphics[width=0.92\linewidth]{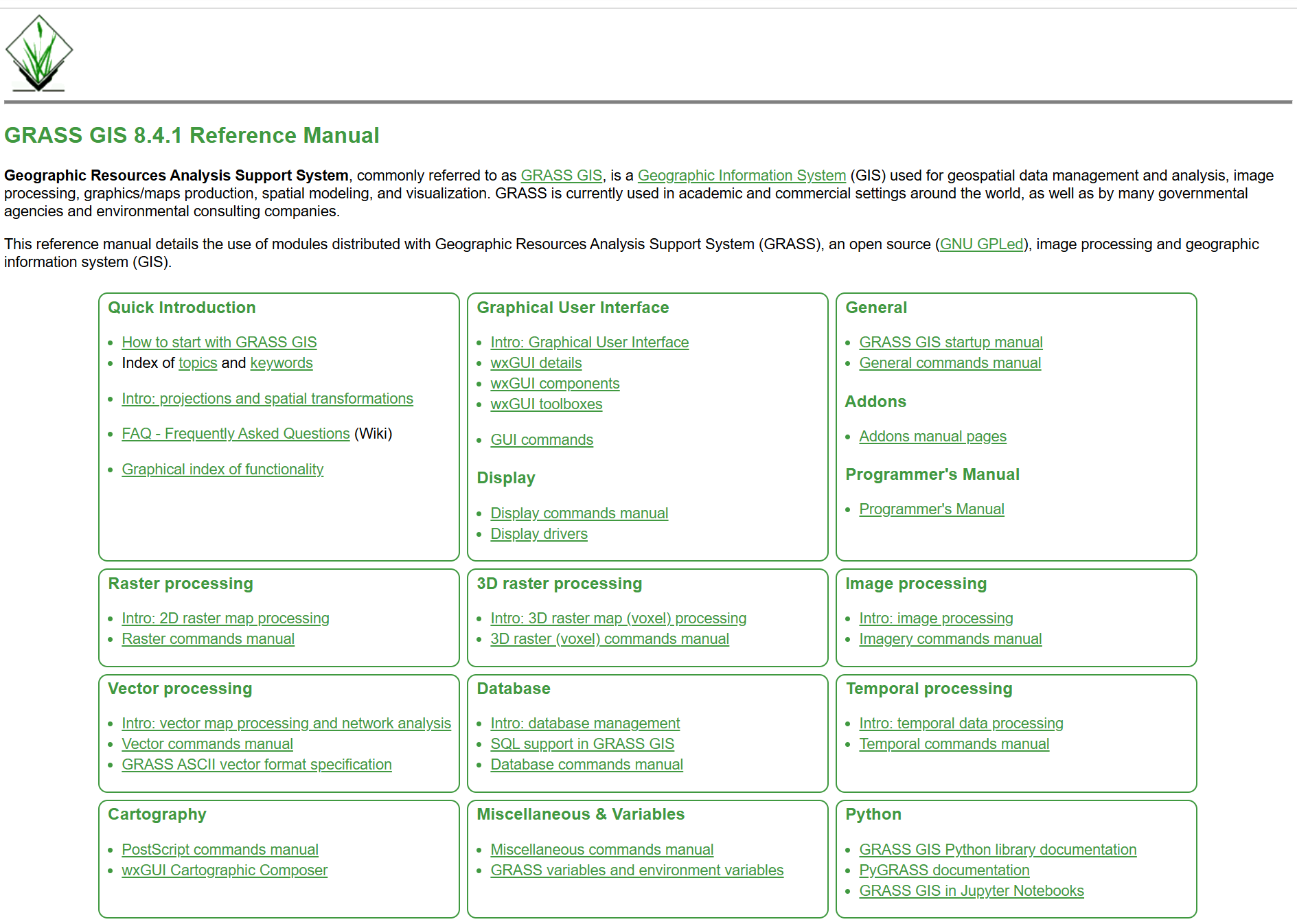}\\[-2pt]
{\scriptsize The manual spans twelve major sections; task demand is concentrated in a subset. The \texttt{d.vect} command is documented under Display.}
\end{center}

Prior Verified runs expose a recurring gap in this task family. B-04 requires an area fill color, B-06 a line width, and B-08 a point-symbol size. Their initial routed resources contained a broad Display command catalog, but the corresponding executions did not reach the required \texttt{d.vect} parameters. These failures motivate adding knowledge at the module level rather than writing a separate task-specific procedure for each map.

\paragraph{B-18: diagnosing and filling the knowledge gap.} B-18 is an Expanded--Verified task in the evolution set: \emph{``Display countries@PERMANENT as red boundary lines only in GRASS GIS.''} The initial resource set exposed the broad Display catalog, but not enough executable detail to compose the required command. Grok Build examined the failure evidence and the official GRASS manual, then selected a Reference rather than a Skill. The diagnosis was that the missing capability was declarative module knowledge---command syntax, option semantics, accepted values, defaults, and parameter composition---rather than an unlearned multi-step workflow. Grok Build browsed three pages from the same official manual. All 12 retained source-evidence records passed the later source audit, and the generator did not receive Verified task prompts or answers.

The resulting module-level Reference, \emph{d.vect Vector Display and Styling}, generalizes beyond the B-18 map and color. Its relevant guidance includes:

\begin{quote}
\small
\texttt{color=name} controls feature color; \texttt{fill\_color=name} controls area fill. Areas use \texttt{fill\_color} for their fill and \texttt{color} for their outline. Use \texttt{fill\_color=none} to make areas transparent, or \texttt{color=none} to suppress area outlines.\\
\texttt{width=integer} sets line width (default: 0).\\
\texttt{size=float} sets point-symbol size (default: 5).\\
\texttt{attribute\_column=name} selects the label column; \texttt{font}, \texttt{xref}, and related options control label appearance and placement.
\end{quote}

This knowledge is directly relevant to several task families: independent outline and fill controls support B-18 and B-04; line-width and symbol-size options support B-06 and B-08; label parameters provide a reusable route for other vector-styling requests. The Reference describes reusable \texttt{d.vect} behavior rather than a fixed answer to B-18 in the evolution set: its options can be composed for other maps, colors, and styling requests.

\paragraph{B-18 validation: applying the added module knowledge.} During validation, the candidate Reference is force-injected into B-18's resource set. Using the command syntax and independent outline and fill controls provided by the Reference, the agent constructs the command below and executes it in the Command Console. The map is rendered with red boundaries and no area fill, and B-18 passes. \textbf{The added Reference supplies the parameter knowledge missing from the initial catalog, enabling the agent to compose the required command.}

\begin{quote}
\small\ttfamily
d.vect map=countries@PERMANENT type=boundary color=red fill\_color=none
\end{quote}

\begin{center}
\begin{minipage}[t]{0.72\linewidth}
\centering
\includegraphics[width=\linewidth]{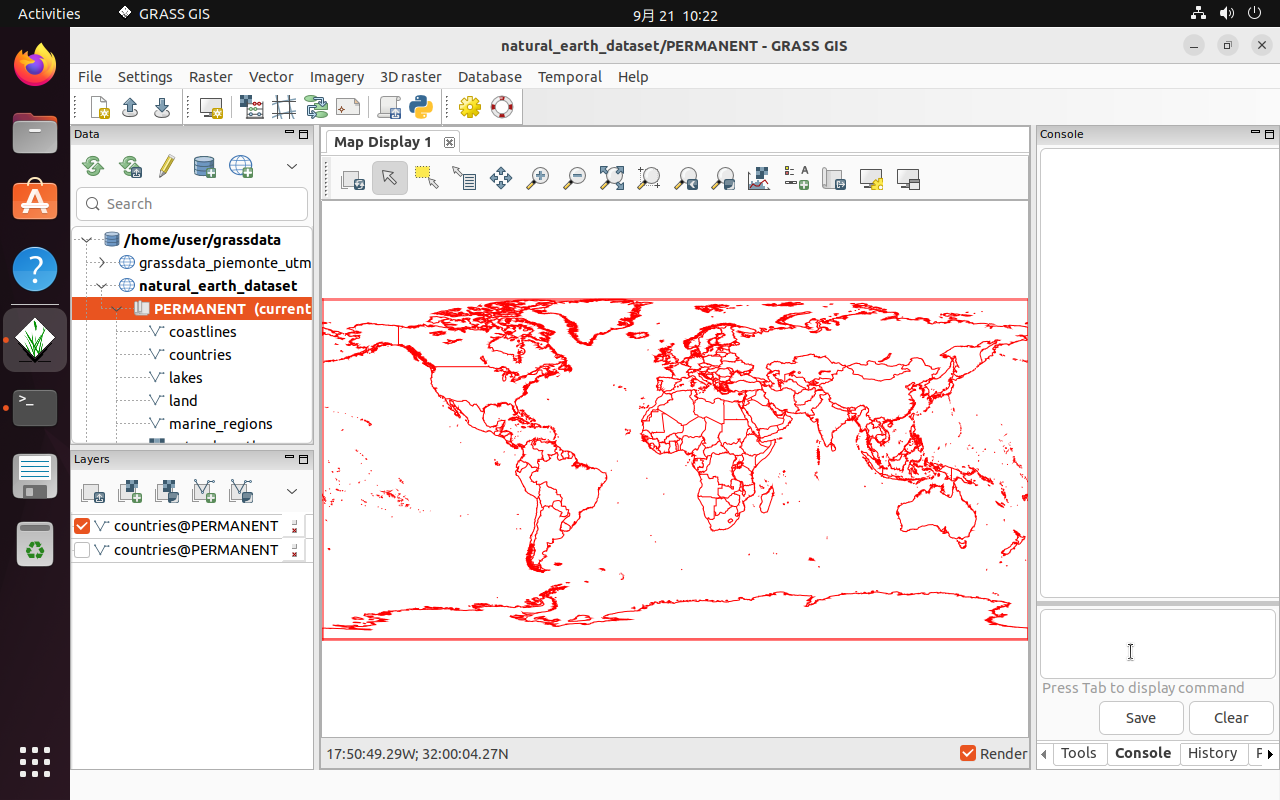}\\[-2pt]
{\scriptsize B-18: using the added Reference, the agent renders red boundaries with no area fill.}
\end{minipage}
\end{center}

\paragraph{Transfer to Verified: three related styling tasks.} B-04, B-06, and B-08 all failed in earlier Verified runs with the initial library. After validation, the new \texttt{d.vect} Reference was made available in the evolved library. In the subsequent Verified evaluation, resource selection was free: each task routed to the new Reference. The successful runs used three different parameter families:

\begin{center}
\small
\begin{tabular*}{\linewidth}{@{\extracolsep{\fill}}p{0.10\linewidth}p{0.27\linewidth}p{0.57\linewidth}@{}}
\toprule
Task & Initial gap & Command in a successful transfer run \\
\midrule
B-04 & Could not set area fill color & \texttt{d.vect map=boundary\_region fill\_color=191:191:191} \\
B-06 & Did not set \texttt{width=2} & \texttt{d.vect map=railroads width=2} \\
B-08 & Did not set \texttt{size=10} & \texttt{d.vect map=hospitals size=10} \\
\bottomrule
\end{tabular*}
\end{center}

\begin{center}
\begin{minipage}[t]{0.72\linewidth}
\centering
\includegraphics[width=\linewidth]{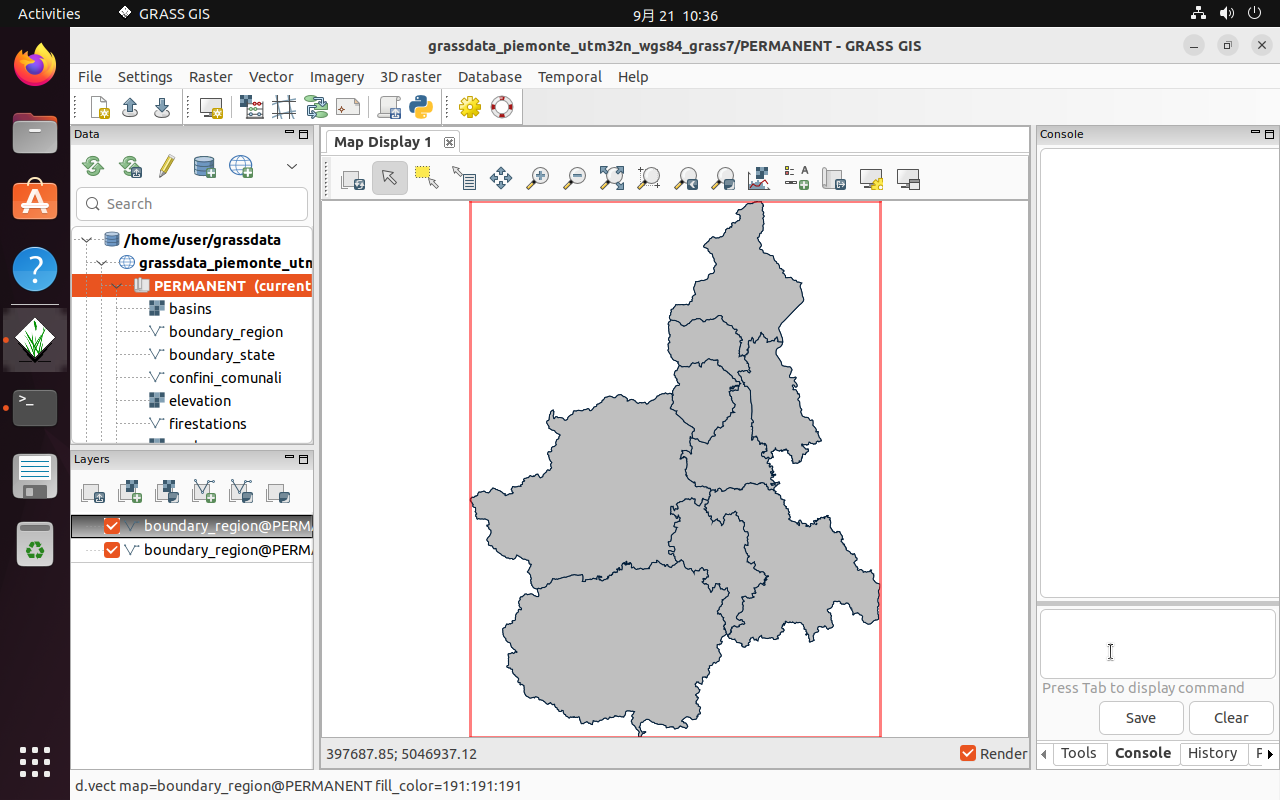}\\[-2pt]
{\scriptsize B-04: the area is gray after applying \texttt{fill\_color=191:191:191}.}
\end{minipage}
\end{center}

\begin{center}
\begin{minipage}[t]{0.49\linewidth}
\centering
\includegraphics[width=\linewidth]{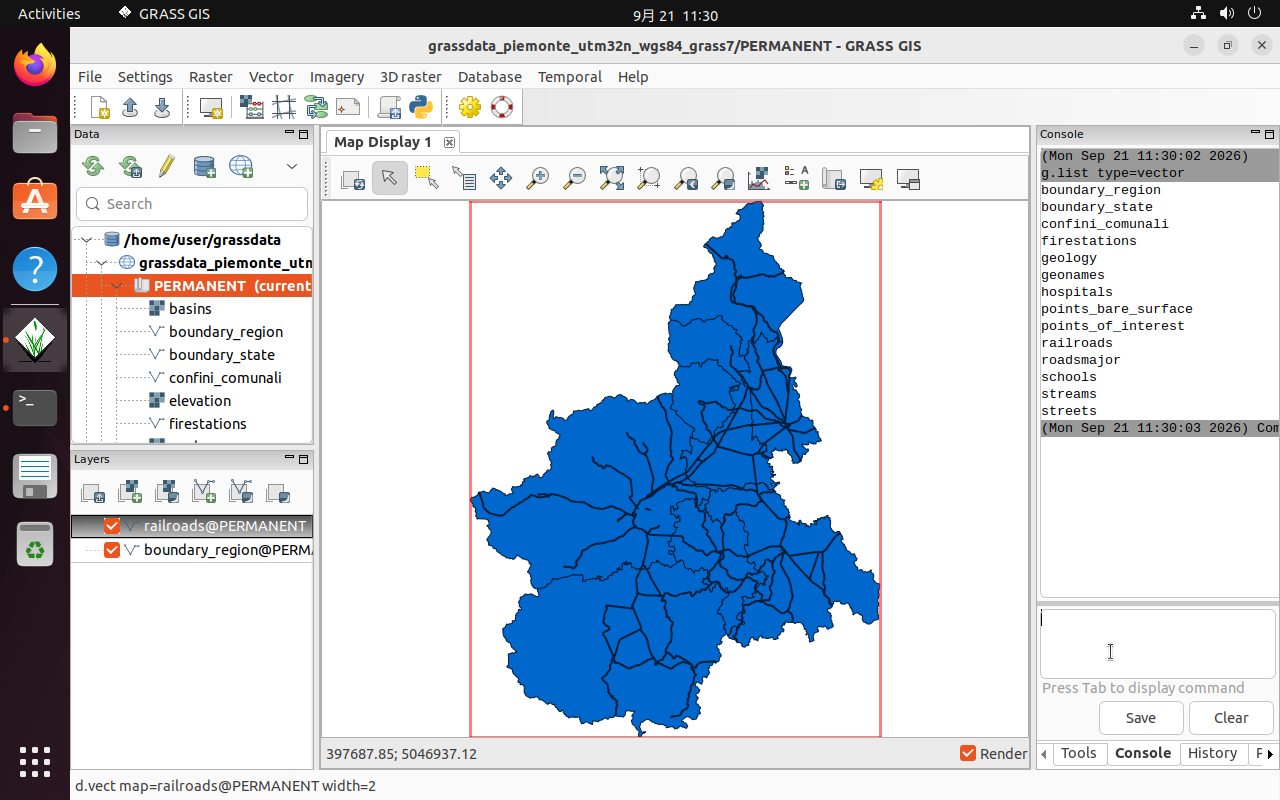}\\[-2pt]
{\scriptsize B-06: the Console executes \texttt{width=2}.}
\end{minipage}\hfill
\begin{minipage}[t]{0.49\linewidth}
\centering
\includegraphics[width=\linewidth]{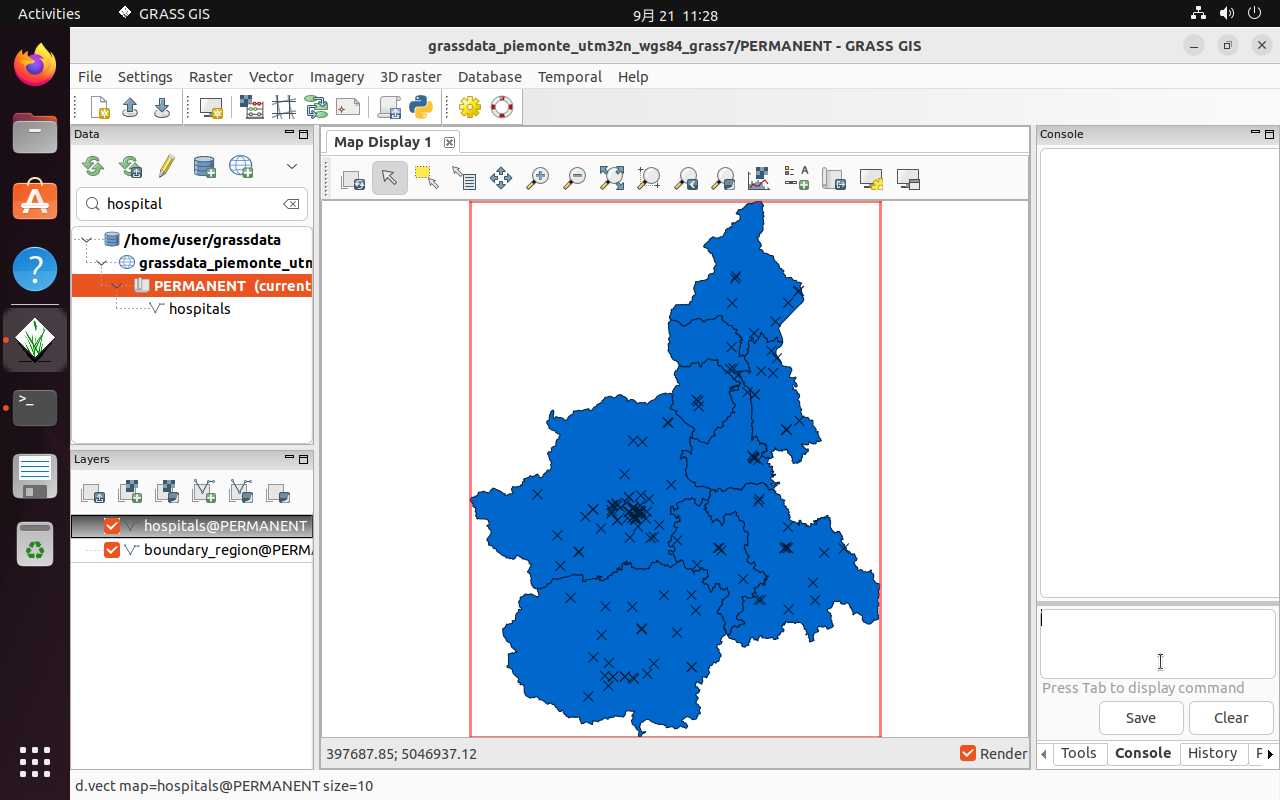}\\[-2pt]
{\scriptsize B-08: the Console executes \texttt{size=10}.}
\end{minipage}
\end{center}

\paragraph{Takeaway.} This case shows how Self-Evolve addresses the mismatch between the broad scope of source resources and the specific knowledge required by downstream tasks through knowledge addition. The initial library covered GRASS GIS broadly but omitted the \texttt{d.vect} details required by these tasks. Execution feedback from the evolution set guided Grok Build to add a reusable module-level Reference. \textbf{The added knowledge filled this task-relevant gap, enabling success on B-18 and related Verified tasks that the initial library had failed to support.}

\newpage
\begingroup
\raggedbottom
\section{Case Study: Outdated Interface Guidance in Celestia}
\label{app:case-celestia-a05}

Task A-05 asks the agent to display dwarf-planet orbits but not moon orbits in Celestia. We examine an initial-library run by Grok-4.6 (xhigh), before any Skill or Reference evolution. The first-stage router selected the Skill \emph{Control rendering features and display quality with keyboard shortcuts} and the Reference \emph{Celestia label, info-text, constellation, and star-filter controls}. The source User's Guide describes Celestia 1.6.1, whereas the target environment runs Celestia 1.7.0 with a Qt interface.

\paragraph{From source guidance to the target interface.}
The selected Reference places the checkboxes for orbit categories in the lower part of a \emph{View Options} window and also mentions the \texttt{O} key for orbit paths. The selected Skill condenses this guidance to pressing \texttt{O} to toggle planet orbits; it does not explain how to choose dwarf-planet or moon orbits separately in the target interface. Figure~\ref{fig:celestia-a05-source} shows the old \emph{View Options} window in the source guide. In the deployed Qt interface, neither the Display nor the View menu contains this window. Instead, the toolbar's \texttt{O} dropdown exposes separate checkboxes for \emph{Dwarf Planets} and \emph{Moons} (Figure~\ref{fig:celestia-a05-target}). The target-interface screenshot is an independent illustration of the available control, not a frame from the failed run.

\noindent\begin{minipage}{\linewidth}
    \begin{minipage}[t]{0.48\linewidth}
        \vspace{0pt}
        \centering
        \includegraphics[width=\linewidth,trim=570bp 375bp 80bp 130bp,clip]{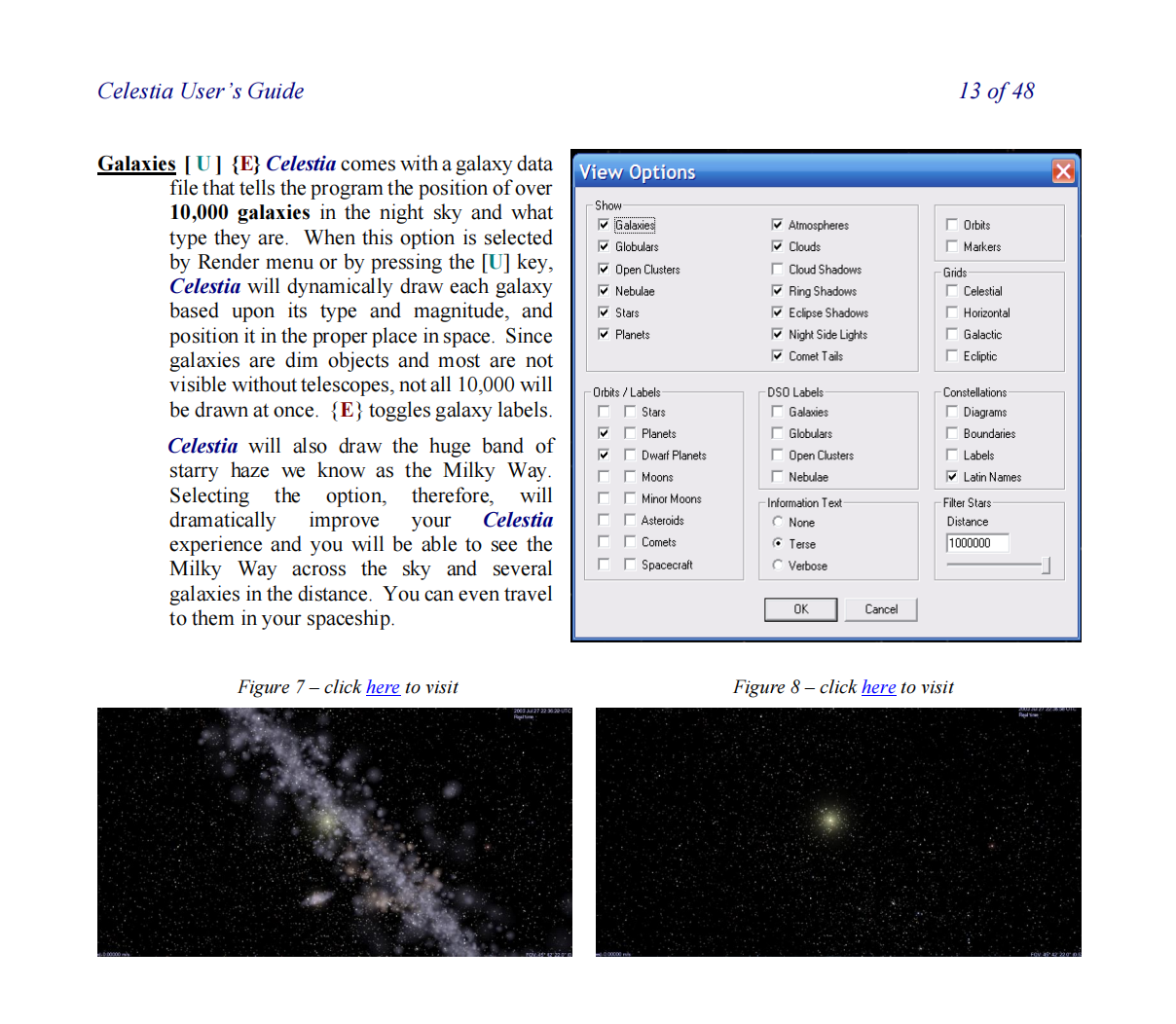}
        \captionof{figure}{The \emph{View Options} window shown in the collected Celestia 1.6.1 User's Guide. The selected Reference retains this old location for orbit-category controls.}
        \label{fig:celestia-a05-source}
    \end{minipage}\hfill
    \begin{minipage}[t]{0.48\linewidth}
        \vspace{0pt}
        \centering
        \includegraphics[width=\linewidth,trim=470bp 440bp 400bp 40bp,clip]{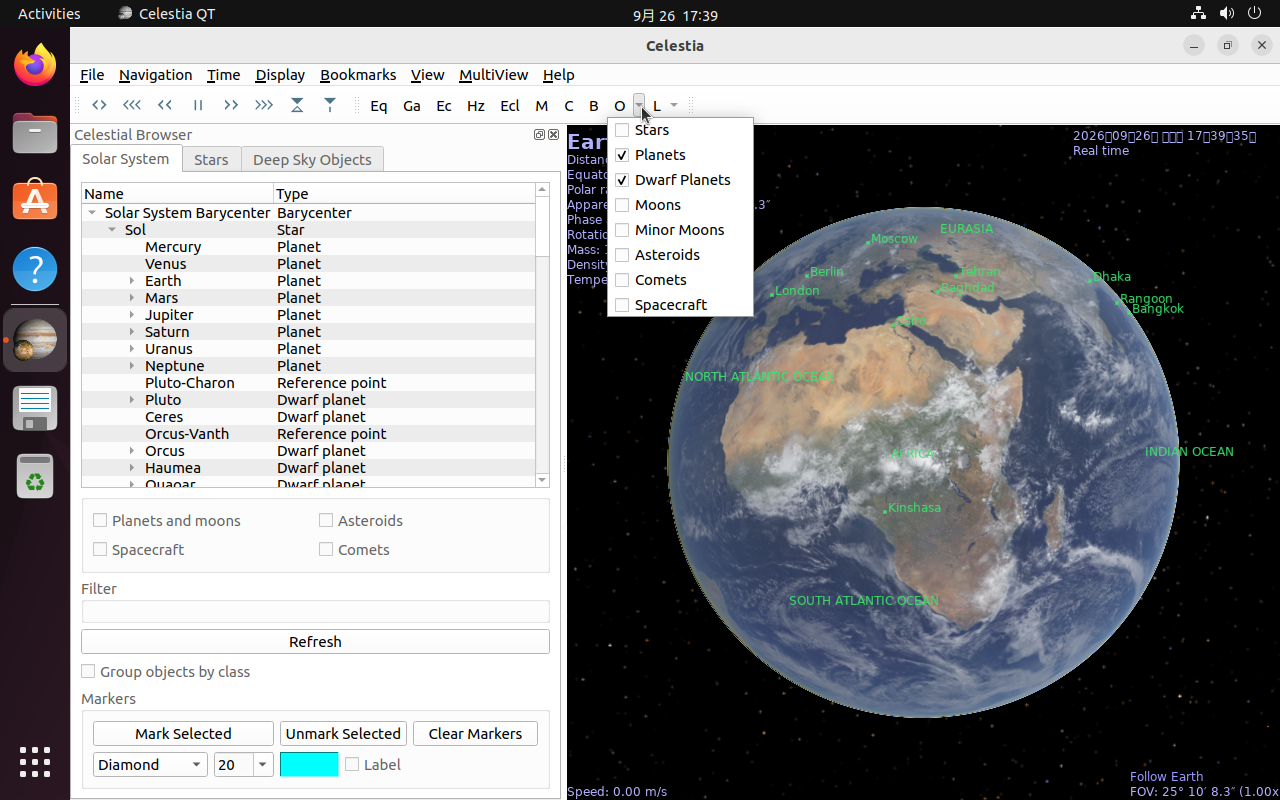}
        \captionof{figure}{The target Celestia 1.7.0 Qt interface. The toolbar's \texttt{O} dropdown allows dwarf-planet and moon orbits to be selected independently; the screenshot shows \emph{Dwarf Planets} checked and \emph{Moons} unchecked. This image illustrates the available control and is separate from the failed trajectory.}
        \label{fig:celestia-a05-target}
    \end{minipage}
\end{minipage}\par\smallskip

\paragraph{Execution with the initial Skill and Reference.}
Figure~\ref{fig:celestia-a05-trace} presents every screenshot from the five-action run. Each screenshot is a pre-action observation; its caption describes the action chosen next. Although the toolbar's \texttt{O} control is visible in the initial state, the agent first states that Display should contain \emph{View Options} and opens that menu. Finding no such entry, it moves to View, retries View with \texttt{Alt+V}, and then selects Guides as another possible location for orbit-path options. That action hides the toolbar. Only in its final response does the agent identify the \texttt{O} dropdown, but it attempts to restore the toolbar rather than configure the two orbit categories. The run ends without completing A-05, and the task evaluator reports failure.

\newcommand{\CelAZeroFivePanel}[2]{%
\begin{minipage}[t]{0.485\linewidth}
\centering\includegraphics[width=\linewidth]{Figure/celestia_a05/#1}\par\vspace{1pt}
\raggedright\scriptsize #2
\end{minipage}}

\noindent\begin{minipage}{\linewidth}
    \centering
    \CelAZeroFivePanel{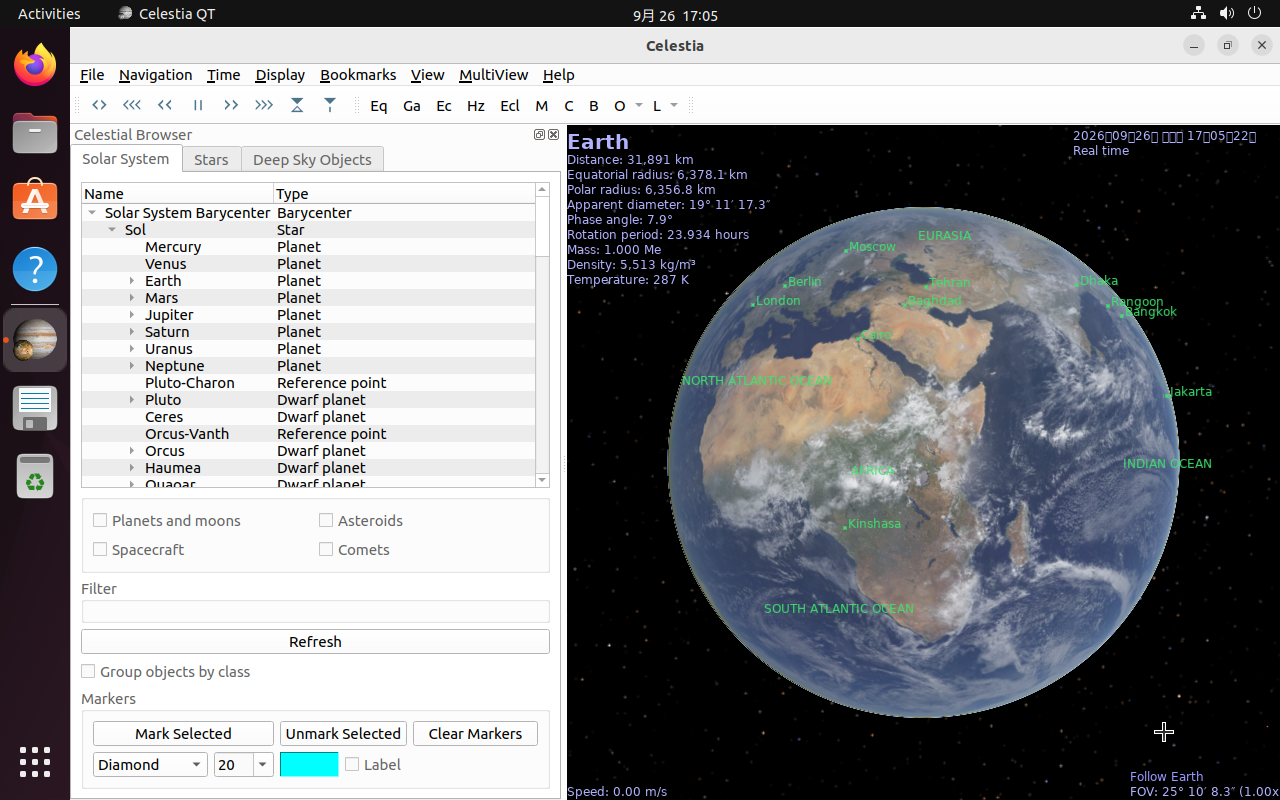}{\textbf{(a) Initial state.} The toolbar's \texttt{O} control is visible; the agent chooses to open Display.}\hfill
    \CelAZeroFivePanel{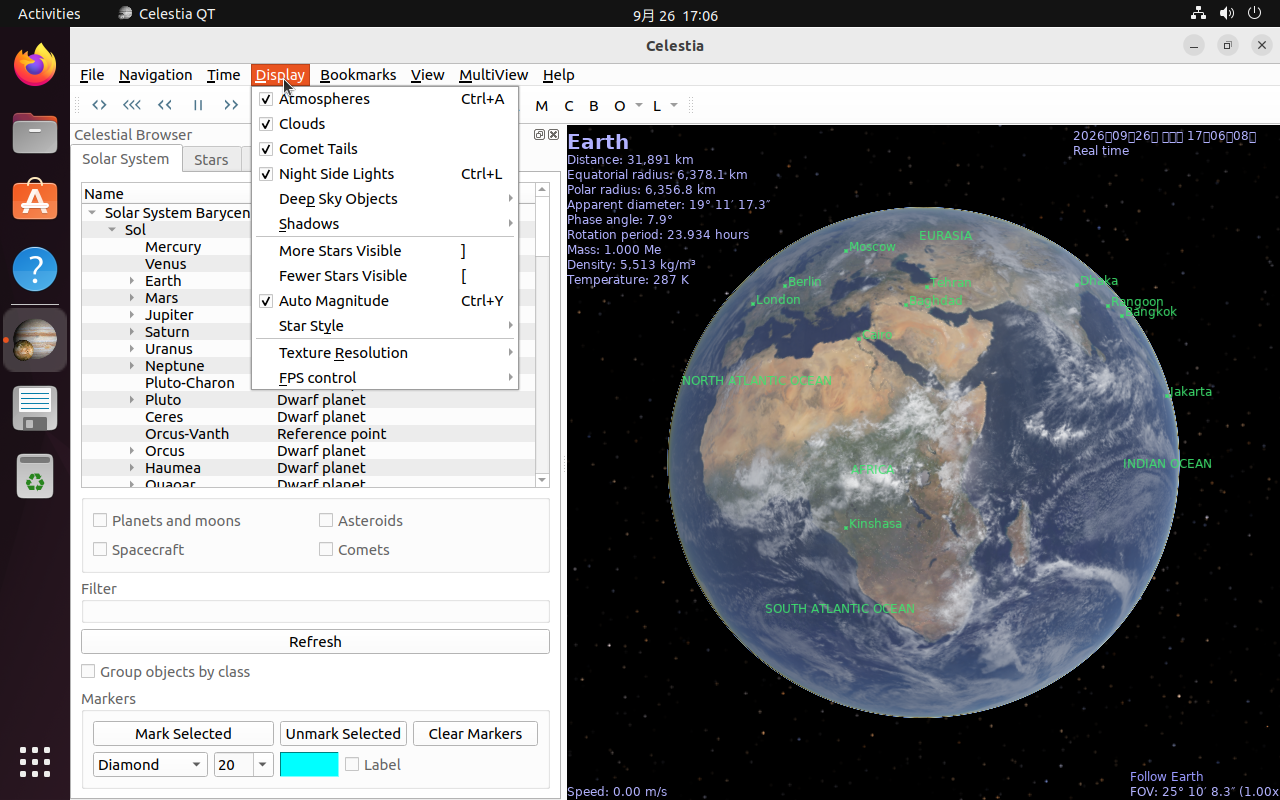}{\textbf{(b) Display menu.} No \emph{View Options} appears; the agent turns to View.}

    \vspace{7pt}
    \CelAZeroFivePanel{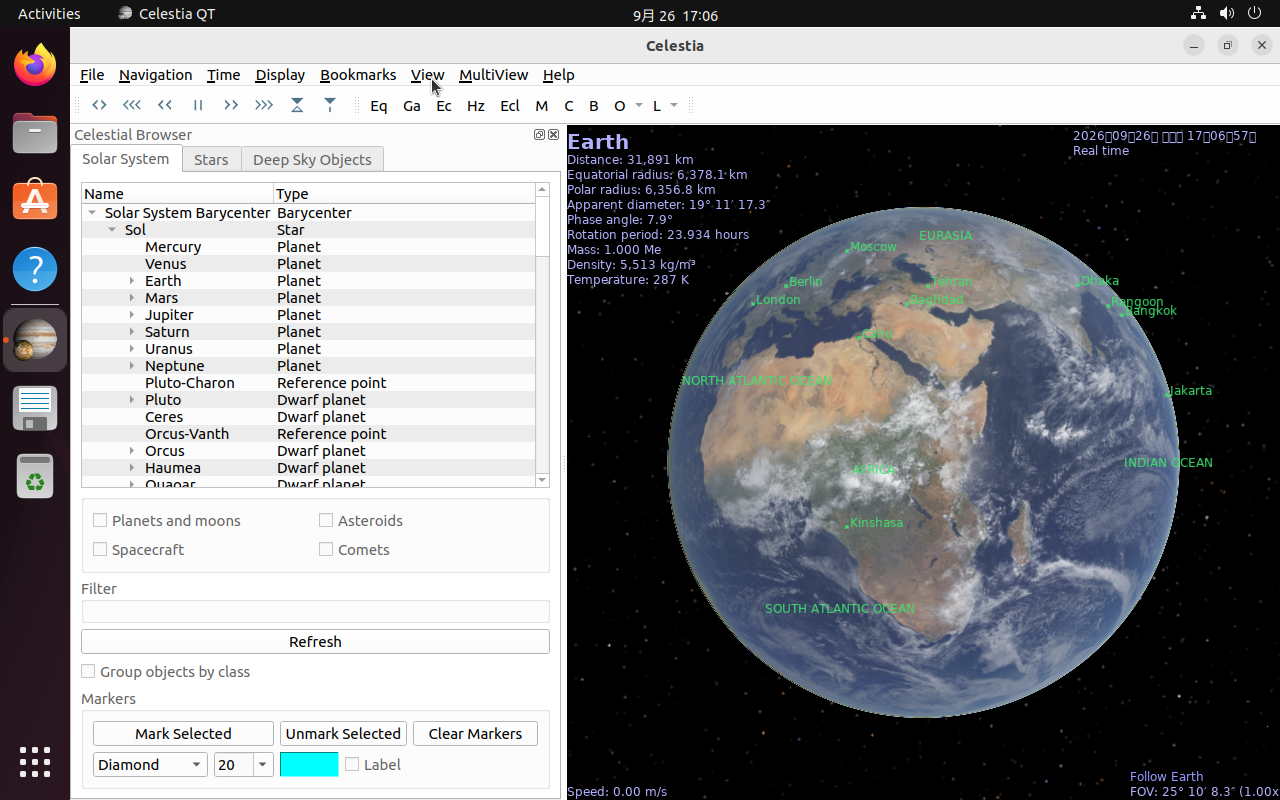}{\textbf{(c) After clicking View.} The menu is not open; the agent retries with \texttt{Alt+V}.}\hfill
    \CelAZeroFivePanel{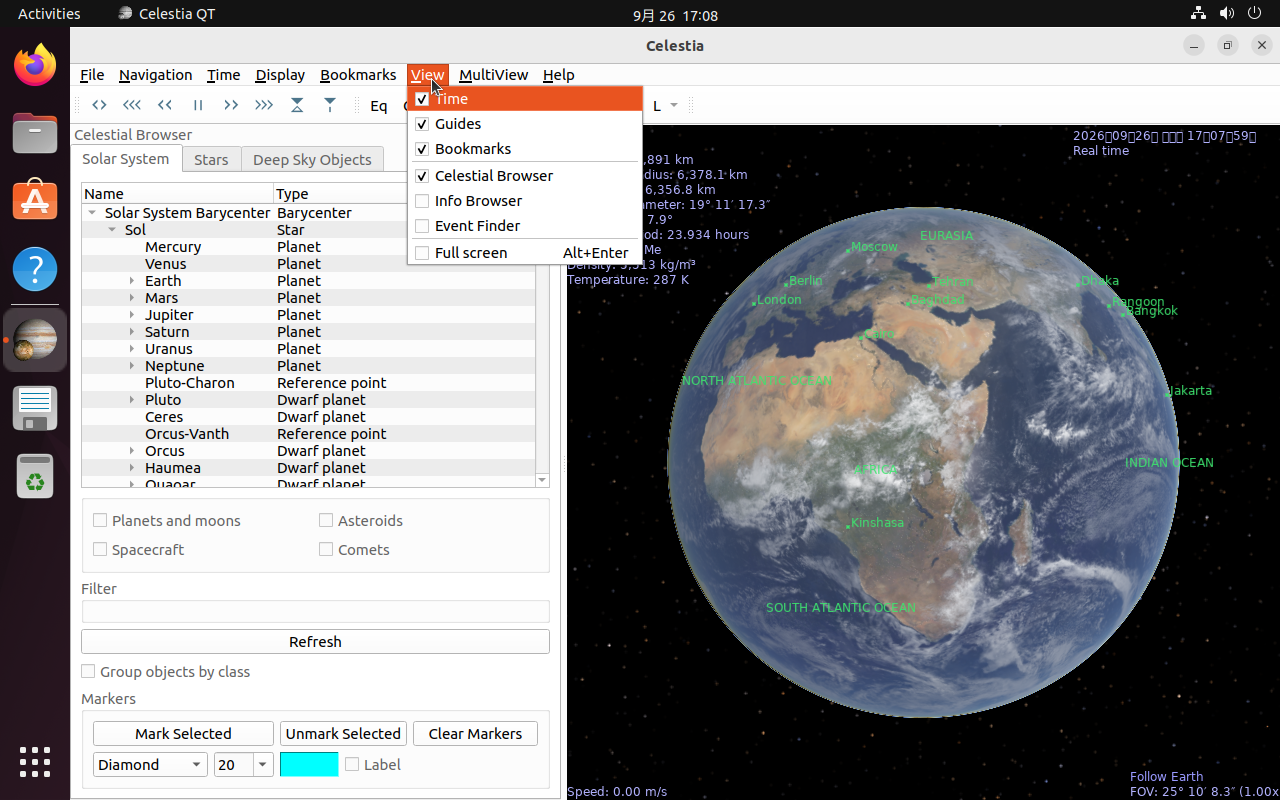}{\textbf{(d) View menu.} The old entry is absent here too; the agent selects Guides.}

    \vspace{7pt}
    \CelAZeroFivePanel{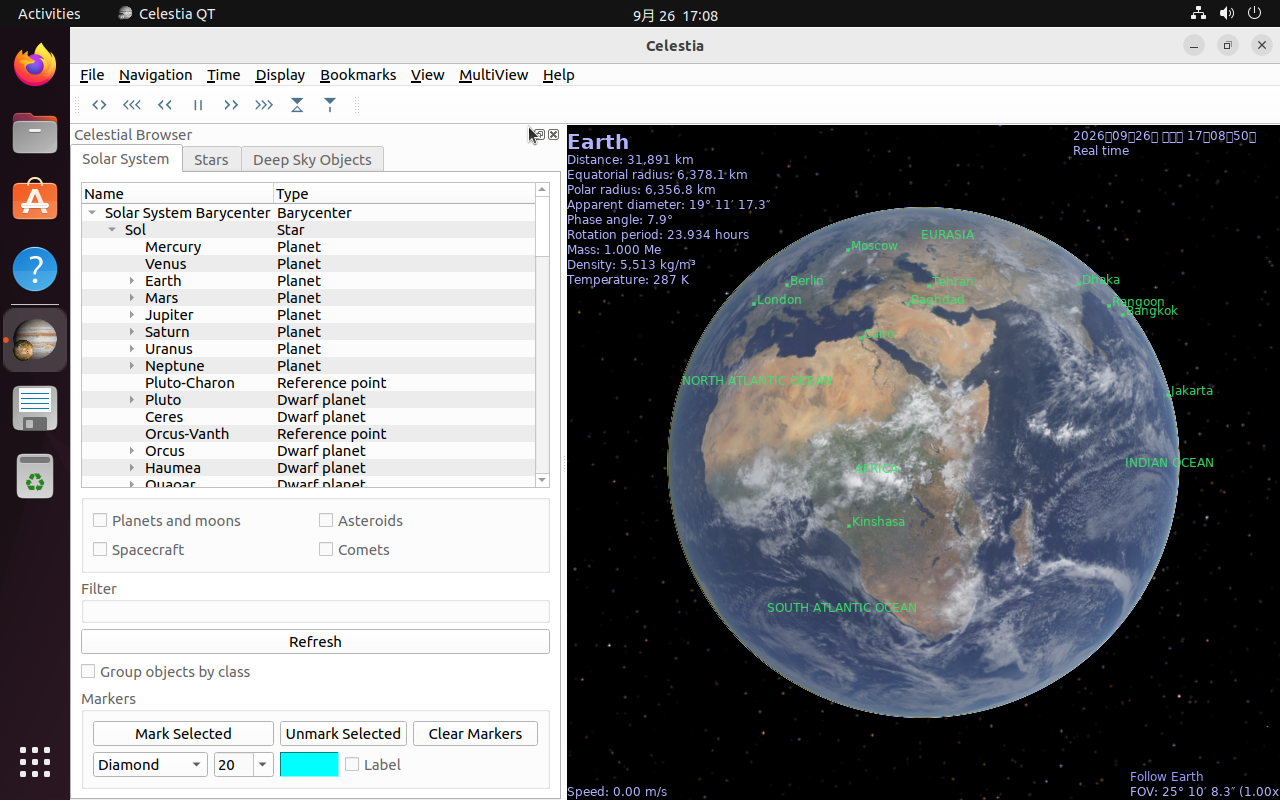}{\textbf{(e) After selecting Guides.} The toolbar has disappeared; the agent next tries to restore it.}
    \captionof{figure}{The complete five-step trajectory of Grok-4.6 (xhigh) with the initial Celestia Skill and Reference on A-05. Panels show observations before the actions described beneath them. The agent spends its actions searching old menu paths and never sets the required orbit categories; evaluation fails.}
    \label{fig:celestia-a05-trace}
\end{minipage}\par\smallskip

\paragraph{Implication.}
This run makes the cost of outdated knowledge concrete. Guidance inherited from the old interface directs the agent toward Display, View, and Guides, consuming its limited action budget even though the target's \texttt{O} dropdown is visible from the start. The agent recognizes that control only at the end and does not complete the category-specific orbit settings. Thus, an initial Skill and Reference grounded in documentation can still waste execution steps and lead to task failure when their operational guidance does not match the deployed software.
\clearpage
\endgroup

\section{Task Requirements and Documentation Scope for Lean 4 and TeXstudio}
\label{app:lean-texstudio-scope}

This appendix compares the Verified task requirements for Lean 4 and TeXstudio with the documentation used to construct their initial Skills and References. Tables~\ref{tab:lean4-task-scope} and~\ref{tab:texstudio-task-scope} summarize the task goals; the full task instructions are provided with ScienceBoard-VE. The comparison helps explain why additional knowledge from the same software-specific sources provides more limited gains for these applications in Table~\ref{tab:self_evolution_ablation}.

\subsection{Lean 4: General Proof Guidance and Domain-Specific Theorems}
\label{app:lean4-task-scope}

The Lean 4 initial library is constructed from \emph{Theorem Proving in Lean 4} (\url{https://leanprover.github.io/theorem_proving_in_lean4/}). As Figure~\ref{fig:lean4-source-scope} shows, its contents cover general proof methods, including propositions, quantifiers, tactics, induction, and type classes. The Verified tasks in Table~\ref{tab:lean4-task-scope}, by contrast, ask the agent to complete particular formal statements across set theory, algebra, topology, analysis, measure theory, and other mathematical domains. Completing them requires identifying applicable domain definitions and lemmas and assembling proofs that match the given Lean declarations.

\noindent\begin{minipage}{\linewidth}
    \centering
    \includegraphics[width=0.95\linewidth]{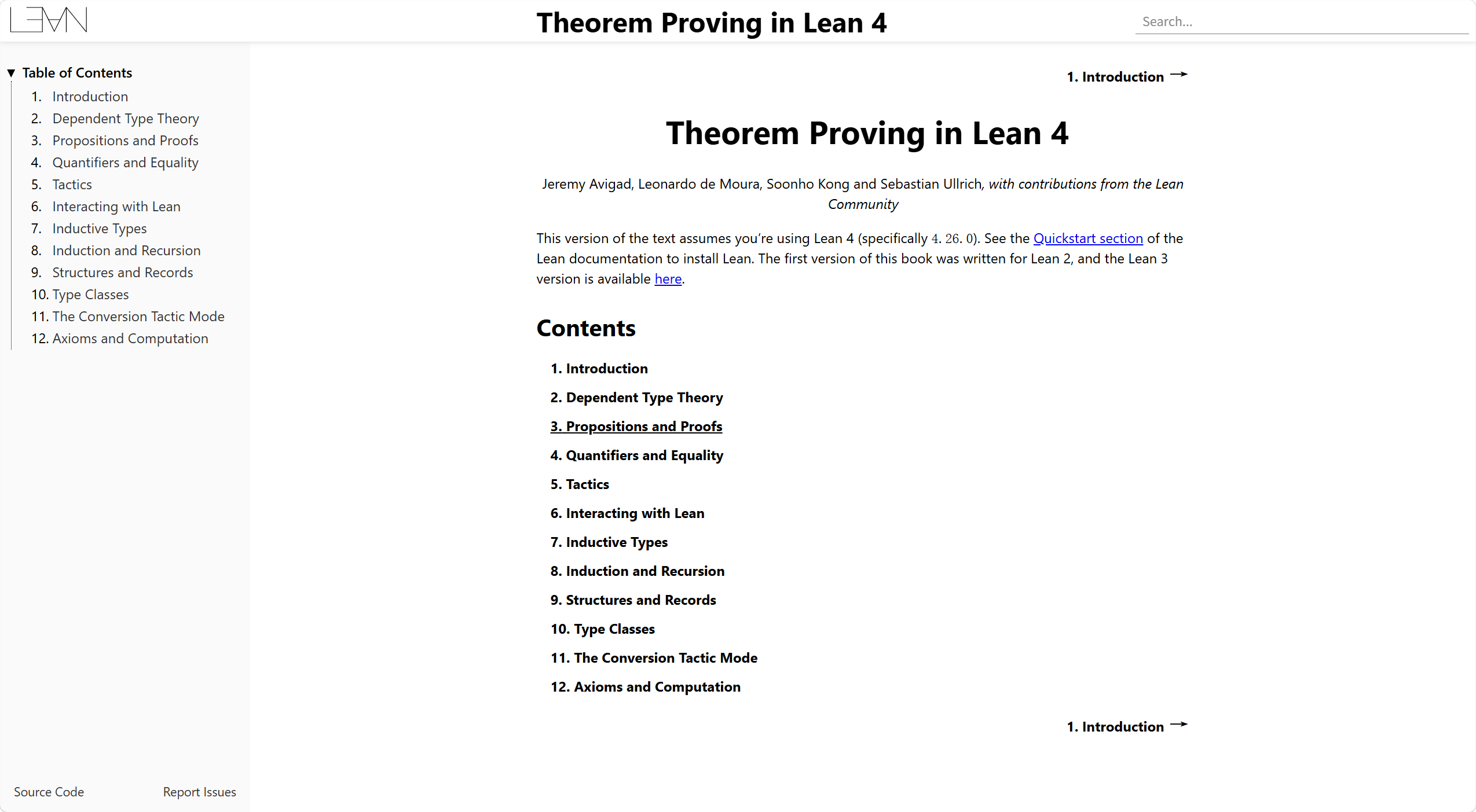}
    \captionof{figure}{Landing page and table of contents of \emph{Theorem Proving in Lean 4}, the source used for the initial Lean 4 library.}
    \label{fig:lean4-source-scope}
\end{minipage}\par\smallskip

\begingroup
\small
\setlength{\LTpre}{4pt}
\setlength{\LTpost}{4pt}
\renewcommand{\arraystretch}{1.08}
\begin{longtable}{@{}p{0.10\linewidth}p{0.84\linewidth}@{}}
\caption{Lean 4 Verified tasks and their proof goals. The entries summarize the full formal task statements supplied with ScienceBoard-VE.}\label{tab:lean4-task-scope}\\
\toprule
Task & Proof goal \\
\midrule
\endfirsthead
\multicolumn{2}{@{}c@{}}{\small\itshape Table \thetable{} (continued)}\\
\toprule
Task & Proof goal \\
\midrule
\endhead
\bottomrule
\endlastfoot
A-01 & Derive reflexivity from symmetry, transitivity, and an existence condition for a relation. \\
A-02 & Prove a recurrence inequality for the Ackermann function. \\
A-03 & Relate transitivity of a ZF set to powerset and union conditions. \\
A-04 & Prove a cofinality property of a limit ordinal. \\
A-05 & Prove a statement involving cardinal operations. \\
B-01 & Prove a modular congruence under primality and coprimality assumptions. \\
B-03 & Prove an embedding statement for isomorphic structures. \\
B-04 & Establish a property of many-one equivalence under \texttt{Sum.elim}. \\
B-05 & Prove that a quotient of a cyclic group is cyclic. \\
B-06 & Prove a statement about an ideal and a quotient division ring. \\
B-07 & Prove a statement about a normal field extension and an intermediate field. \\
C-01 & Prove that metrizability is preserved under homeomorphism. \\
C-02 & Establish closedness from a derived-set inclusion condition. \\
C-03 & Prove connectedness of the continuous image of a connected set. \\
C-04 & Relate openness conditions between topologies to the $T_0$ and $T_1$ properties. \\
C-05 & Prove compactness of a closed subset of a compact set. \\
D-01 & Derive a quantitative bound from a function limit. \\
D-02 & Prove uniform continuity under a bounded-derivative condition. \\
D-03 & Prove that a continuous function vanishes when the integral of its square is zero. \\
D-04 & Establish the cardinality of a countably generated set algebra. \\
D-05 & Construct a measure from the limit of an increasing sequence of measures. \\
D-06 & Establish convergence in measure after composition with a continuous function. \\
D-07 & Prove a limit statement involving a random variable's moment and tail probability. \\
E-01 & Construct a decidability instance for representation as a sum of four integer cubes. \\
E-02 & Construct a decidability instance for an extremal set-family property. \\
E-03 & Prove an edge-count lower bound under a Hamiltonian-walk condition. \\
\end{longtable}
\endgroup

The source supplies general Lean proof and interaction knowledge, but it cannot directly enumerate the domain-specific lemmas and proof sequences required by all of these statements. Execution feedback can expose where an attempted proof fails; when the original source does not contain the missing mathematical knowledge, source-grounded additions to the library have limited scope. This task--source gap helps explain why Lean 4 retains a relatively low success rate after self-evolution.

\subsection{TeXstudio: Editor Operations and Cross-Application Information}
\label{app:texstudio-task-scope}

The TeXstudio initial library is constructed from the TeXstudio manual (\url{https://texstudio-org.github.io/}). Its contents cover document editing, compilation, PDF viewing, and editor configuration (Figure~\ref{fig:texstudio-source-scope}). Table~\ref{tab:texstudio-task-scope} shows that some Verified tasks require only TeX editing or compilation, while others first require a measurement, object property, computation, or bibliographic record from another application or website.

\noindent\begin{minipage}{\linewidth}
    \centering
    \includegraphics[width=0.95\linewidth]{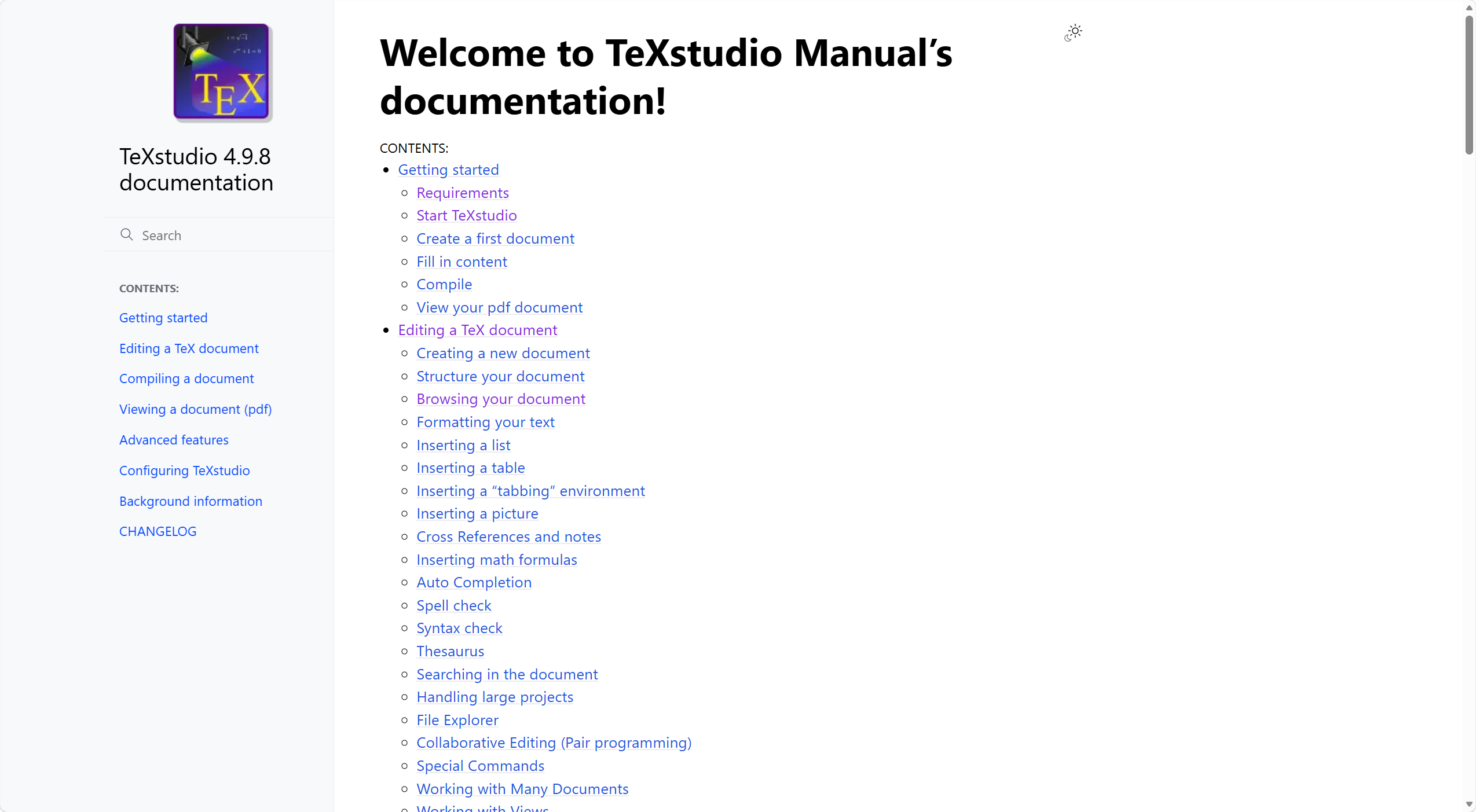}
    \captionof{figure}{Landing page and table of contents of the TeXstudio manual used for the initial library.}
    \label{fig:texstudio-source-scope}
\end{minipage}\par\smallskip

\begingroup
\small
\setlength{\LTpre}{4pt}
\setlength{\LTpost}{4pt}
\renewcommand{\arraystretch}{1.08}
\begin{longtable}{@{}p{0.09\linewidth}p{0.43\linewidth}p{0.39\linewidth}@{}}
\caption{TeXstudio Verified tasks and information required beyond editor operation. The entries summarize the full task instructions supplied with ScienceBoard-VE.}\label{tab:texstudio-task-scope}\\
\toprule
Task & Task goal & Additional information or tool \\
\midrule
\endfirsthead
\multicolumn{3}{@{}c@{}}{\small\itshape Table \thetable{} (continued)}\\
\toprule
Task & Task goal & Additional information or tool \\
\midrule
\endhead
\bottomrule
\endlastfoot
A-01 & Enter a molecular angle in the article. & Measurement in ChimeraX. \\
A-02 & Enter a residue name in the article's table. & Residue information from ChimeraX. \\
A-03 & Correct a DNA sequence in the article. & Sequence information from ChimeraX. \\
A-04 & Correct a Fibonacci value in the article. & Computation in KAlgebra. \\
B-01 & Complete a \texttt{\textbackslash cite} command for an existing bibliography entry. & None. \\
B-02 & Enter the COSPAR ID of Gemini 9A and add a citation. & Object information from Celestia. \\
B-03 & Enter the number of provinces in Piemonte and add a citation. & Geographic information from GRASS GIS. \\
B-04 & Add the BibTeX record for a specified paper to \texttt{refs.bib}. & Bibliographic search on Google Scholar. \\
C-01 & Correct an atom count in the article. & Count obtained in ChimeraX. \\
C-02 & Correct one letter in a $\lambda C$ deduction rule. & Knowledge of the deduction rule. \\
C-03 & Repair an incomplete figure path. & The project's actual file path. \\
C-04 & Change only the width of a specified figure. & None. \\
C-05 & Correct a Lean 4 code fragment in the article. & Lean code knowledge; VS Code can be used for debugging. \\
C-06 & Revise a formula until the document compiles. & TeX compilation feedback. \\
D-01 & Read the document and answer the value of $BB(4)$. & Knowledge needed to resolve the question. \\
D-02 & Read the document and answer the value of $B(5)$. & Knowledge needed to resolve the question. \\
\end{longtable}
\endgroup

Many tasks therefore have two components: obtaining the content to insert and editing or compiling the TeX document. TeXstudio documentation can support the latter, but it does not supply measurements or object information from other scientific applications. A library constructed and evolved from TeXstudio documentation alone cannot cover all of these cross-application requirements, which helps explain its more limited improvement in Table~\ref{tab:self_evolution_ablation}.

\clearpage
\section{Research Setting and Baseline Reproduction Constraints of Resource2Skill}
\label{app:resource2skill-scope}

\subsection{Research Setting and Knowledge Objectives}

Resource2Skill and ScienceCUA both investigate how to transform human-created resources into reusable skills, but they differ in their research settings and knowledge objectives.

Resource2Skill primarily addresses authoring tasks across domains such as PPT, Web, Excel, Blender, CAD, UE5, and audio production. It distills skills from tutorial videos, repositories, articles, documentation, and reference artifacts. Its library provides retrievable and composable authoring techniques, code patterns, and visual references to help agents create new artifacts, with downstream evaluation primarily measuring artifact quality.

ScienceCUA aims to improve the ability of models and agents to operate scientific software. Starting from supplied PDF/HTML software documentation, it discovers supported operational tasks, derives task-level procedures, and organizes shared knowledge such as commands, syntax, and parameter constraints. These forms of knowledge are represented as Skills and References, respectively, to support agents in completing user tasks in deployed software environments. Downstream evaluation primarily measures task success.

Thus, although both approaches investigate resource-to-skill construction, they differ in how resources are used, how knowledge is organized, and what downstream evaluation emphasizes.

\subsection{Limitations in Adapting the Released Implementation}

At the implementation level, the relevant public interfaces of Resource2Skill do not directly provide the document-level knowledge-construction process required for our collections of PDF/HTML software documentation. Although the paper describes a construction operator that can generate multiple candidate skills from a single resource, the two interfaces directly relevant to our documentation inputs have the following limitations in the public code we inspected.

\begin{enumerate}
  \item \textbf{\texttt{StaticArtifactConnector} does not directly support PDF manual parsing.} Its directory mode filters files using domain-specific extension lists, covering \texttt{.xlsx}, \texttt{.html/.htm}, \texttt{.pptx}, \texttt{.blend}, and selected audio and project formats, but excluding PDF. Consequently, the KAlgebra and Celestia PDF manuals used in our experiments cannot be directly processed through this mode. Its catalogue mode constructs candidates from supplied JSON fields, including descriptions and applicability; it does not itself parse the referenced manuals, discover knowledge units, or generate operational knowledge. \href{https://github.com/microsoft/Resource2Skill/blob/main/core/sources/static_artifact.py}{StaticArtifactConnector implementation}.

  \item \textbf{\texttt{ArticleConnector} uses individual supplied pages as candidate-generation units.} It accepts a URL list, fetches each page, and produces at most one \texttt{CandidateSkill} per URL. Its optional model call primarily extracts a title, applicability, tags, and a short summary. The interface does not implement recursive documentation crawling, cross-page evidence aggregation, or the discovery of multiple procedural knowledge units within a page. Supplying a documentation homepage or a list of page URLs therefore does not directly provide a complete workflow for constructing operational knowledge from an entire documentation collection. \href{https://github.com/microsoft/Resource2Skill/blob/main/core/sources/article.py}{ArticleConnector implementation}.
\end{enumerate}

Adapting these interfaces to our setting would require additional components for PDF parsing, HTML documentation discovery and collection, knowledge-unit decomposition, cross-page content organization, and procedural-content generation, before packaging the results as \texttt{CandidateSkill} records. Resource2Skill's subsequent candidate processing and Skill Wiki construction remain reusable. However, the added components would directly determine the content, granularity, and coverage of the resulting library---properties central to the evaluation of resource-to-skill construction.

Such an adaptation would therefore extend beyond input-format conversion and introduce a knowledge-extraction method designed by us. Its downstream performance would make the respective contributions of Resource2Skill and the added extraction procedure difficult to distinguish. Given these differences in knowledge objectives and the adaptation requirements of the released implementation, we did not include Resource2Skill as an original-pipeline baseline alongside VisualSkill and SkillFoundry under the same evaluation setting.

\section{Supplementary Evaluation of Frontier Models}
\label{app:frontier-models}

We additionally evaluate Claude Opus 5.5 and GPT-6 Astra on ScienceBoard-Verified, both with the \texttt{xhigh} reasoning setting. Claude Opus 5.5 shows comparatively limited computer-use performance in this scientific-software evaluation, so we do not further evaluate it with Skills and References in this supplementary study. GPT-6 Astra demonstrates substantially stronger performance. We therefore evaluate it with ScienceCUA's self-evolved Skills and References to examine whether a stronger base model can still benefit from external operational knowledge.

\begin{figure}[htbp]
    \centering
    \includegraphics[width=0.6\linewidth]{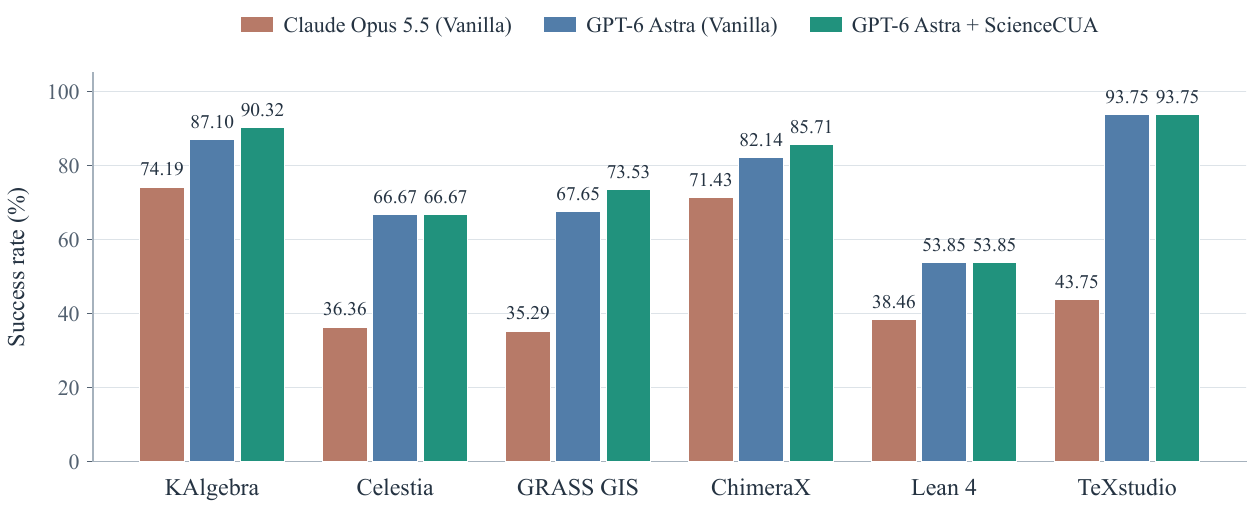}
    \caption{Success rates across the six applications in ScienceBoard-Verified. All models use the \texttt{xhigh} reasoning setting. ScienceCUA provides Skills and References that have undergone self-evolution. Labels above the bars give success rates in percent.}
    \label{fig:frontier-models}
\end{figure}

As shown in Figure~\ref{fig:frontier-models}, adding ScienceCUA's self-evolved library increases GPT-6 Astra's average success rate from 75.19\% to 77.30\%, a gain of 2.11 percentage points. These results show that the self-evolved library still provides a modest additional benefit on this stronger base model, supporting the discussion in Appendix~\ref{app:research-takeaways}.

\end{document}